\documentclass{article}

\newif \ifarxiv
\arxivtrue

\ifarxiv 
    \usepackage[legalpaper, margin=1.45in]{geometry}
    \usepackage{palatino}
    \date{}
\else

\fi

\usepackage[numbers]{natbib}
\usepackage[utf8]{inputenc} %
\usepackage[T1]{fontenc}    %
\usepackage{hyperref}       %
\usepackage{url}            %
\usepackage{booktabs}       %
\usepackage{amsfonts}       %
\usepackage{nicefrac}       %
\usepackage{microtype}      %
\usepackage{epsfig}
\usepackage{graphicx}
\usepackage{amsmath,amssymb}

\usepackage{glossaries}
\usepackage[font=footnotesize,skip=4pt]{caption}
\usepackage{subcaption}
\usepackage{multirow}
\usepackage{xcolor, colortbl}
\usepackage{pifont}%
\usepackage{xspace}
\usepackage{algorithm}
\usepackage{dblfloatfix}
\usepackage{transparent}
\usepackage{listings}
\usepackage{enumitem}
\usepackage{tabu}
\usepackage{subcaption}
\usepackage{booktabs}
\usepackage{array}

\usepackage{comment}

\def \pzo {\phantom{0}} 
\def \dzo {\phantom{00}}

\def \alambic {$\Upsilon$\xspace}
\def \alambicb {$\Upsilon$\xspace}

\def \ours          {XCiT\xspace}
\def \OURS          {XCiT\xspace}
\def \ImNet {ImageNet\xspace}

\def \etal {\textit{et al.}\xspace}
\def \eg {\textit{e.g.}\xspace}

\def \ie {\textit{i.e.}\xspace}

\def \wrt {{w.r.t.}\xspace}

\def\be{\begin{equation}}
\def\ee{\end{equation}}
\def\bea{\begin{eqnarray}}
\def\eea{\end{eqnarray}}
\def\fig#1{Figure~\ref{fig:#1}}
\def\tab#1{Table~\ref{tab:#1}}

\graphicspath{{.}{figs/}}

\usepackage{color, colortbl}
\definecolor{Gray}{gray}{0.9}
\definecolor{Goldenrod}{RGB}{245,245,220}
\title{\ours: Cross-Covariance Image Transformers}

\ifarxiv
    \makeatletter
    \renewcommand{\paragraph}{%
    \@startsection{paragraph}{4}%
    {\z@}{0.74em}{-1em}%
    {\normalfont\normalsize\bfseries}%
    }
    \makeatother
\fi 

\makeatletter
\let\inserttitle\@title
\makeatother

\author{%
\begin{minipage}{\linewidth}
\begin{center}
\scalebox{1.}{\normalsize Alaaeldin El-Nouby$^{1,2}$ \hspace{0.35cm} Hugo Touvron$^{1,3}$ \hspace{0.35cm} Mathilde Caron$^{1,2}$ \hspace{0.35cm}  Piotr Bojanowski$^{1}$ }\\[0.2cm]
\scalebox{1.}{\normalsize Matthijs Douze$^{1}$ \hspace{0.35cm} Armand Joulin$^{1}$ \hspace{0.35cm} Ivan Laptev$^{2}$ \hspace{0.35cm} Natalia Neverova$^{1}$ }\\[0.2cm]
\scalebox{1.}{\normalsize Gabriel Synnaeve$^{1}$ \hspace{0.35cm} Jakob Verbeek$^{1}$ \hspace{0.35cm} Herv\'e J\'egou$^{1}$}\\[0.5cm]
\scalebox{0.95}{\normalsize \textmd{$^1$Facebook AI\hspace{0.6cm} $^2$Inria\hspace{0.6cm} $^3$Sorbonne University\hspace{0.6cm}}}
\\[1cm]
\end{center}
\end{minipage}
}

\begin{document}

\maketitle

\iftrue

\newcommand{\alaa}[1]{{\color{green!40!red}[\textbf{Alaa}: #1]}}
\newcommand{\rv}[1]{{\color{red}[\textbf{Rv}: #1]}}
\newcommand{\hugo}[1]{{\color{red}[\textbf{Hugo}: #1]}}
\newcommand{\nn}[1]{{\color{blue}[\textbf{NN}: #1]}}
\newcommand{\ivan}[1]{{\color{magenta}[\textbf{Ivan}: #1]}}
\newcommand{\jkb}[1]{{\color{olive}[\textbf{jkb}: #1]}}
\newcommand{\gab}[1]{{\color{purple}[\textbf{gab}: #1]}}

\else
\newcommand{\alaa}[1]{}
\newcommand{\rv}[1]  {}
\newcommand{\hugo}[1]{}
\newcommand{\nn}[1]  {}
\newcommand{\ivan}[1]{}
\newcommand{\jkb}[1] {}
\fi

\begin{abstract}
Following tremendous success in natural language processing, transformers have recently shown much promise for computer vision. 
The self-attention operation underlying  transformers yields global  %
interactions between all tokens, \ie words or image patches, and enables flexible modelling of image data beyond the local %
interactions of convolutions.
This flexibility, however, comes with a quadratic complexity in time and memory,  hindering application to long sequences and 
 high-resolution images.
We  propose  a ``transposed'' %
version of self-attention that operates across feature channels rather than tokens, where the
interactions 
are based on 
the cross-covariance matrix between keys and queries. %
The resulting cross-covariance attention (XCA) has linear complexity in the number of tokens, and allows %
 efficient processing of high-resolution images.
Our cross-covariance image transformer (\ours) -- built upon XCA --
 combines the accuracy of conventional transformers with the scalability of convolutional architectures.
We validate the effectiveness and generality of 
\ours
by reporting excellent results on multiple %
vision benchmarks, including (self-supervised) image classification  on \ImNet-1k, object detection and instance segmentation on COCO, and semantic segmentation on ADE20k. 

\makeatletter{\renewcommand*{\@makefnmark}{}
\footnotetext{Code: \url{https://github.com/facebookresearch/xcit}}\makeatother}

\end{abstract}

\section{Introduction}

Transformers architectures~\cite{vaswani2017attention} have provided quantitative and qualitative breakthroughs in speech and natural language processing (NLP). 
Recently, \citet{dosovitskiy2020image}  established transformers as  a viable architecture for %
learning visual representations, reporting competitive results for image classification while relying on large-scale pre-training. 
\citet{touvron2020deit} have shown on par or better accuracy/throughput compared to strong convolutional baselines such as EfficientNets~\cite{tan2019efficientnet} when training transformers on \ImNet-1k using extensive data augmentation and improved training schemes.
Promising results have been obtained for other vision tasks, including image retrieval~\cite{el2021training}, object detection and semantic segmentation~\cite{liu2021swin, wang2021pyramid, zhang2021multi, zheng2020rethinking}, as well as video understanding~\cite{arnab2021vivit, bertasius2021space, fan2021multiscale}. 

One major drawback of transformers is the time and memory complexity of the core self-attention operation, that increases quadratically with the number of input tokens, or similarly number of patches in computer vision. 
For $w{\times}h$ images, this translates to a complexity of ${\mathcal O}(w^2h^2)$, which  is prohibitive for most tasks involving high-resolution images, such as object detection and segmentation.
Various strategies have been proposed to alleviate this complexity, for instance using approximate forms of self-attention~\cite{liu2021swin, zhang2021multi}, or pyramidal architectures which progressively downsample the feature maps~\cite{wang2021pyramid}. 
However, none of the existing solutions are fully satisfactory, as they either trade complexity for accuracy, or their complexity %
remains excessive for processing very large images. 

\begin{figure*}[t!]
    \centering
    \ifarxiv
        \includegraphics[width=\linewidth]{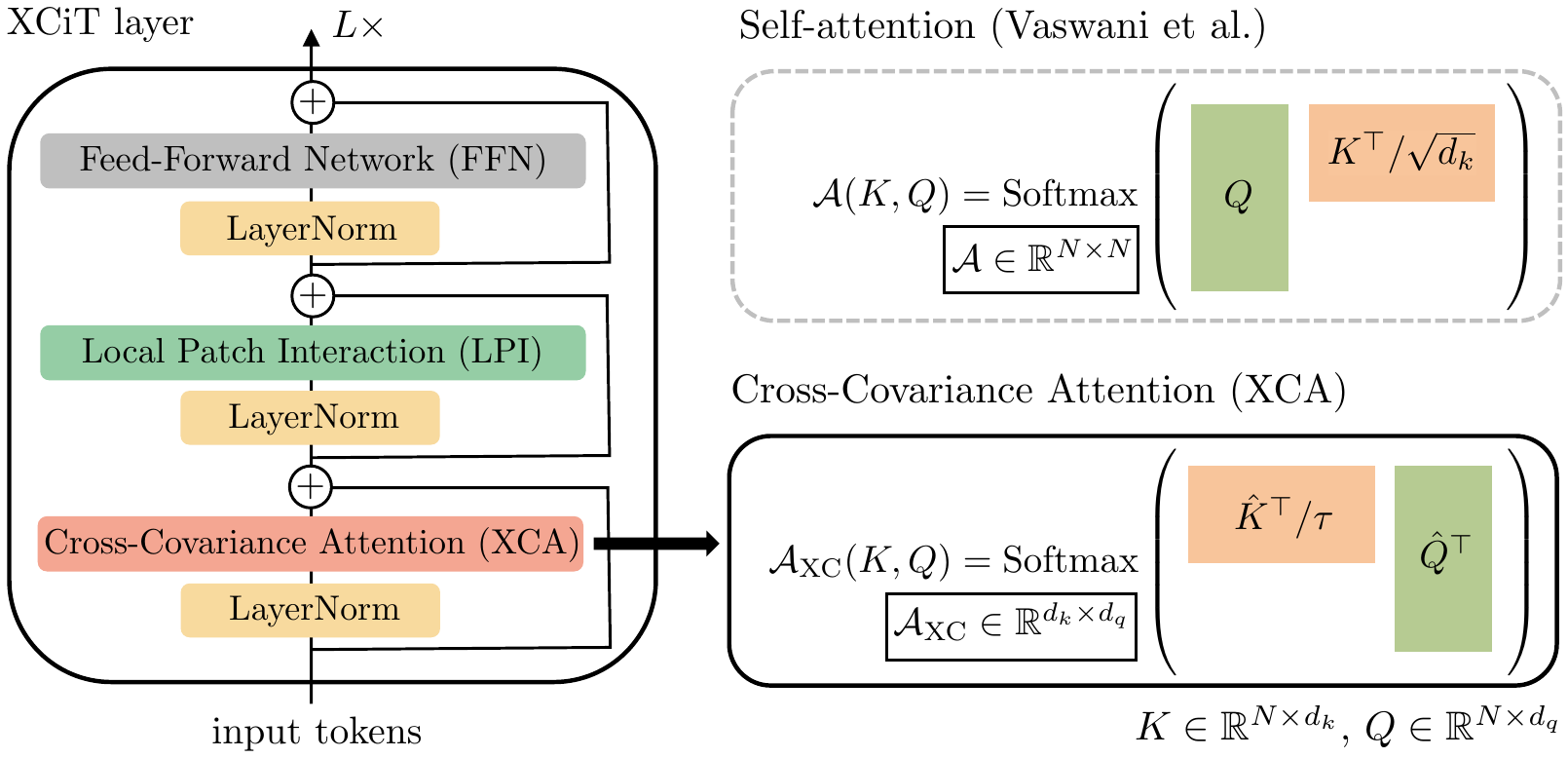}
    \else
        \includegraphics[width=0.82\linewidth]{figs/xcit_layer.pdf}
    \fi
    \caption{    \footnotesize
    Our \ours layer consists of three main blocks, each preceded by LayerNorm and followed by a residual connection: (i) the core  cross-covariance attention (XCA) operation, (ii) the local patch interaction (LPI) module, and  (iii) a feed-forward network (FFN). 
    By transposing the query-key interaction, the computational complexity of XCA is linear in the number of data elements $N$, rather than quadratic as in conventional self-attention. 
    }
    \label{fig:scat_layer}
\end{figure*}

We replace the self-attention, as originally introduced by~\citet{vaswani2017attention}, with a ``transposed'' attention that we denote as ``cross-covariance attention'' (XCA). Cross-covariance attention substitutes the explicit full pairwise  interaction between tokens by self-attention among features, where the attention map 
is derived from the cross-covariance matrix computed over the key and query  projections of the token features. 
Importantly, XCA has a linear complexity in the number of patches.
To construct our  Cross-Covariance Image Transformers (\ours), we combine XCA with  local patch interaction modules that rely on efficient depth-wise convolutions and point-wise feedforward networks commonly used in transformers, see \fig{scat_layer}.
XCA  can be regarded as a form of a dynamic $1\!\times\!1$ convolution, which multiplies all tokens with the same data-dependent weight matrix. 
We find that the performance of our XCA layer can be further improved by applying it on blocks of channels, rather than directly mixing all channels together.
This ``block-diagonal'' shape of XCA further reduces the computational complexity with a factor linear in the number of blocks. 

Given its linear complexity in the number of tokens, %
\ours  can efficiently process images with more than thousand pixels in each dimension. 
Notably, our experiments show that \ours does not compromise the accuracy and achieves similar results to DeiT~\cite{touvron2020deit} and CaiT~\cite{touvron2021going} in comparable settings.
Moreover, for dense prediction tasks such as object detection and image segmentation, our models outperform popular ResNet~\cite{he2016deep} backbones as well as the recent transformer-based models \cite{liu2021swin, wang2021pyramid, zhang2021multi}.
Finally, we also successfully apply \ours to the self-supervised feature learning using DINO~\cite{caron2021emerging}, and demonstrate improved performance compared to a DeiT-based backbone~\cite{touvron2020deit}.

Overall, we summarize our contributions as follows:
\begin{itemize}[leftmargin=*]
\item We introduce cross-covariance attention (XCA), which  provides a ``transposed'' alternative  to conventional self-attention, attending over channels instead of tokens. Its complexity is linear in the number of tokens, %
allowing for
efficient processing of high-resolution images, see Figure~\ref{fig:peak_mem}. 

\item  XCA attends to a fixed number of channels, irrespective of the number of tokens. As a result, our models are significantly more robust to changes in image resolution at test time, and are therefore more amenable to process variable-size images.

\item For image classification, we demonstrate that our models are on par with state-of-the-art vision transformers for multiple model sizes using a simple columnar architecture, \emph{i.e.}, in which we keep the resolution constant across layers. %
In particular, our \ours-L24 model achieves 86.0\% top-1 accuracy on ImageNet, outperforming its CaiT-M24~\cite{touvron2021going} and NFNet-F2~\cite{brock2021high} counterparts with comparable numbers of parameters.

\item  For dense prediction tasks with high-resolution images, our models outperform ResNet and multiple  transformer-based backbones. On the COCO benchmark, we achieve a strong performance of 48.5\% and 43.7\% mAP for object detection and instance segmentation respectively. Moreover, we report 48.4\% mIoU for semantic segmentation on the ADE20k benchmark, outperforming the state-of-the-art Swin Transformer~\cite{liu2021swin} backbones across all comparable model sizes. 

\item Finally, our \ours model is highly %
effective 
in self-supervised learning setups, achieving 80.9\% top-1 accuracy on ImageNet-1k using DINO~\cite{caron2021emerging}.
\end{itemize}

\section{Related work}
\label{sec:related}

\paragraph{Deep vision transformers.}
Training deep vision transformers can be challenging due to instabilities and optimization issues.
\citet{touvron2021going} successfully train models with up to 48 layers using LayerScale, which  weighs contributions of residual blocks across layers and improves optimization. 
Additionally, the authors 
introduce class attention layers which decouple the learning of patch features and the feature aggregation stage for classification.

\paragraph{Spatial structure in vision transformers.}
\citet{yuan2021tokens} propose applying a soft split for patch projection with overlapping patches which is applied repeatedly across model layers, reducing the number of patches progressively. 
\citet{han2021transformer} introduce a transformer module for intra-patch structure, exploiting pixel-level information and integrating with an  inter-patch transformer to attain higher representation power.
\citet{d2021convit} consider the initialization of  self-attention blocks as a convolutional operator, and demonstrate that such initialization improves the performance of vision transformers in low-data regimes. 
\citet{graham2021levit} introduce LeViT,  which adopts a multi-stage architecture with progressively reduced feature resolution similar to popular convolutional architectures, allowing for models with high inference speed while retaining a strong performance. Moreover, the authors adopt a convolution-based module for extracting patch descriptors.  
\citet{yuan2021incorporating} improve both the performance and the convergence speed of vision transformers by replacing the linear patch projection with convolutional layers and max-pooling, as well as modifying the feed-forward networks in each transformer layer to incorporate depth-wise convolutions. 

\paragraph{Efficient attention.}
Numerous methods for efficient self-attention have been proposed in the  literature to address the quadratic complexity of self-attention in  the number of input tokens. 
These include restricting  the span of the self-attention to local windows~\cite{parmar2018image,qiu2019blockwise}, strided patterns~\cite{child2019generating}, axial patterns~\cite{ho2019axial}, or an adaptive computation across layers \cite{sukhbaatar2019adaptive}. 
Other methods provide an approximation of the self-attention matrix which can be achieved by a projection across the token dimension~\cite{wang2020linformer}, or through a factorization of the softmax-attention kernel~\cite{choromanski2020rethinking, katharopoulos2020transformers, shen2021efficient, xiong2021nystr}, which avoids  explicit computation of the attention matrix. 
While conceptually different, our XCA performs similar computations without being sensitive to the choice of the kernel. 
Similarly, \citet{lee2021fnet}~achieve faster training by substituting self-attention with unparametrized Fourier Transform.
Other efficient attention methods rely on local attention and adding a small number of global tokens, thus allowing interaction among all tokens only by hopping through the global tokens \cite{ainslie2020etc, beltagy2020longformer, jaegle2021perceiver, zaheer2020big}.

\paragraph{Transformers for high-resolution images.}
Several works adopt visual transformers to high-resolution image tasks beyond image classification, such as object detection and image segmentation. 
\citet{wang2021pyramid} design a model with a pyramidal architecture and %
address complexity by gradually reducing the spatial resolution of keys and values.
Similarly, for video recognition \citet{fan2021multiscale} utilize pooling to reduce the resolution across the spatial and temporal dimensions to allow for an efficient computation of the attention matrix. \citet{zhang2021multi} adopt global tokens and local attention to reduce the model complexity, while \citet{liu2021swin} provide an efficient method for local attention with shifted windows. In addition, \citet{zheng2020rethinking} and \citet{ranftl2021vision} study problems like semantic segmentation and monocular depth estimation with the quadratic self-attention operation.

\paragraph{Data-dependent layers.} 
Our \ours layer can be regarded as a ``dynamic'' $1{\times}1$ convolution, which multiplies all token features with the same data-dependent weight matrix, derived from the key and query cross-covariance matrix.
In the context of convolutional networks, Dynamic Filter Networks~\cite{brabandere16nips} explore a related idea, using a filter generating subnetwork to produce convolutional filters based on features in previous layers.
Squeeze-and-Excitation networks~\cite{hu2018squeeze}  use data dependent  $1\!\times\!1$ convolutions in convolutional architectures. Spatially average-pooled features are fed to a 2-layer MLP which produces per channel scaling parameters.
Closer in spirit to our work, Lambda layers propose a way to ensure global interaction in ResNet models \cite{bello2021lambdanetworks}.
Their ``content-based lambda function'' is computing  a similar term as our cross-covariance attention, but differing in how the softmax and $\ell_2$ normalizations are applied.
Moreover, Lambda layers also include specific position-based lambda functions, and  LambdaNetworks are based on ResNets while XCiT follows the ViT architecture.
Recently \emph{data-independent} analogues of self-attention have also been found to be an effective alternative to convolutional and self-attention layers for vision tasks~\cite{ding2021repmlp,melaskyriazi2021doyoueven,Tolstikhin21mixer,Touvron21ResMLP}. %
These methods treat entries in the attention map as learnable parameters, rather than deriving the attention map dynamically from queries and keys, but their complexity remains quadratic in the number of tokens.
Zhao \etal~\cite{zhao20cvpr} %
consider alternative  attention forms in computer vision.

\section{Method}
\label{sec:methods}

In this section, we first recall the self-attention mechanism, and the connection between the Gram and covariance matrices, which motivated our work.
We then propose our  cross-covariance attention operation (XCA) -- 
 which operates along the feature dimension instead of token dimension in conventional transformers --
and combine it with local patch interaction and feedforward layers
to construct our Cross-Covariance Image Transformer (\ours). 
See Figure~\ref{fig:scat_layer} for an overview.

\subsection{Background}

\paragraph{Token self-attention.} 
Self-attention, as introduced by~\citet{vaswani2017attention}, operates on an input matrix $X \in \mathbb{R}^{N \times d}$, where $N$ is the number of tokens, each of dimensionality $d$. 
The input  $X$ is linearly projected to queries, keys and values,  using the weight matrices $W_{q} \in \mathbb{R}^{d \times d_q}$, $W_{k} \in \mathbb{R}^{d \times d_k}$ and $W_{v} \in \mathbb{R}^{d \times d_v}$, such that $Q{=}X W_q $, $K{=}X W_k $ and $V{=}X W_v $, where $d_q\!=\!d_k$.
Keys and values are used to compute an attention map 
$\mathcal{A}(K,Q) = \text{Softmax}(Q K^\top/\sqrt{d_k})$,  
and the output of the self-attention operation is defined as the weighted sum of $N$ token features in $V$ with the weights corresponding to the attention map: 
$\text{Attention}(Q, K, V) = \mathcal{A}(K, Q) V$.
The computational complexity of self-attention scales quadratically in $N$, due to pairwise interactions between all  $N$ elements. 

\paragraph{Relationship between Gram and covariance matrices.}
To motivate our cross-covariance attention operation, we recall the relation between Gram and covariance matrices.
The unnormalised $d\!\times\!d$ covariance matrix
is obtained as $C{=}X^\top X$. 
The $N{\times}N$ Gram matrix contains all pairwise innerproducts: $G{=}X X^\top$. The non-zero part of the  eigenspectrum of the  Gram and covariance matrix are equivalent, and the eigenvectors of $C$ and $G$ can be computed in terms of each other. 
If $V$ are the eigenvectors of $G$, %
then the eigenvectors of $C$ are given by $U{=}XV$. 
To minimise the computational cost, the eigendecomposition of either the Gram or covariance matrix can be obtained in terms of the decomposition of the other, depending on which of the two matrices is the smallest.\footnote{
For $C$ to represent the covariance, $X$ should be centered, 
\ie $X\mathbf{1}  {=} \mathbf{0}$.
For the  relation between $C$ and $G$, however, centering is not required. 
}

We draw upon this strong connection between the Gram and covariance matrices to consider if it is possible to avoid the quadratic cost to compute the $N\!\times\!N$ attention matrix, which is computed from  the analogue of the $N\!\times\!N$ Gram matrix $Q K^\top {=} X W_qW_k^\top X^\top$.  
Below we consider how we can use the $d_k\!\times\!d_q$ cross-covariance matrix, $K^\top Q {=}  W_k^\top X^\top X W_q $, which can be computed in linear time in the number of elements $N$, 
 to define an attention mechanism.

\begin{figure*}[t!]
    \begin{minipage}[t]{0.48\textwidth}
    \centering
        \includegraphics[trim=0.15cm 0.1cm 0 0, clip, width=1.1\linewidth]{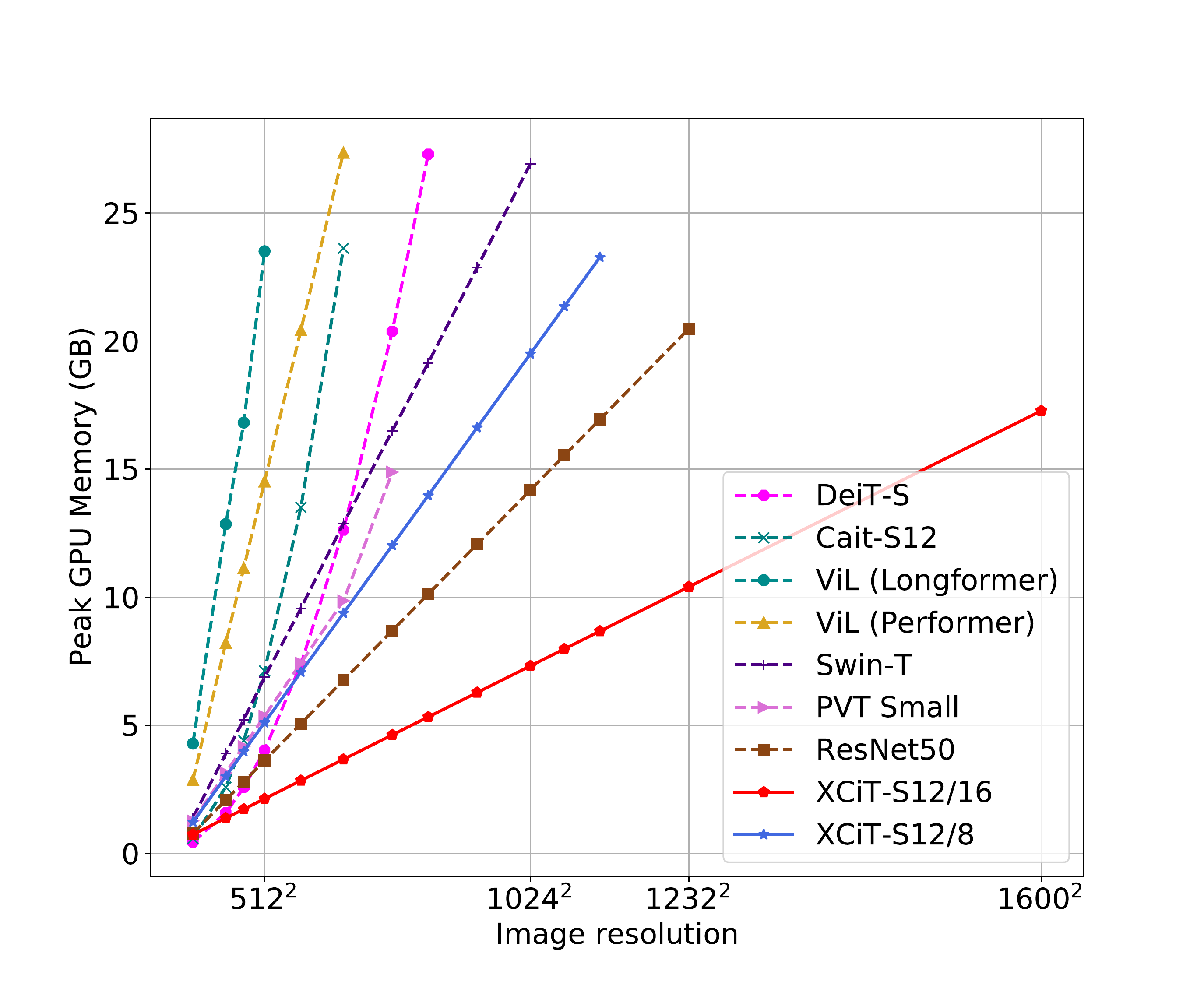}
    \caption{%
    Inference memory usage of vision transformer variants. %
    Our \ours models scale linearly in the number of tokens, which makes it possible to scale to much larger image sizes, even in comparison to approaches employing approximate self-attention or a pyramidal design.
    All measurements are performed with a batch size of 64 on a single V100-32GB GPU. 
    \label{fig:peak_mem}}
    \end{minipage}%
    \hfill
    \begin{minipage}[t]{0.48\textwidth}
    \centering
    \includegraphics[trim=0.15cm 0.1cm 0 0, clip, width=1.1\linewidth]{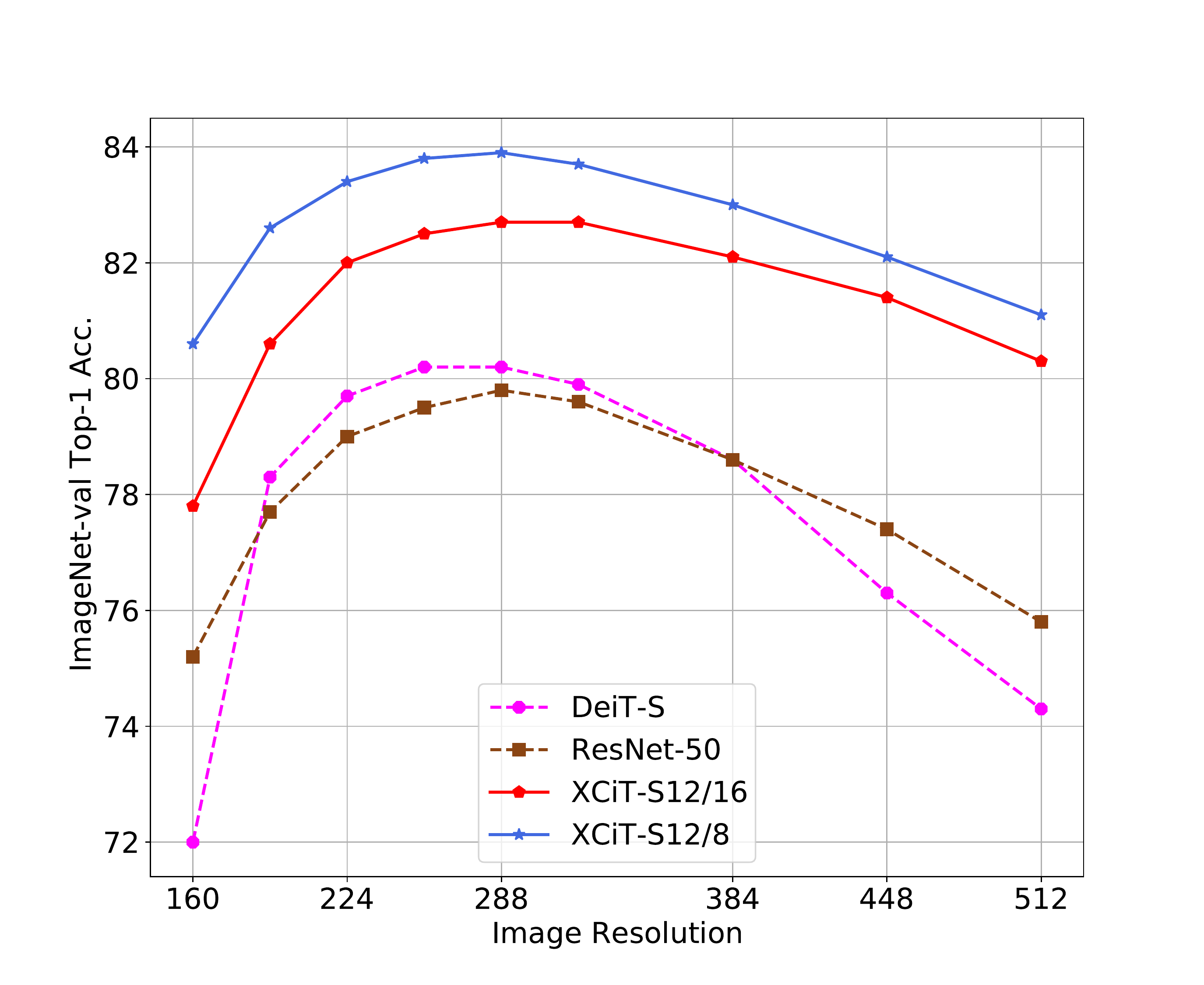}
    \captionof{figure}{Performance when changing the resolution at test-time for models with a similar number of parameters. All networks were trained at resolution 224, w/o distillation. \OURS is more tolerant to changes of resolution than the Gram-based DeiT and benefit more from the ``FixRes'' effect~\cite{touvron2019fixing} when inference is performed at a larger resolution than at train-time.}
    \label{fig:change_res_testtime}
    \strut\end{minipage}%
    \vspace{-5mm}
\end{figure*}

\subsection{Cross-covariance attention}

We propose a cross-covariance based self-attention function that operates along the feature dimension, rather than along the token dimension as in token self-attention. 
Using the definitions of queries, keys and values from above, 
the 
cross-covariance attention function is defined as:
\bea    
\text{XC-Attention}(Q, K, V) = V \mathcal{A}_\text{XC}( K, Q), \quad\quad
\mathcal{A}_\text{XC}( K,Q) = \text{Softmax}\left(\hat{K}^\top \hat{Q} /\tau\right),
    \label{eq:xatt}
\eea
where each output token embedding is %
a convex combination 
of the $d_v$ features of its corresponding token embedding in $V$. The attention weights $\mathcal{A}$ are computed based on the cross-covariance matrix.

\paragraph{$\ell_2$-Normalization and  temperature scaling.} 
In addition to building our attention operation on the cross-covariance matrix, we make a second modification compared to  token self-attention. 
We restrict the magnitude of the query and key matrices by $\ell_2$-normalising them, such that each column of length $N$ of the normalised matrices $\hat{Q}$ and $\hat{K}$ has unit norm, and  every element in $d{\times}d$ cross-covariance matrix $\hat{K}^\top \hat{Q}$ is in the range  $[-1, 1]$.
We observed that controlling the norm strongly enhances the stability of training, especially when trained with a variable numbers of tokens. 
However, restricting the norm  reduces the representational power of the operation by removing a degree of freedom. Therefore, we  introduce a learnable temperature parameter $\tau$ which scales the inner products before the Softmax, allowing for sharper or more uniform distribution of attention weights.

\paragraph{Block-diagonal cross-covariance attention.} 
Instead of allowing all features to interact among each other,  %
we divide them into a $h$ groups, or  ``heads'', in a similar fashion as multi-head token self-attention. We apply the cross-covariance attention separately per head where
for each head, we learn separate weight matrices to project $X$ to 
queries, keys and values, and collect the corresponding weight matrices in the tensors  $W_{q} \in \mathbb{R}^{h \times d \times d_q}$, $W_{k} \in \mathbb{R}^{h \times d \times d_k}$ and $W_{v} \in \mathbb{R}^{h \times d \times d_v}$, where we set $d_k {=} d_q {=} d_v {=} d / h$. 
Restricting the attention within heads has two advantages: 
(i)~the complexity of aggregating the values with the attention weights is reduced by a factor  $h$; 
(ii)~more importantly,  we empirically observe that the block-diagonal version is easier to optimize, and typically leads to improved results. 
This observation is in line with observations made for  Group Normalization~\cite{wu2018group}, which normalizes groups of channels separately based on their statistics, and achieves favorable results for computer vision tasks compared to Layer Normalization~\cite{ba2016layer}, which combines all channels in a single group. 
\fig{cls_gram_vis} shows that each head learns to focus on semantically coherent parts of the image, while being flexible to change what type of features it attends to based on the image content. 

\paragraph{Complexity analysis.} The usual token self-attention  %
with $h$ heads has a time complexity of $\mathcal{O}(N^2 d)$ and memory complexity of $\mathcal{O}(h N^2 {+} N d)$. 
Due to the quadratic complexity, it is problematic to scale token self-attention  to images with a large number of tokens. 
Our cross-covariance attention overcomes this drawback as its computational cost of $\mathcal{O}({N d^2}/{h})$  scales linearly with the number of tokens, as does the memory complexity of $\mathcal{O}({d^2}/{h} {+} N d)$. 
Therefore, our   model scales much better to cases where the number of tokens $N$ is large, and the  feature dimension $d$ is relatively small, as is typically the case, in particularly when  splitting the features into $h$ heads.

\subsection{Cross-covariance image transformers} 

To construct our cross-covariance image transformers (\ours),
we adopt a columnar architecture which maintains the same spatial resolution across layers,
similarly to \cite{dosovitskiy2020image, touvron2020deit, touvron2021going}. 
We combine  our cross-covariance attention (XCA) block with the following   additional modules, each one being preceded by a LayerNorm~\cite{ba2016layer}.
See \fig{scat_layer} for an  overview. 
Since in this section we specifically design the model for computer vision tasks, tokens  correspond to image patches in this context.

\paragraph{Local patch interaction.} 
In the XCA block communication between  patches is only implicit through the shared statistics. 
To enable explicit  communication across patches we add a simple Local Patch Interaction (LPI) block after each XCA block.
LPI consists of two depth-wise $3{\times}3$ convolutional layers with Batch Normalization and GELU non-linearity in between.
Due to its depth-wise structure, the LPI block has a negligible overhead in terms of parameters, 
as well as a very limited overhead in terms of throughput and memory usage during inference. 

\paragraph{Feed-forward network.} 
As is common in transformer models, we add a point-wise feedforward network (FFN), which has a single hidden layer with $4d$ hidden units.
While interaction between features is confined within groups in the XCA block, and no feature interaction takes place in the  LPI block, the FFN allows for  interaction across all features.

\paragraph{Global aggregation with class attention.} 
When training our models for image classification, we  utilize the class attention layers as proposed by~\citet{touvron2021going}. These layers aggregate the patch embeddings of the last \ours layer through writing to a  \texttt{CLS} token by one-way attention between the \texttt{CLS} tokens and the patch embeddings. 
The class attention is also applied per head, \ie feature group. 

\begin{table}[t!]
    \caption{\textbf{\OURS models}. %
    Design choices include model depth, patch embeddings dimensionality $d$, and the  number of heads $h$ used in XCA. By default our models are trained and tested at resolution 224 with patch sizes of 16$\times$16. We also train with distillation using a convolutional teacher (denoted \alambic) as proposed by \citet{touvron2020deit}. Finally, we report performance of our strongest models obtained with 8$\times$8 patch size, fine-tuned ($\uparrow$) and tested at resolution 384$\times$384 (column @384/8), using distillation with a teacher that was also fine-tuned @384. 
     \label{tab:xct_models}}\vspace*{2pt}
    \centering 
    \scalebox{0.83}{
    \hspace{-2pt}
    \begin{tabular}{l|ccc|r|rr|ccc}
    \toprule
         Model & Depth & $d$ & \#heads & \#params & \multicolumn{2}{c|}{GFLOPs} & \multicolumn{3}{c}{\ImNet-1k-val top-1 acc. (\%)}    \\ 
         & & &   &  & @224/16 & @384/8 & @224/16 & @224/16\alambic & @384/8\alambic $\uparrow$     \\
        \midrule
        \OURS-N12    & 12 & 128 &  4 &   3M &  0.5 & \dzo6.4   & 69.9 & 72.2 & 77.8 \\
        \OURS-T12    & 12 & 192 &  4 &   7M &  1.2 & \pzo14.3  & 77.1 & 78.6 & 82.4 \\
        \OURS-T24    & 24 & 192 &  4 &  12M &  2.3 & \pzo27.3  & 79.4 & 80.4 & 83.7 \\
        \OURS-S12    & 12 & 384 &  8 &  26M &  4.8 & \pzo55.6  & 82.0 & 83.3 & 85.1 \\ 
        \OURS-S24    & 24 & 384 &  8 &  48M &  9.1 & 106.0     & 82.6 & 83.9 & 85.6 \\ 
        \OURS-M24    & 24 & 512 &  8 &  84M & 16.2 & 188.0     & 82.7 & 84.3 & 85.8 \\
        \OURS-L24    & 24 & 768 & 16 & 189M & 36.1 & 417.9     & 82.9 & 84.9 & 86.0 \\
    \bottomrule     
 \end{tabular}
 } %
\vspace{-4mm}
\end{table}

\paragraph{Handling images of varying resolution.}
In contrast to the attention map involved in token self-attention, in our case the covariance blocks are of fixed size independent of the input image resolution. The softmax always operates over the same number of elements, which may explain why our models behave better when dealing with images of varying resolutions (see Figure~\ref{fig:change_res_testtime}). 
In \OURS we  include additive sinusoidal positional encoding~\cite{vaswani2017attention} with the input tokens. 
We generate them in 64 dimensions from the 2d patch coordinates and then linearly project to the transformer working dimension $d$. 
This choice is orthogonal to the use of  learned positional encoding, as in ViT~\cite{dosovitskiy2020image}. 
However, it is more flexible since there is no need to interpolate or fine-tune the network when changing the image size.

\paragraph{Model configurations.}
In Table~\ref{tab:xct_models} we list different variants of our model which we use in our experiments, with different choices for model width  and depth. 
For the patch encoding layer, unless mentioned otherwise, we adopt the alternative used by \citet{graham2021levit} with  convolutional patch projection layers. 
We also experimented with a linear patch projection as described in \cite{dosovitskiy2020image}, see our ablation in Table~\ref{tab:ablations}.
Our default patch size is $16\!\times\!16$, as in other vision transformer models including  ViT~\cite{dosovitskiy2020image}, DeiT~\cite{touvron2020deit} and~CaiT \cite{touvron2021going}.
We also experiment with smaller $8\!\times\!8$ patches, which  has been observed to improve  performance~\cite{caron2021emerging}.
Note that this is  efficient with \ours as its complexity scales linearly which the number of patches, while ViT, DeiT and CaiT scale quadratically.

\section{Experimental evaluation}
\label{sec:experiments}

\begin{figure}[t!]
\begin{minipage}[t]{0.5\textwidth}
    \captionof{table}{\textbf{\ImNet classification}. Number of parameters, FLOPs, image resolution, and top-1 accuracy on \ImNet-1k and \ImNet-V2. Training %
    strategies  
    vary across models, transformer-based models and the reported RegNet mostly follow  recipes from DeiT~\cite{touvron2020deit}.
    \ifarxiv
        \vspace*{-2pt}
    \else
        \vspace*{-10pt}
    \fi
    \label{tab:classification}} 
    \def \mysp {\hspace{5pt}}
    \resizebox{\textwidth}{!}{
    \begin{tabular}{l@{\mysp}r@{\mysp}r@{\mysp}c|c@{\mysp}c} %
    \toprule
    Model  & \#params & FLOPs & Res.   &  ImNet   &  V2 \\ %
    \toprule
    EfficientNet-B5 RA \cite{cubuk2020randaugment}\!\!\!\!\!\!\!\!\! & \pzo30M & \dzo9.9B   & $456$ & 83.7 & \_ \\ %
    RegNetY-4GF  \cite{radosavovic2020designing}  & \pzo21M & \pzo4.0B   & $224$  & 80.0 &  72.4 \\
    DeiT-S\alambicb  \cite{touvron2020deit}  & \pzo22M & \dzo4.6B   & $224$  & 81.2   & 68.5 \\
    Swin-T  \cite{liu2021swin}          & \pzo29M & \dzo4.5B  & $224$  & 81.3 & \_ \\
    CaiT-XS24\alambicb$\uparrow$ \cite{touvron2021going}& \pzo26M & \dzo19.3B  & $384$  & 84.1 & 74.1 \\
    \rowcolor{Goldenrod} 
    \OURS-S12/16\alambicb & \pzo26M & \dzo4.8B & $224$  & 83.3 & 72.5 \\
    \rowcolor{Goldenrod} 
    \OURS-S12/16\alambicb$\uparrow$ & \pzo26M & \dzo14.3B  & $384$  & 84.7 &  74.1 \\
    \rowcolor{Goldenrod} 
    \OURS-S12/8\alambicb$\uparrow$ & \pzo26M & \dzo55.6B  & $384$  &  \textbf{85.1} & \textbf{74.8} \\
    \midrule
    EfficientNet-B7 RA  \cite{cubuk2020randaugment}\!\!\!\!\!\!\!\!\! & \pzo66M & \pzo37.0B  & $600$  & 84.7 & \_ \\ %
    NFNet-F0   \cite{brock2021high}          & \pzo72M & \pzo12.4B  & $256$ & 83.6  & 72.6 \\
    RegNetY-8GF \cite{radosavovic2020designing}       & \pzo39M & \pzo8.0B   & $224$  & 81.7  & 72.4 \\
    TNT-B   \cite{yuan2021tokens} & \pzo66M & \pzo14.1B  & $224$       & 82.8 & \_  \\ %
    Swin-S \cite{liu2021swin}  & \pzo50M & \pzo8.7B & $224$ & 83.0 & \_  \\
    CaiT-S24\alambicb$\uparrow$ \cite{touvron2021going}   & \pzo47M & \dzo32.2B  & $384$  & 85.1 & 75.4 \\
    \rowcolor{Goldenrod} 
    \OURS-S24/16\alambicb & \pzo48M & \dzo9.1B  & $224$  & 83.9 &  73.3\\
    \rowcolor{Goldenrod} 
    \OURS-S24/16\alambicb$\uparrow$ & \pzo48M & \dzo26.9B  & $384$  & 85.1 & 74.6 \\
    \rowcolor{Goldenrod} 
    \OURS-S24/8\alambicb$\uparrow$ & \pzo48M & \dzo105.9B  & $384$  &  \textbf{85.6} & \textbf{75.7} \\
    \midrule
    Fix-EfficientNet-B8 \cite{touvron2020fixing}\!\!\!\!\!\!\!\!\! & \pzo87M & 89.5B  & $800$  & 85.7 &  75.9 \\
    RegNetY-16GF  \cite{radosavovic2020designing}     & \pzo84M & \pzo16.0B  & $224$ & 82.9 &  72.4 \\
    Swin-B$\uparrow$ \cite{liu2021swin}  & \pzo88M & \pzo47.0B & $384$ & 84.2 & \_ \\
    DeiT-B\alambicb$\uparrow$ \cite{touvron2020deit} & 87M  & 55.5B &  $384$ &  85.2 &  75.2 \\
    CaiT-S48\alambicb$\uparrow$ \cite{touvron2021going}  & \pzo89M & \dzo63.8B  & $384$  & 85.3 & \textbf{76.2} \\
    \rowcolor{Goldenrod} 
    \OURS-M24/16\alambicb & \pzo84M & \dzo16.2B & $224$  & 84.3 & 73.6 \\
    \rowcolor{Goldenrod} 
    \OURS-M24/16\alambicb$\uparrow$ & \pzo84M & \dzo47.7B  & $384$  & 85.4 & 75.1 \\
    \rowcolor{Goldenrod}
    \OURS-M24/8\alambicb$\uparrow$ & \pzo84M & \dzo187.9B  & $384$  &  \textbf{85.8} & 76.1 \\
    \midrule
    NFNet-F2    \cite{brock2021high}       & 194M    & \pzo62.6B  & $352$  & 85.1 & 74.3 \\
    NFNet-F3     \cite{brock2021high}        & 255M    & 114.8B     & $416$  & 85.7 & 75.2 \\
    CaiT-M24\alambicb$\uparrow$ \cite{touvron2021going}  & \pzo186M & \dzo116.1B & $384$  & 85.8 & 76.1 \\
    \rowcolor{Goldenrod} 
    \OURS-L24/16\alambicb & \pzo189M & \dzo36.1B  & $224$  & 84.9 & 74.6 \\
    \rowcolor{Goldenrod} 
    \OURS-L24/16\alambicb$\uparrow$ & \pzo189M & \dzo106.0B  & $384$  & 85.8 &  75.8 \\
    \rowcolor{Goldenrod}
    \OURS-L24/8\alambicb$\uparrow$ & \pzo189M & \dzo417.8B  & $384$  &  \textbf{86.0} & \textbf{76.6} \\
    \bottomrule
    \end{tabular}}

\strut\end{minipage}%
\hfill\allowbreak%
\begin{minipage}[t]{0.46\textwidth}
    \centering
    \vspace{0pt}
    \setlength\tabcolsep{0pt} %
    \renewcommand{\arraystretch}{0.5}
    \scalebox{1}{
\begin{tabular}{@{\hspace{-2pt}}l@{\hspace{1pt}}c@{\hspace{1pt}}r@{}}
          \includegraphics[width=0.33\linewidth]{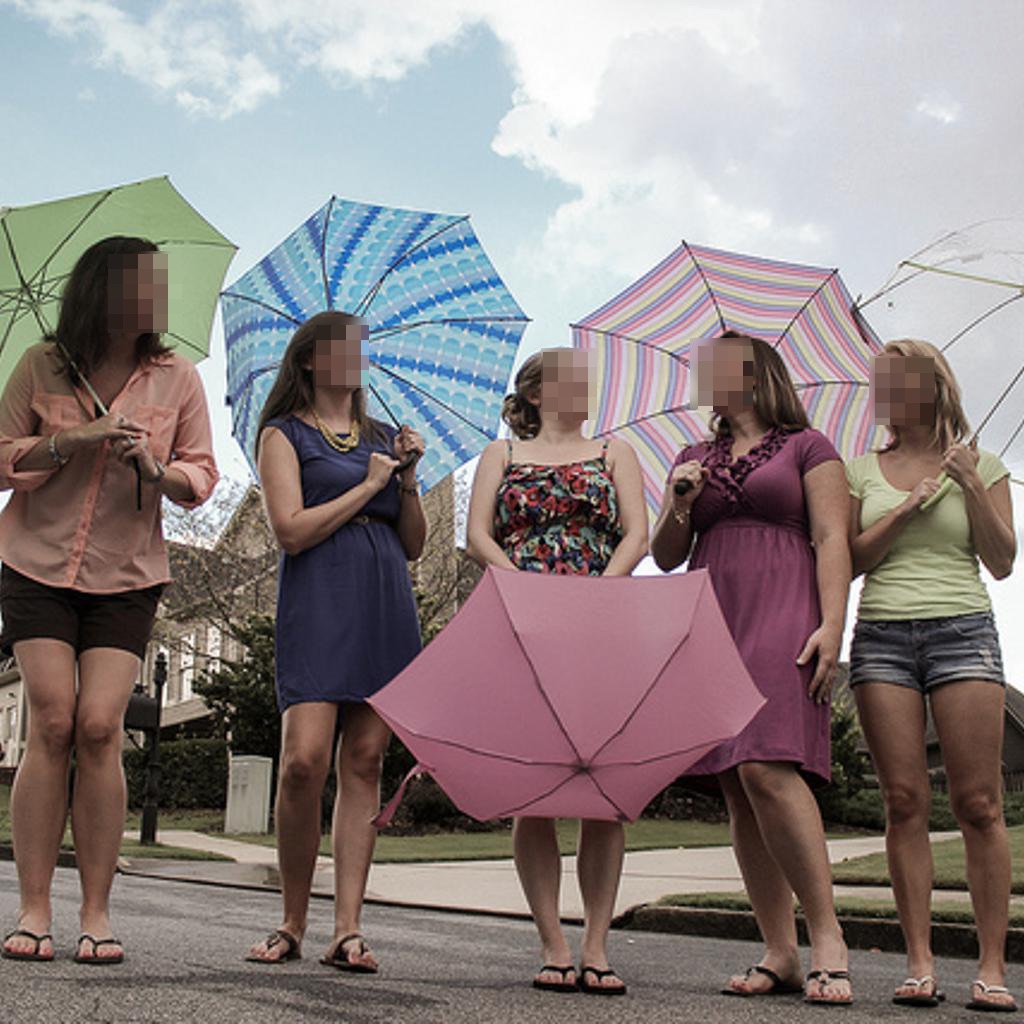} &
          \includegraphics[width=0.33\linewidth]{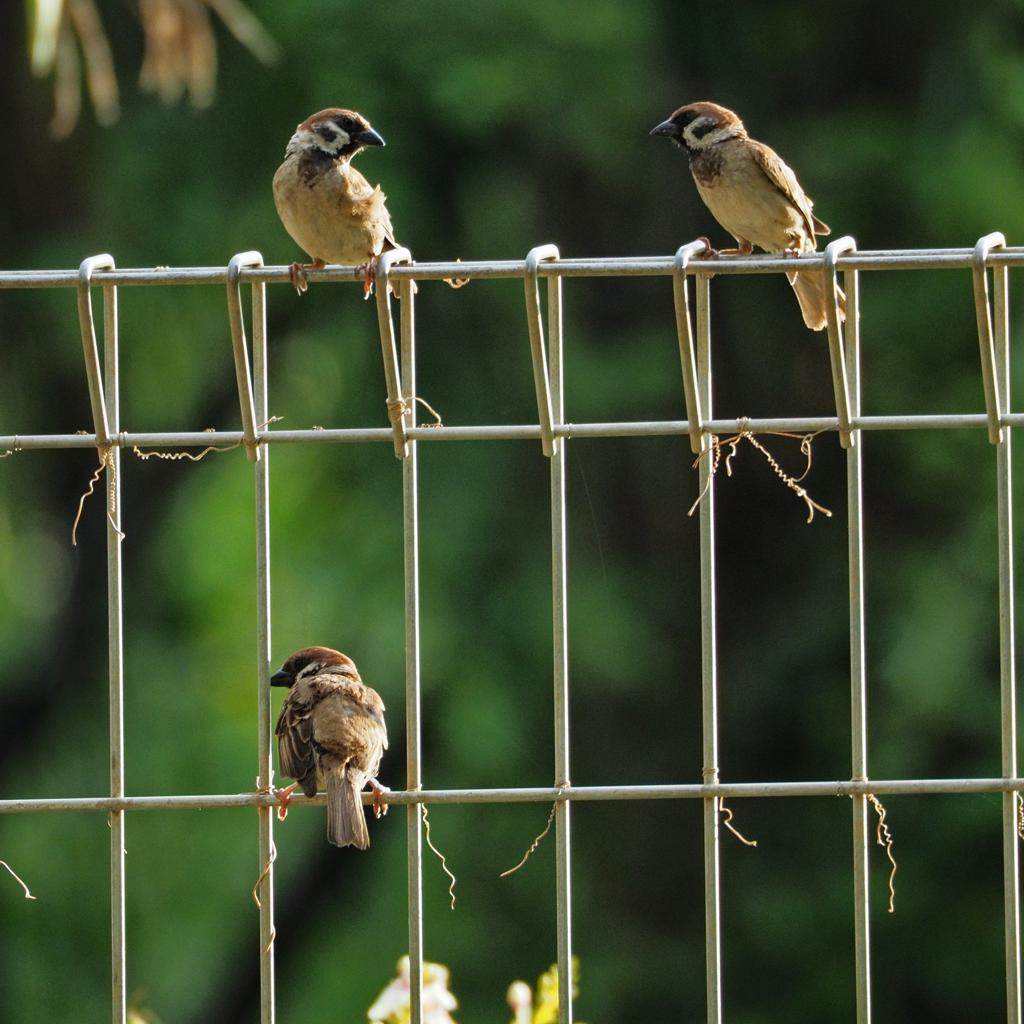} &
          \includegraphics[width=0.33\linewidth]{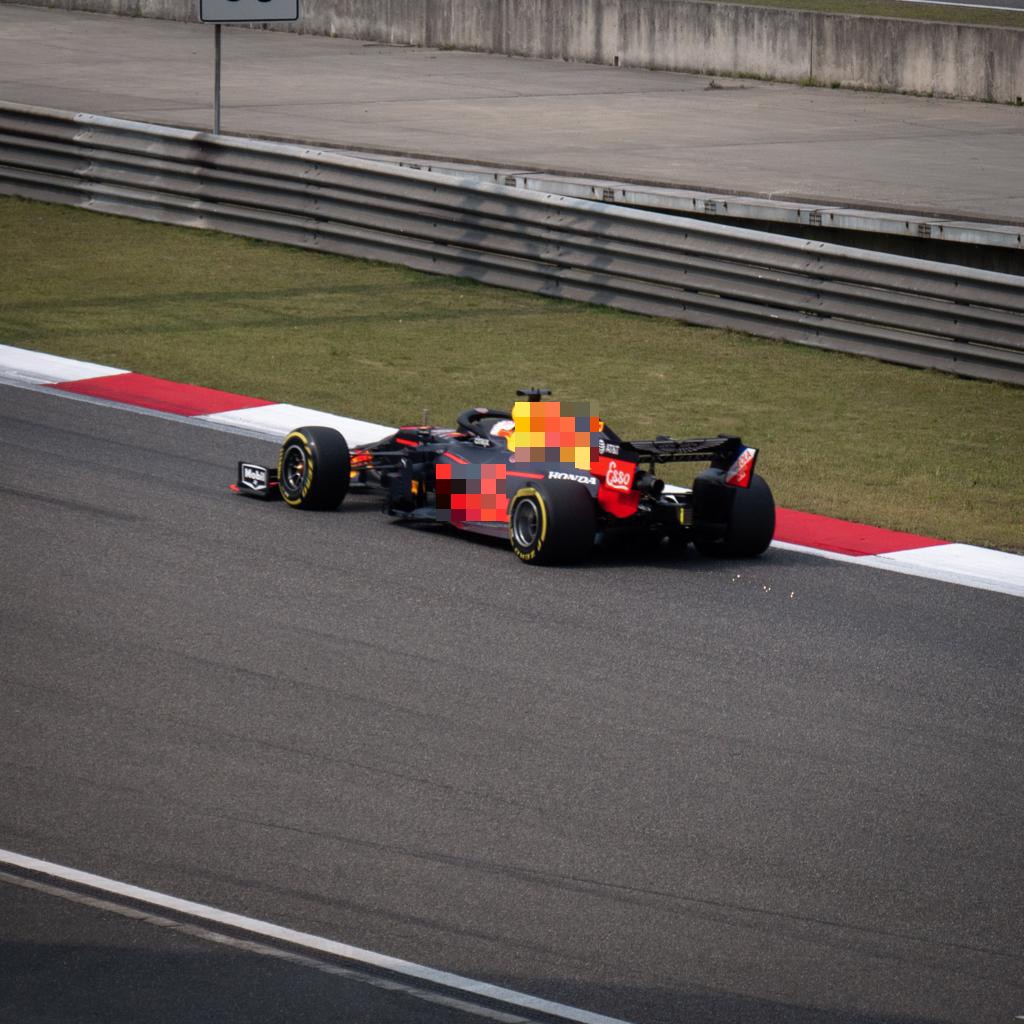} \\ 
          \includegraphics[width=0.33\linewidth]{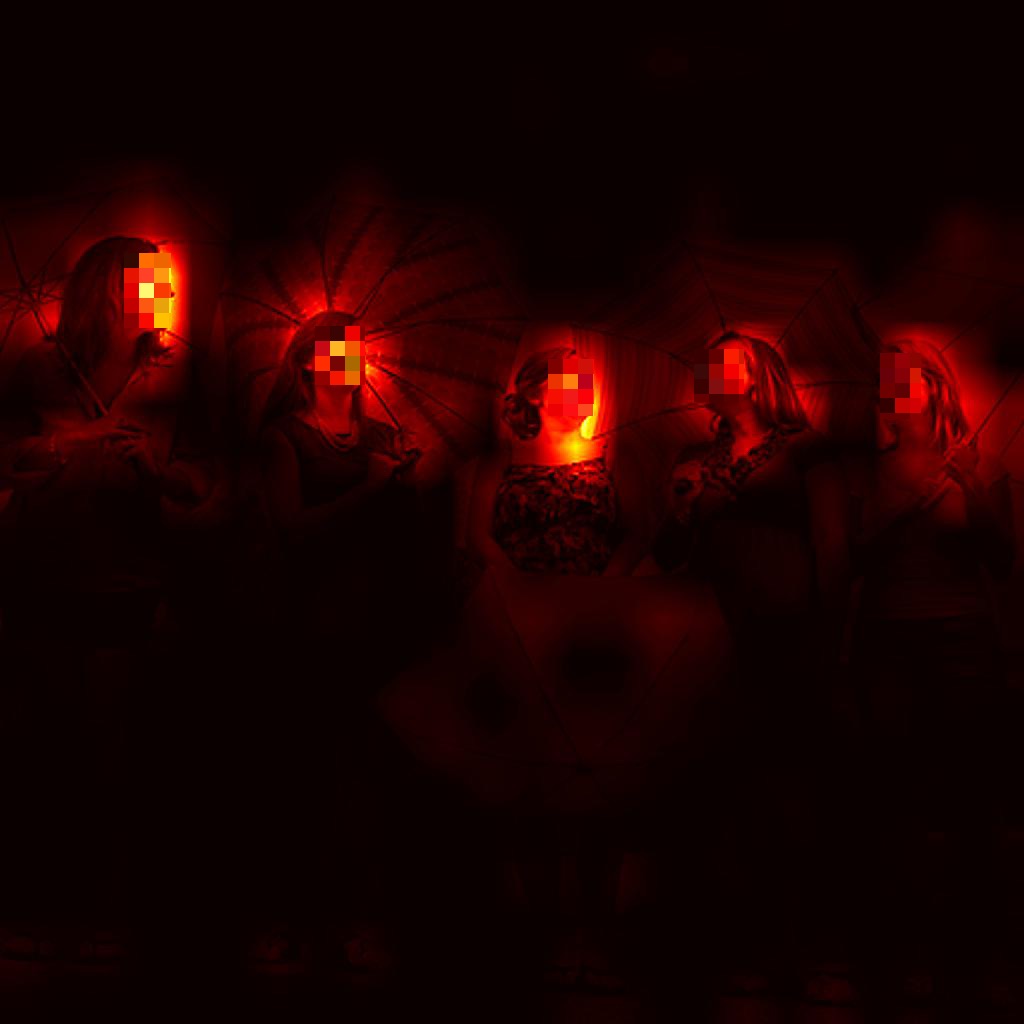} &
          \includegraphics[width=0.33\linewidth]{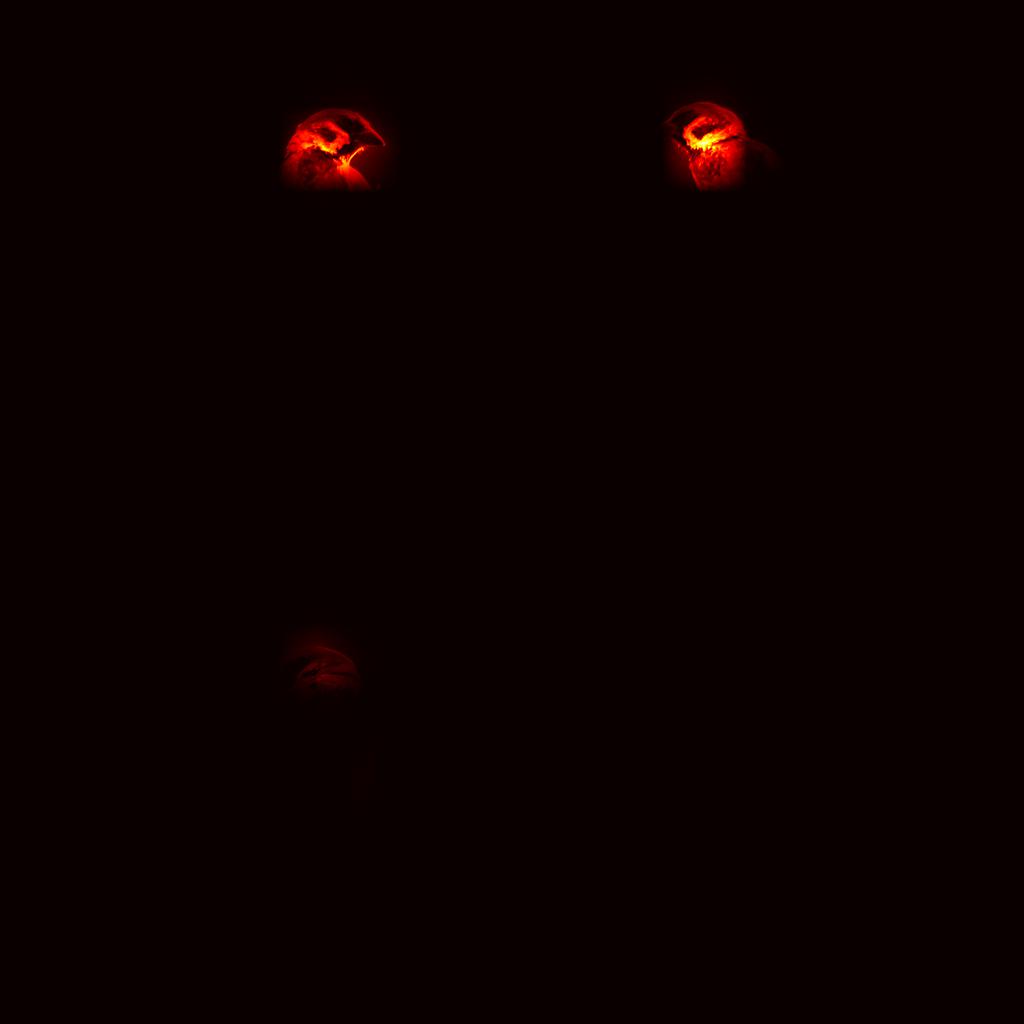} &
          \includegraphics[width=0.33\linewidth]{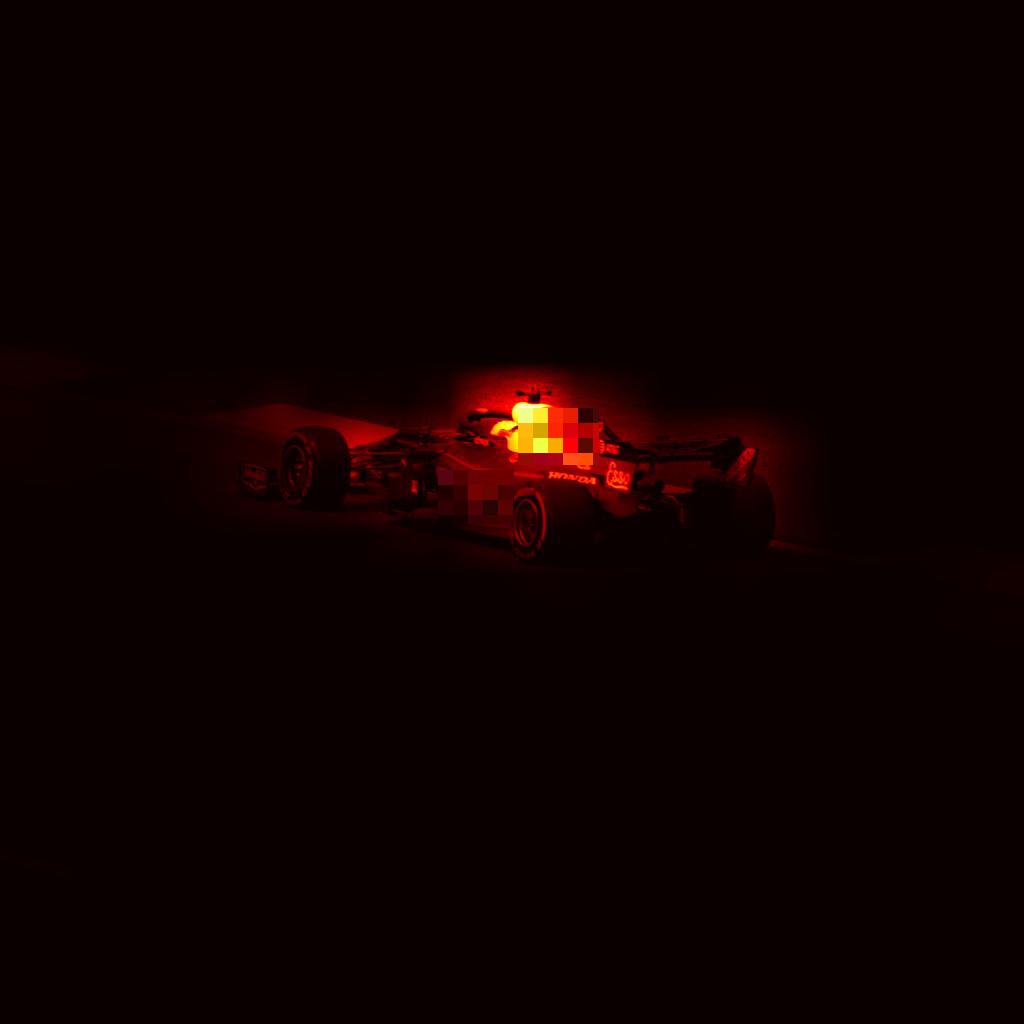} \\
          \includegraphics[width=0.33\linewidth]{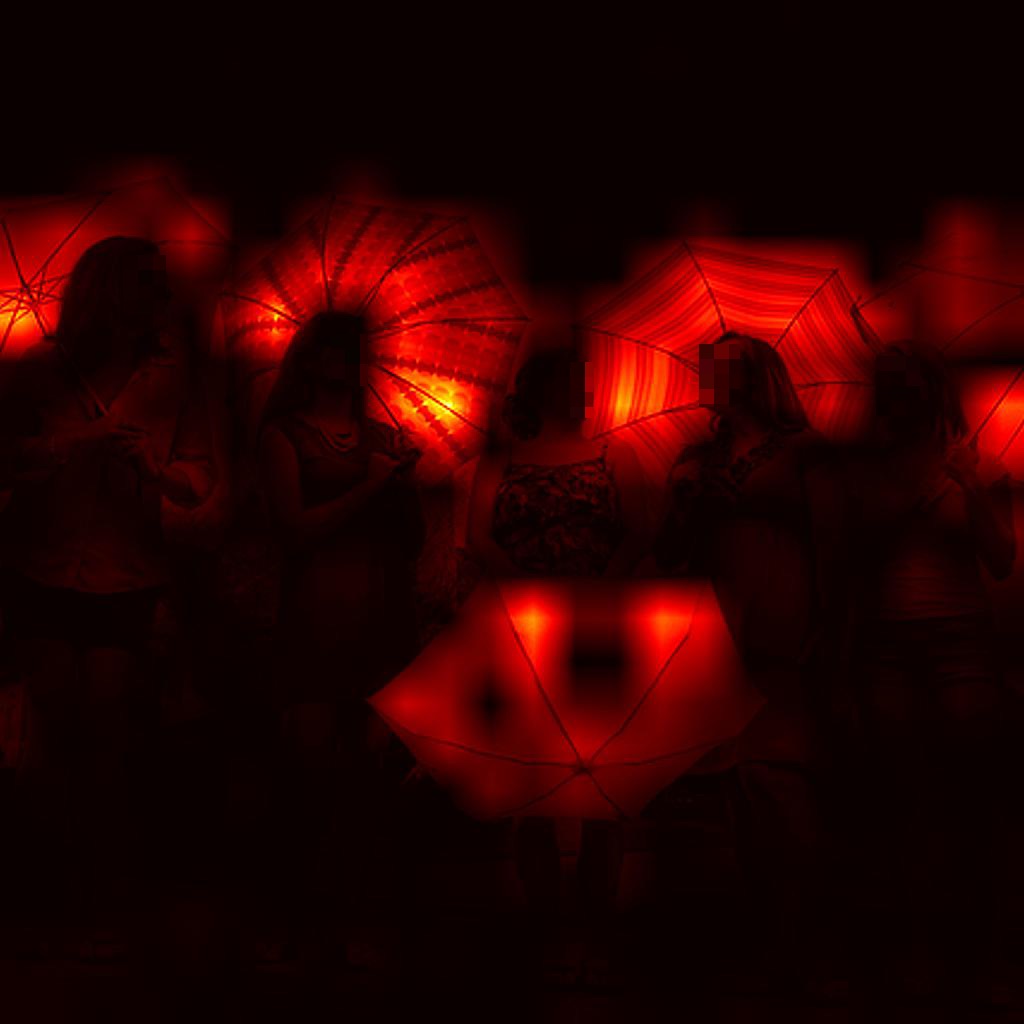} &
          \includegraphics[width=0.33\linewidth]{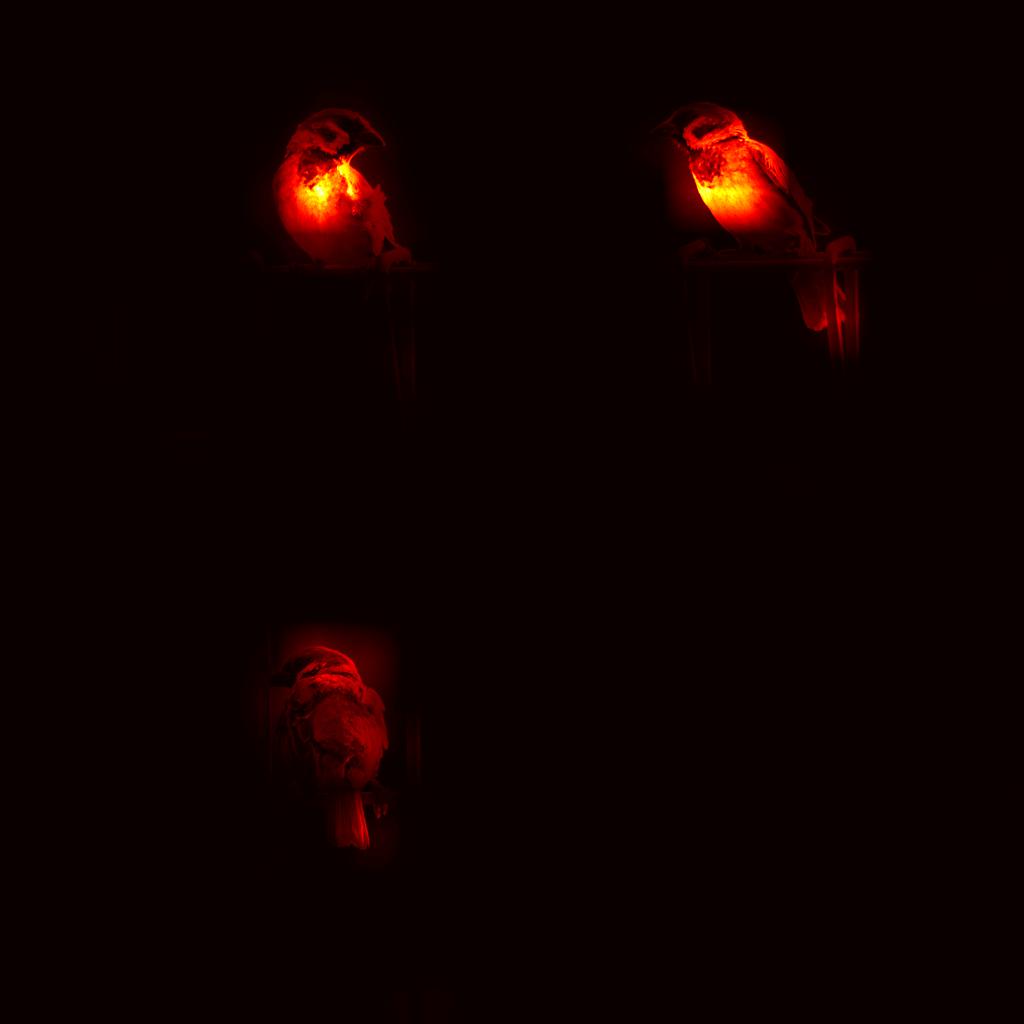} &
          \includegraphics[width=0.33\linewidth]{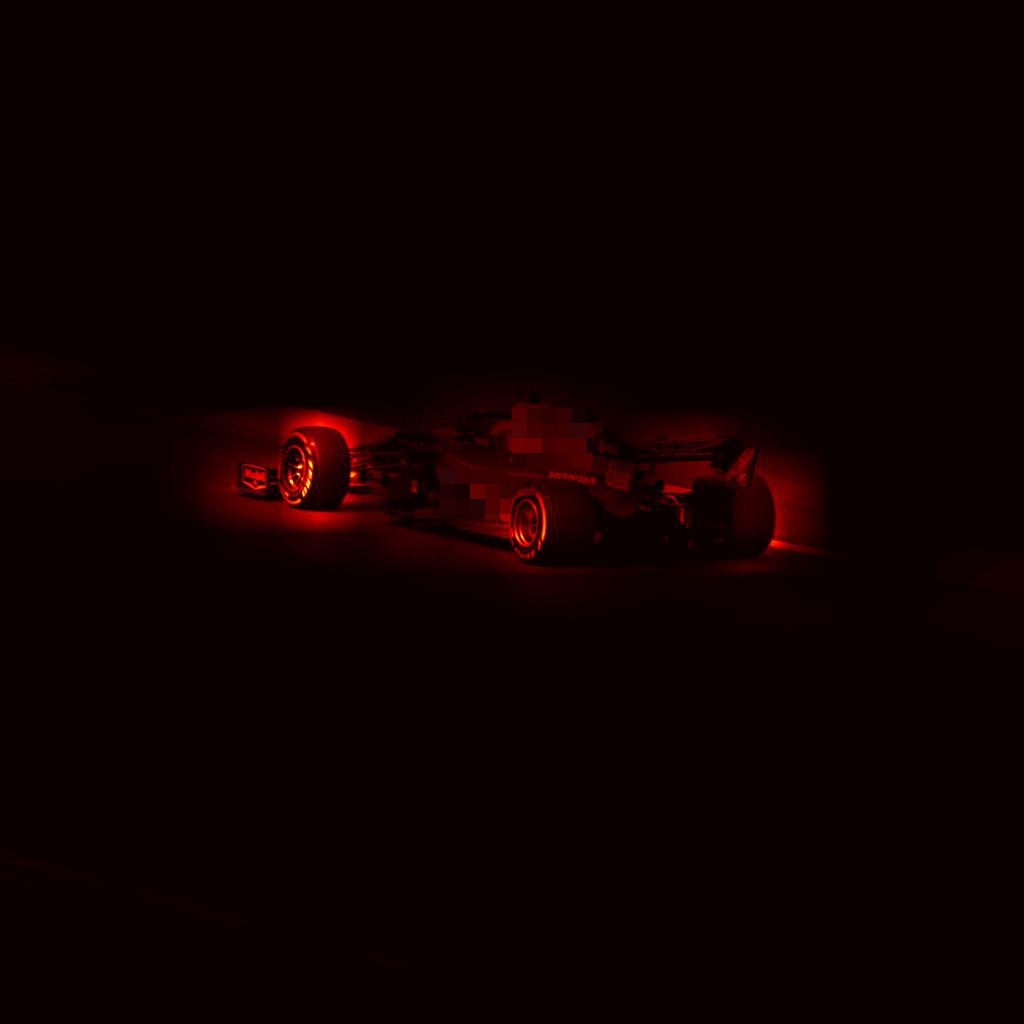} \\
          \includegraphics[width=0.33\linewidth]{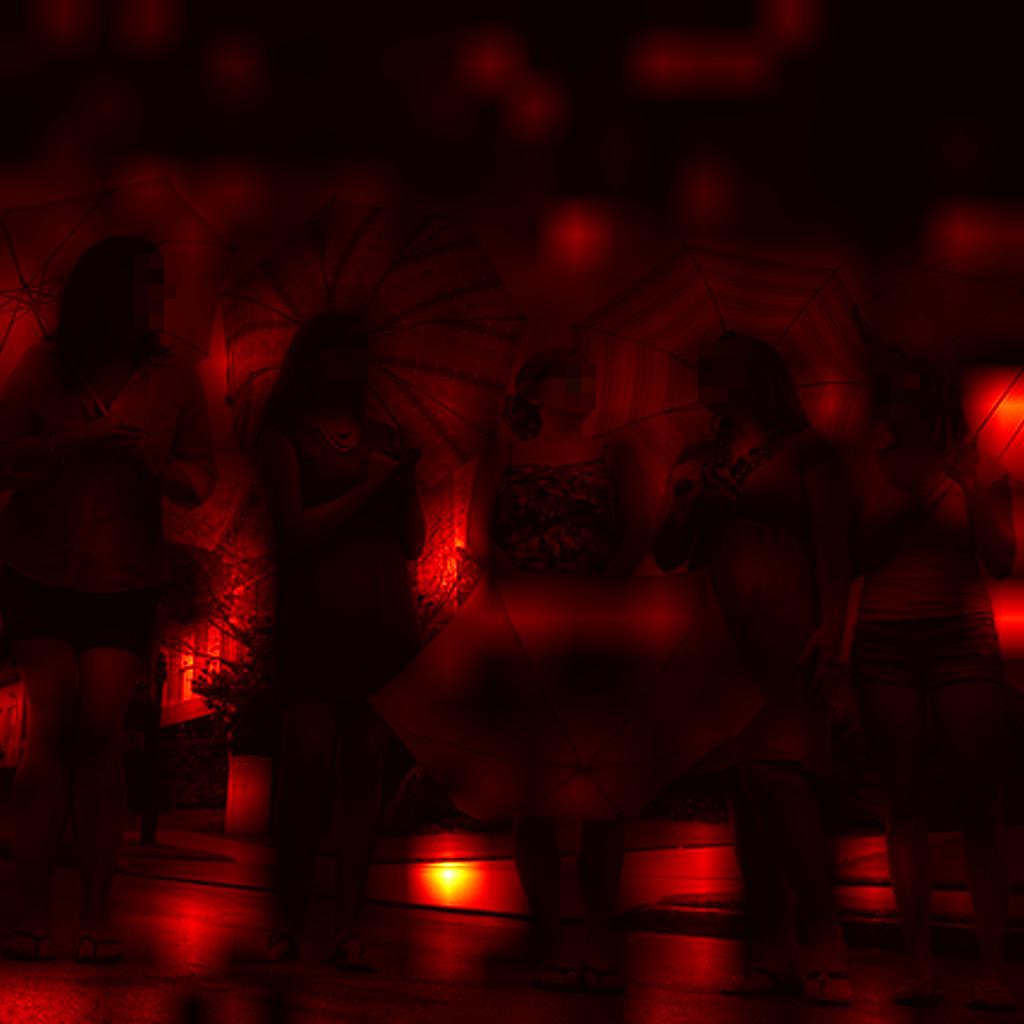} &
          \includegraphics[width=0.33\linewidth]{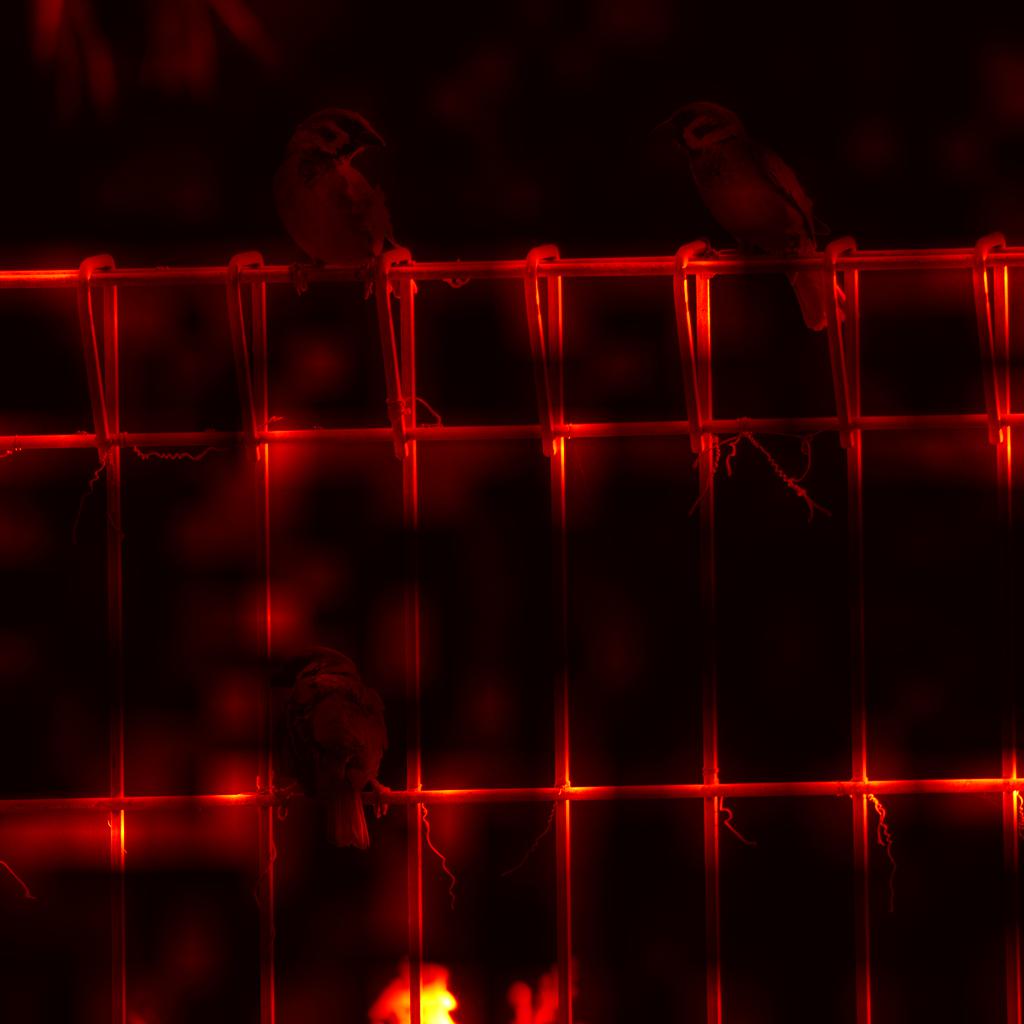} &
          \includegraphics[width=0.33\linewidth]{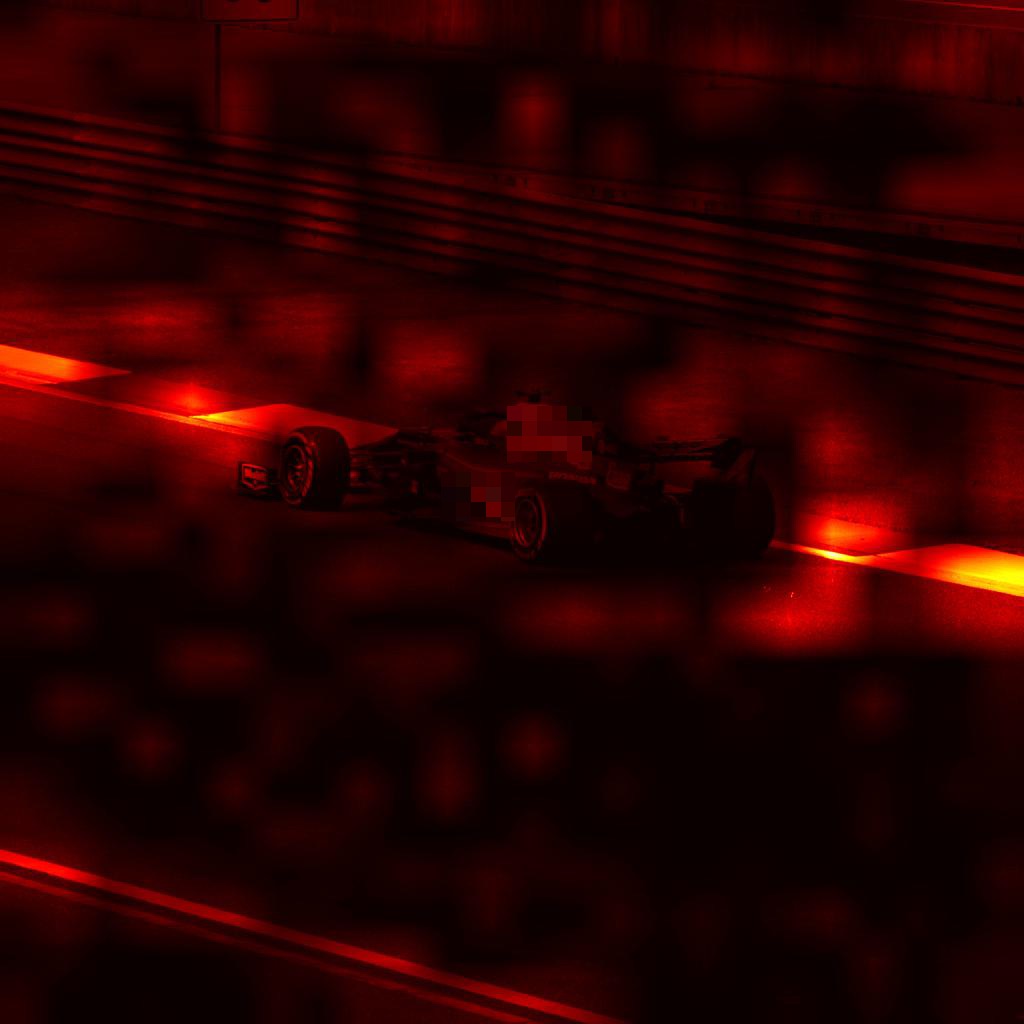} \\
        \end{tabular}
        }\vspace*{-2pt}
    \captionof{figure}{ Visualization of the attention map between the \texttt{CLS} token and  individual patches in the class-attention stage. 
    For each column, each row represents the attention map \wrt one head, corresponding to the image in the first raw. 
    Each head seems salient to semantically coherent regions. Heads are sensitive to similar features within the same or across images (\eg people or bird faces). They trigger on different concepts when such features are missing (\eg, cockpit for race cars).
    }
    \label{fig:cls_gram_vis}
\end{minipage}
\vspace{-5mm}
\end{figure}

In this section we demonstrate the effectiveness and versatility of \ours on multiple computer vision benchmarks, and present ablations providing insight on the importance of its different components. 
In the supplementary material  we provide additional analysis,  including the impact on performance of image resolution in Section  \ref{sec:res_patch} and of multiple approximate attention baselines in Section \ref{sec:approx_att}.

\subsection{Image classification}

We use  \ImNet-1k~\cite{deng2009imagenet} to train and evaluate our models for  image classification.  It consists of 1.28M training images and 50k validation images, labeled across  1,000 semantic categories. Our training setup follows the DeiT recipe~\cite{touvron2020deit}. We train our model for 400 epochs with the AdamW optimizer~\cite{loshchilov2017decoupled} using a cosine learning rate decay. 
In order to enhance the training of larger models, we utilize LayerScale~\cite{touvron2021going} and adjust the stochastic depth~\cite{huang2016deep} for each of our models accordingly (see the supplementary material for details).  %
Following \cite{touvron2021going}, images are cropped with crop ratio of 1.0 for evaluation.  In addition to the \ImNet-1k validation set, we report results for \ImNet-V2~\cite{recht2019imagenet} which has a distinct test set. Our implementation is based on the Timm library~\cite{rw2019timm}.

\paragraph{Results on \ImNet.}
We present a family of seven models in \tab{xct_models} with different operating points  in terms of parameters and FLOPs. We observe that the performance of the \ours models benefits from increased capacity both in depth and width. Additionally, consistent with \cite{touvron2020deit, touvron2021going} we find that using hard distillation with a convolutional teacher improves the performance.
Because of its linear complexity in the number of tokens, it is feasible to train \ours  at $384\!\times\!384$ resolution  with  small  $8\!\times\!8$ patches, \ie 2304 tokens, which  provides a strong boost in performance across all configurations.

We compare to the state-of-the-art convolutional and transformer-based architectures \cite{brock2021high,  liu2021swin, radosavovic2020designing, tan2019efficientnet, touvron2021going} in \tab{classification}. 
By varying the input image resolution and/or patch size, our models provide  competitive or superior performance across  model sizes and FLOP budgets. 
First, the models operating on $224\!\times\!224$ and $16\!\times\!16$ (\eg \ours-S12/16) enjoy high accuracy at relatively few FLOPs compared to their counterparts with comparable parameter count and FLOPs. 
Second, our models with $16\!\times\!16$ and  $384\!\times\!384$ resolution images  (\eg \ours-S12/16$\uparrow$) yield an improved accuracy at the expense of higher FLOPs, and  provide superior or on-par performance compared to state-of-the-art models with comparable computational requirements. 
Finally, \ours linear complexity allows us to scale to process $384\!\times\!384$ images with $8\times8$ patch sizes (\eg \ours-S12/8$\uparrow$), achieving the highest accuracy across the board, albeit at a relatively high FLOPs count.

\paragraph{Class attention visualization.} In \fig{cls_gram_vis} we show  the class attention map obtained in the feature aggregation stage. %
Each head focuses on different semantically coherent regions in the image (\eg faces or umbrellas). Furthermore,  heads tend to focus on similar patterns across images (\eg bird head or human face), but adapts by focusing on other salient regions when such patterns are absent.

\paragraph{Robustness to resolution changes.}
In \fig{change_res_testtime} we report the accuracy of \ours-S12, DeiT-S and ResNet-50    trained on 224$\times$224 images and evaluated at different image resolutions.
While DeiT outperforms ResNet-50 when train and test resolutions are similar, it suffers from a larger drop in performance as the image resolution deviates farther from the training resolution.
\ours displays a substantially increased accuracy when train and test resolutions are similar, while also being robust to resolution changes, in particular for the model with $8\!\times\!8$ patches.

\def \myminus {\textcolor{red!80}{\scalebox{0.7}{\fbox{\small \bf $-$}}\ \xspace}}
\def \myplus {\textcolor{blue!80}{\scalebox{0.7}{\fbox{\small \bf $+$}}\ \xspace}}

\begin{table}[t!]
    \begin{minipage}[t]{0.52\textwidth}
        \caption{\textbf{Self-supervised learning.} Top-1 acc. on \ImNet-1k.  Wwe report with a crop-ratio 0.875 for consistency with DINO. For the last row it is set to 1.0 (improves from 80.7\% to 80.9\%). All models are trained for 300 epochs.\vspace*{2pt} %
        \label{tab:ssl}}
        \def \mysp {\hspace{6pt}}
        \centering \scalebox{0.65}
        {
        \begin{tabular}{@{\ }ll@{\ \ }r@{\ \ \ }r@{\mysp}cc}
        \toprule
        SSL Method & Model  & \#params &  FLOPs & Linear & $k$-NN \\
        \toprule
        MoBY~\cite{xie2021self} & Swin-T \cite{liu2021swin} & 29M & \dzo4.5B & 75.0 & -- \\
        DINO~\cite{caron2021emerging} & ResNet-50 \cite{he2016deep} & 23M & \dzo4.1B & 74.5 & 65.6 \\
        DINO~\cite{caron2021emerging} & ViT-S/16 \cite{dosovitskiy2020image} & 22M & \dzo4.6B & 76.1 & 72.8 \\
        DINO~\cite{caron2021emerging} & ViT-S/8 \cite{dosovitskiy2020image} & 22M & \dzo22.4B & \textbf{79.2} & \textbf{77.2} \\
        \rowcolor{Goldenrod}
        DINO~\cite{caron2021emerging} & \ours-S12/16 & 26M & \dzo4.9B & 77.8 & 76.0 \\
        \rowcolor{Goldenrod}
        DINO~\cite{caron2021emerging} &  \ours-S12/8 & 26M & \dzo18.9B & \textbf{79.2} & 77.1 \\
        \midrule
        DINO~\cite{caron2021emerging} & ViT-B/16 \cite{dosovitskiy2020image} & 87M & \dzo17.5B & 78.2 & 76.1 \\
        DINO~\cite{caron2021emerging} & ViT-B/8 \cite{dosovitskiy2020image} & 87M & \dzo78.2B & 80.1 & 77.4 \\
        \rowcolor{Goldenrod}
        DINO~\cite{caron2021emerging} & \ours-M24/16 & 84M & \dzo16.2B & 78.8 & 76.4 \\
        \rowcolor{Goldenrod}
        DINO~\cite{caron2021emerging} & \ours-M24/8 & 84M & \dzo64.0B & \textbf{80.3} & \textbf{77.9} \\
        \rowcolor{Goldenrod}
        DINO~\cite{caron2021emerging} & \ours-M24/8$\uparrow$384 & 84M & \dzo188.0B & \textbf{80.9} & - \\

        \bottomrule
        \end{tabular}}  
    \strut\end{minipage}%
    \hfill %
    \begin{minipage}[t]{0.45\textwidth}
        \vspace{0pt}
        \centering
        \caption{\textbf{Ablations} of various architectural design choices on the task of \ImNet-1k classification using the \ours-S12 model. Our baseline model uses the convolutional projection adopted from LeVit. 
        }%
\vspace{1.5pt}
\scalebox{0.75}{
\centering
\begin{tabular}{llc}
\toprule
\multirow{1}{*}{Model} & \multirow{1}{*}{Ablation} & \!\! ImNet top-1 acc.\\
\midrule
\OURS-S12/16 & \multirow{2}{*}{Baseline} & 82.0 \\
\OURS-S12/8  &    & 83.4 \\
\midrule
\OURS-S12/16 & \multirow{2}{*}{Linear patch proj.} & 81.1 \\
\OURS-S12/8  &   &  83.1 \\
\midrule
\multirow{2}{*}{\OURS-S12/16} 
    &  w/o  LPI layer &  80.8 \\
&  w/o  XCA layer &  75.9 \\%[2pt]

\midrule
\multirow{2}{*}{\OURS-S12/16} &  w/o $\ell_2$-normal. & failed \\ 
    & w/o learned temp.\ $\tau$ \!\!\!\!\!\!  & 81.8 \\ 
\bottomrule
\end{tabular}
        }
        \label{tab:ablations}
    \end{minipage}
    \vspace{-6mm}
\end{table}

\paragraph{Self-supervised learning.}
We train \ours in a self-supervised manner using DINO~\cite{caron2021emerging} 
on \ImNet-1k.
In \tab{ssl} we report  performance using the linear and $k$-NN protocols as in~\cite{caron2021emerging}.
Across model sizes \ours obtains excellent accuracy with both protocols, substantially improving DINO with  ResNet-50 or ViT architectures,  as well as over those reported for Swin-Transformer trained with MoBY \cite{xie2021self}.
Comparing the larger models to ViT, we also observed improved performance for \ours achieving a strong 80.3\% accuracy.
For fair comparison, all reported models  have been trained for 300 epochs. Further improved performance of small models is reported by \citet{caron2021emerging} when  training for 800 epochs, which we expect to carryover to \ours based on the results presented here.

\paragraph{Analysis and ablations.}
In \tab{ablations} we provide  ablation experiments  to  analyse the impact of different design choices for our \OURS-S12  model. 
First, we  observe the positive effect of using the convolutional patch projection as compared to using linear patch projection, for both $8\!\times\!8$ and $16\!\times\!16$ patches.
Second, while removing the LPI layer reduces the accuracy by only 1.2\% (from 82.0 to 80.8), removing the XCA layer results in a large drop of 6.1\%,  underlining the effectiveness of  XCA.
We  noticed that the inclusion of two convolutional components -- convolutional patch projection and LPI -- not only brings improvements in accuracy, but also accelerates  training.
Third, although we were able to ensure proper convergence without $\ell_2$-normalization of  queries and keys by tweaking the hyper-parameters, 
we found that it provides stability across model size (depth and width) and other hyper-parameters.
Finally, while the learnable softmax temperature parameter is not critical, removing it drops accuracy by 0.2\%. %
Additional ablations are provided in the supplementary material.

  \begin{table}[t]

    \begin{minipage}[t]{.52\linewidth}
        \captionof{table}{\footnotesize \textbf{COCO object detection and instance segmentation} performance on the mini-val set. All backbones are pre-trained on ImageNet-1k, use Mask R-CNN model~\cite{he2017mask} and are trained with the same 3x schedule.}\vspace*{2pt}
        \centering
        \scalebox{0.675}{
        \begin{tabular}{l@{}|c@{\ \ } | c@{\ \ \ }c@{\ \ }c|c@{\ \ }c@{\ \ } c}
        \toprule
             Backbone & \!\!\!\#params\!\!\! & $\text{AP}^{b}$ & $\text{AP}^{b}_{50}$ & $\text{AP}^{b}_{75}$ & $\text{AP}^{m}$ & $\text{AP}^{m}_{50}$ & $\text{AP}^{m}_{75}$\\ 
            \midrule 
            ResNet18 \cite{he2016deep} & 31.2M & 36.9 & 57.1 & 40.0 & 33.6 & 53.9 & 35.7 \\
            PVT-Tiny \cite{wang2021pyramid} & 32.9M & 39.8 & 62.2 & 43.0 & 37.4 & 59.3 & 39.9 \\
            ViL-Tiny \cite{zhang2021multi} & 26.9M & 41.2 & 64.0 & 44.7 & 37.9 & 59.8 & 40.6 \\
            \rowcolor{Goldenrod}
            \OURS-T12/16 & 26.1M & 42.7 & 64.3 & 46.4 & 38.5 & 61.2 & 41.1 \\
            \rowcolor{Goldenrod}
            \OURS-T12/8  & 25.8M & \textbf{44.5} & \textbf{66.4} & \textbf{48.8} & \textbf{40.3} & \textbf{63.5} & \textbf{43.2} \\
            \midrule
            ResNet50 \cite{he2016deep} & 44.2M & 41.0 & 61.7 & 44.9 & 37.1 & 58.4 & 40.1 \\
            PVT-Small \cite{wang2021pyramid} & 44.1M & 43.0 & 65.3 & 46.9 & 39.9 & 62.5 & 42.8 \\
            ViL-Small \cite{zhang2021multi} & 45.0M & 43.4 & 64.9 & 47.0 & 39.6 & 62.1 & 42.4 \\
            Swin-T \cite{liu2021swin} & 47.8M & 46.0 & 68.1  & 50.3 & 41.6 & 65.1 & 44.9 \\
            \rowcolor{Goldenrod}
            \OURS-S12/16 & 44.3M & 45.3 & 67.0 & 49.5 & 40.8 & 64.0 & 43.8  \\
            \rowcolor{Goldenrod}
            \OURS-S12/8  & 43.1M & \textbf{47.0} & \textbf{68.9}  & \textbf{51.7}  & \textbf{42.3}  & \textbf{66.0} & \textbf{45.4}  \\
            \midrule
            ResNet101 \cite{he2016deep} & 63.2M & 42.8 & 63.2 & 47.1 & 38.5 & 60.1 & 41.3 \\
            ResNeXt101-32 & 62.8M & 44.0 & 64.4 & 48.0 & 39.2 & 61.4 & 41.9 \\
            PVT-Medium \cite{wang2021pyramid} & 63.9M & 44.2 & 66.0 & 48.2 & 40.5 & 63.1 & 43.5 \\
            ViL-Medium \cite{zhang2021multi} & 60.1M & 44.6 & 66.3 & 48.5 & 40.7 & 63.8 & 43.7 \\
            Swin-S \cite{liu2021swin} & 69.1M & \textbf{48.5} & \textbf{70.2} &\textbf{53.5} & \textbf{43.3} &\textbf{ 67.3} & \textbf{46.6} \\
            \rowcolor{Goldenrod}
            \OURS-S24/16 & 65.8M  & 46.5 & 68.0 & 50.9 & 41.8 & 65.2  & 45.0 \\
            \rowcolor{Goldenrod}
            \OURS-S24/8  & 64.5M & 48.1  & 69.5  & 53.0 & 43.0 & 66.5 & 46.1 \\
            \midrule
            ResNeXt101-64 \cite{xie2017aggregated} & \!101.9M\! & 44.4 & 64.9 & 48.8 & 39.7 & 61.9 & 42.6 \\
            PVT-Large \cite{wang2021pyramid} & 81.0M & 44.5 & 66.0 & 48.3 & 40.7 & 63.4 & 43.7 \\
            ViL-Large \cite{zhang2021multi}  & 76.1M & 45.7 & 67.2 & 49.9 & 41.3 & 64.4 & 44.5 \\
            \rowcolor{Goldenrod}
            \OURS-M24/16 & 101.1M & 46.7 & 68.2 & 51.1 & 42.0 & 65.6 & 44.9  \\
            \rowcolor{Goldenrod}
            \OURS-M24/8  & 98.9M & \textbf{48.5} & \textbf{70.3} & \textbf{53.4}  & \textbf{43.7} & \textbf{67.5} & \textbf{46.9}  \\
        \bottomrule     
     \end{tabular}     
     } %
     \label{tab:coco_det}
     \end{minipage}
\hfill
    \begin{minipage}[t]{.45\linewidth}
        \caption{\textbf{ADE20k semantic segmentation} performance using Semantic FPN \cite{kirillov2019panoptic} and UperNet \cite{xiao2018unified} (in comparable settings). We do not include comparisons with other state-of-the-art models that  are pre-trained on larger  datasets~\cite{liu2021swin, ranftl2021vision,zheng2020rethinking}.\vspace*{4pt} \label{tab:sem_seg}}
        \centering
        \scalebox{0.676}{
        \begin{tabular}{l |c|c|c|c}
        \toprule
             Backbone & \multicolumn{2}{c|}{Semantic FPN} & \multicolumn{2}{c}{UperNet}  \\
             \cmidrule{2-3}
             \cmidrule{4-5}
             & \#params & mIoU & \#params & mIoU \\
            \midrule 
             ResNet18 \cite{he2016deep} &  15.5M & 32.9 & - & - \\
             PVT-Tiny \cite{wang2021pyramid}  &  17.0M & 35.7M & - & -\\
             \rowcolor{Goldenrod}
             \OURS-T12/16 & 8.4M & 38.1 & 33.7M & 41.5 \\
             \rowcolor{Goldenrod}
             \OURS-T12/8 & 8.4M & \textbf{39.9} & 33.7 &\textbf{43.5}  \\
             \midrule
             ResNet50 \cite{he2016deep} & 28.5M & 36.7  & 66.5M & 42.0 \\
             PVT-Small \cite{wang2021pyramid}  &  28.2M & 39.8 & - & - \\
             Swin-T  \cite{liu2021swin} & - & - & 59.9M & 44.5 \\
             \rowcolor{Goldenrod}
             \OURS-S12/16 & 30.4M & 43.9 & 52.4M & 45.9 \\
             \rowcolor{Goldenrod}
             \OURS-S12/8 & 30.4M & \textbf{44.2} & 52.3M &  \textbf{46.6} \\
            \midrule
             ResNet101 \cite{he2016deep} & 47.5M & 38.8  & 85.5M & 43.8 \\
             ResNeXt101-32 \cite{xie2017aggregated}\! & 47.1M & 39.7 & - & - \\
             PVT-Medium \cite{wang2021pyramid} &  48.0M & 41.6 & - & - \\
             Swin-S \cite{liu2021swin} & - & - & 81.0M & 47.6 \\
             \rowcolor{Goldenrod}
             \OURS-S24/16  & 51.8M & 44.6  & 73.8M & 46.9 \\
             \rowcolor{Goldenrod}
             \OURS-S24/8  & 51.8M & \textbf{47.1} & 73.8M & \textbf{48.1} \\
             \midrule
             ResNeXt101-64 \cite{xie2017aggregated}\! & 86.4M & 40.2 & - & - \\
             PVT-Large \cite{wang2021pyramid} &  65.1M & 42.1 & - & - \\
             Swin-B \cite{liu2021swin} & - & - & 121.0M & 48.1 \\
             \rowcolor{Goldenrod}
             \OURS-M24/16 & 90.8M & 45.9 & 109.0M & 47.6 \\
              \rowcolor{Goldenrod}
             \OURS-M24/8 & 90.8M & \textbf{46.9} & 108.9M & \textbf{48.4} \\
        \bottomrule     
        \end{tabular}   
        } %
  \end{minipage}
  \vspace{-5pt}
\end{table}  

\subsection{Object detection and instance segmentation}
\label{sec:coco_imp_det}

Our \ours models can efficiently process high-resolution images (see Figure~\ref{fig:peak_mem}). Additionally, \ours has a better adaptability to varying image resolutions compared to ViT models (see Figure~\ref{fig:change_res_testtime}). These two properties make \ours a good fit for dense prediction tasks including detection and segmentation. 

We evalutate \ours  for object detection and instance segmentation using the COCO benchmark \cite{lin2014microsoft} which consists of 118k training and 5k validation images including bounding boxes and mask labels for 80 categories.
We integrate  \ours as backbone in the Mask R-CNN~\cite{he2017mask} detector with FPN~\cite{lin2017feature}.  
Since the \ours architecture is inherently columnar, we make it FPN-compatible by extracting features from different layers (\eg, [4, 6, 8, 12] for \ours-S12). 
All features have a constant stride of 8 or 16 based on the patch size, and the feature resolutions are adjusted to have strides of [4, 8, 16, 32], similar to ResNet-FPN backbones, where the downsampling is achieved by max pooling and the upsampling is obtained using a single transposed convolution layer (see %
suppl. mat. for details). The model is trained for 36 epochs (3x schedule) using the AdamW optimizer with learning rate of $10^{-4}$, 0.05 weight decay and 16 batch size. We adopt the multiscale training and augmentation strategy of DETR~\cite{carion2020end}. Our implementation is based on the mmdetection library~\cite{mmdetection}.

\paragraph{Results on COCO.}
In Table~\ref{tab:coco_det} we report  object detection and instance segmentation results of four variants of \ours  using $16\!\times\!16$ and $8\!\times\!8$ patches. We compare to ResNets~\cite{he2016deep} and concurrent efficient vision transformers \cite{liu2021swin,wang2021pyramid,zhang2021multi}. All models are trained using the 3x schedule after \ImNet-1k pre-training. Note that other results with higher absolute numbers have been achieved when pre-training on larger datasets~\cite{liu2021swin} or with longer schedules \cite{bello2021lambdanetworks}, and are therefore not directly comparable to the reported results. 
First, across all model sizes \ours outperforms the convolutional ResNet~\cite{he2016deep} and ResNeXt~\cite{xie2017aggregated} by a large margin with either patch size. Second, we observe a similar increase in accuracy compared to  PVT~\cite{wang2021pyramid} and ViL~\cite{zhang2021multi}  backbones. Finally, \ours provides a competitive performance with Swin \cite{liu2021swin} \footnote{We use report the results  provided by the authors in their open-sourced code  \url{https://github.com/SwinTransformer/Swin-Transformer-Object-Detection}}. For relatively small models, \ours-S12/8 outperforms its Swin-T counterpart with a decent margin. On the other hand, Swin-S provides slightly stronger results compared to \ours-S24/8. %
Utilizing smaller 8$\times$8 patches leads to a consistent gain across all models.

\subsection{Semantic segmentation}

We further show transferability of our models with semantic segmentation experiments on the ADE20k dataset \cite{zhou2017scene}, which consists of 20k training and 5k validation images with labels over 150 semantic categories. 
We integrate our backbones in two segmentation methods: Semantic FPN \cite{kirillov2019panoptic} and UperNet \cite{xiao2018unified}. We train for 80k and 160k iterations for Semantic FPN and UperNet respectively. Following \cite{liu2021swin}, the models are trained using  batch size 16 and an AdamW optimizer with learning rate of $6\times10^{-5}$  and 0.01 weight decay. We apply the same method of extracting FPN features as explained in Section~\ref{sec:coco_imp_det}. %
We report the performance using the standard single scale protocol (without multi-scale and flipping). Our implementation is based on the mmsegmentation library \cite{mmseg2020}.

\paragraph{Results on ADE20k.}
We present the semantic segmentation performance using \ours backbones %
in Table~\ref{tab:sem_seg}. First, for Semantic FPN \cite{kirillov2019panoptic}, \ours provides a superior performance compared to ResNet, ResNeXt and PVT backbones using either option of patch size. Second, compared to Swin Transformers using the same UperNet decoder \cite{xiao2018unified}, \ours with 8$\times$8 patches consistently achieves a higher mIoU for different models. \ours with 16$\times$16 patches provides a strong performance especially for smaller models where \ours-S12/16 outperforms Swin-T.

\section{Conclusion}
\label{sec:conclusion}

\paragraph{Contributions.} We present an alternative to token self-attention which operates on the feature dimension, eliminating the need for expensive computation of quadratic attention maps. We build our \ours %
models with the cross-covariance attention as its core component and demonstrate the effectiveness and generality of our models on various computer vision tasks. In particular, it exhibits a strong image classification performance on par with state-of-the-art transformer models while similarly robust to changing image resolutions as convnets. %
\ours  is effective as a backbone for %
dense prediction tasks, providing excellent performance on object detection, instance and semantic segmentation. Finally, we showed that \ours can be a strong backbone for self-supervised learning, matching the state-of-the-art results with less compute. %
\ours is a generic architecture that can readily be deployed in %
other research domains where self-attention has shown success.

\ifarxiv

\else
\paragraph{Limitations.} Our models enable training with smaller patches and on higher-resolution images, which leads to clear performance gains. However, for tasks like image classification this gain comes at a cost of relatively high number of FLOPs. In order to address this issue, other components, like FFN, could also be re-examined. %
Another point is that \OURS models seem to overfit more than their CaiT counterparts, see Table~\ref{tab:classification}. They are more similar to some convnets in that respect.  
\fi

\ifarxiv 
{\small
\bibliographystyle{plainnat}
\bibliography{egbib}
}

\else
\clearpage
{\small
\bibliographystyle{plainnat}
\bibliography{egbib}

\begin{thebibliography}{84}
\providecommand{\natexlab}[1]{#1}
\providecommand{\url}[1]{\texttt{#1}}
\expandafter\ifx\csname urlstyle\endcsname\relax
  \providecommand{\doi}[1]{doi: #1}\else
  \providecommand{\doi}{doi: \begingroup \urlstyle{rm}\Url}\fi

\bibitem[Ainslie et~al.(2020)Ainslie, Ontanon, Alberti, Cvicek, Fisher, Pham,
  Ravula, Sanghai, Wang, and Yang]{ainslie2020etc}
Joshua Ainslie, Santiago Ontanon, Chris Alberti, Vaclav Cvicek, Zachary Fisher,
  Philip Pham, Anirudh Ravula, Sumit Sanghai, Qifan Wang, and Li~Yang.
\newblock Etc: Encoding long and structured inputs in transformers.
\newblock In \emph{Conference on Empirical Methods in Natural Language
  Processing}, 2020.

\bibitem[Arnab et~al.(2021)Arnab, Dehghani, Heigold, Sun, Lu{\v{c}}i{\'c}, and
  Schmid]{arnab2021vivit}
Anurag Arnab, Mostafa Dehghani, Georg Heigold, Chen Sun, Mario Lu{\v{c}}i{\'c},
  and Cordelia Schmid.
\newblock Vivit: A video vision transformer.
\newblock \emph{arXiv preprint arXiv:2103.15691}, 2021.

\bibitem[Ba et~al.(2016)Ba, Kiros, and Hinton]{ba2016layer}
Jimmy~Lei Ba, Jamie~Ryan Kiros, and Geoffrey~E Hinton.
\newblock Layer normalization.
\newblock \emph{arXiv preprint arXiv:1607.06450}, 2016.

\bibitem[Bello(2021)]{bello2021lambdanetworks}
Irwan Bello.
\newblock {LambdaNetworks}: Modeling long-range interactions without attention.
\newblock \emph{arXiv preprint arXiv:2102.08602}, 2021.

\bibitem[Beltagy et~al.(2020)Beltagy, Peters, and Cohan]{beltagy2020longformer}
Iz~Beltagy, Matthew~E Peters, and Arman Cohan.
\newblock Longformer: The long-document transformer.
\newblock \emph{arXiv preprint arXiv:2004.05150}, 2020.

\bibitem[Berman et~al.(2019)Berman, J{\'e}gou, Vedaldi, Kokkinos, and
  Douze]{berman2019multigrain}
Maxim Berman, Herv{\'e} J{\'e}gou, Andrea Vedaldi, Iasonas Kokkinos, and
  Matthijs Douze.
\newblock {MultiGrain}: a unified image embedding for classes and instances.
\newblock \emph{arXiv preprint arXiv:1902.05509}, 2019.

\bibitem[Bertasius et~al.(2021)Bertasius, Wang, and
  Torresani]{bertasius2021space}
Gedas Bertasius, Heng Wang, and Lorenzo Torresani.
\newblock Is space-time attention all you need for video understanding?
\newblock \emph{arXiv preprint arXiv:2102.05095}, 2021.

\bibitem[Boureau et~al.(2010)Boureau, Ponce, and LeCun]{Boureau2010ATA}
Y-Lan Boureau, Jean Ponce, and Yann LeCun.
\newblock A theoretical analysis of feature pooling in visual recognition.
\newblock In \emph{International Conference on Machine Learning}, 2010.

\bibitem[Brabandere et~al.(2016)Brabandere, Jia, Tuytelaars, and
  Gool]{brabandere16nips}
B.~De Brabandere, X.~Jia, T.~Tuytelaars, and L.~Van Gool.
\newblock Dynamic filter networks.
\newblock In \emph{Advances in Neural Information Processing Systems}, 2016.

\bibitem[Brock et~al.(2021)Brock, De, Smith, and Simonyan]{brock2021high}
Andrew Brock, Soham De, Samuel~L Smith, and Karen Simonyan.
\newblock High-performance large-scale image recognition without normalization.
\newblock \emph{arXiv preprint arXiv:2102.06171}, 2021.

\bibitem[Carion et~al.(2020)Carion, Massa, Synnaeve, Usunier, Kirillov, and
  Zagoruyko]{carion2020end}
Nicolas Carion, Francisco Massa, Gabriel Synnaeve, Nicolas Usunier, Alexander
  Kirillov, and Sergey Zagoruyko.
\newblock End-to-end object detection with transformers.
\newblock In \emph{European Conference on Computer Vision}, 2020.

\bibitem[Caron et~al.(2021)Caron, Touvron, Misra, J{\'e}gou, Mairal,
  Bojanowski, and Joulin]{caron2021emerging}
Mathilde Caron, Hugo Touvron, Ishan Misra, Herv{\'e} J{\'e}gou, Julien Mairal,
  Piotr Bojanowski, and Armand Joulin.
\newblock Emerging properties in self-supervised vision transformers.
\newblock \emph{arXiv preprint arXiv:2104.14294}, 2021.

\bibitem[Chen et~al.(2019)Chen, Wang, Pang, Cao, Xiong, Li, Sun, Feng, Liu, Xu,
  Zhang, Cheng, Zhu, Cheng, Zhao, Li, Lu, Zhu, Wu, Dai, Wang, Shi, Ouyang, Loy,
  and Lin]{mmdetection}
Kai Chen, Jiaqi Wang, Jiangmiao Pang, Yuhang Cao, Yu~Xiong, Xiaoxiao Li,
  Shuyang Sun, Wansen Feng, Ziwei Liu, Jiarui Xu, Zheng Zhang, Dazhi Cheng,
  Chenchen Zhu, Tianheng Cheng, Qijie Zhao, Buyu Li, Xin Lu, Rui Zhu, Yue Wu,
  Jifeng Dai, Jingdong Wang, Jianping Shi, Wanli Ouyang, Chen~Change Loy, and
  Dahua Lin.
\newblock {MMDetection}: Open mmlab detection toolbox and benchmark.
\newblock \emph{arXiv preprint arXiv:1906.07155}, 2019.

\bibitem[Child et~al.(2019)Child, Gray, Radford, and
  Sutskever]{child2019generating}
Rewon Child, Scott Gray, Alec Radford, and Ilya Sutskever.
\newblock Generating long sequences with sparse transformers.
\newblock \emph{arXiv preprint arXiv:1904.10509}, 2019.

\bibitem[Choromanski et~al.(2020)Choromanski, Likhosherstov, Dohan, Song, Gane,
  Sarlos, Hawkins, Davis, Mohiuddin, Kaiser, et~al.]{choromanski2020rethinking}
Krzysztof Choromanski, Valerii Likhosherstov, David Dohan, Xingyou Song,
  Andreea Gane, Tamas Sarlos, Peter Hawkins, Jared Davis, Afroz Mohiuddin,
  Lukasz Kaiser, et~al.
\newblock Rethinking attention with performers.
\newblock \emph{arXiv preprint arXiv:2009.14794}, 2020.

\bibitem[Chum et~al.(2007)Chum, Philbin, Sivic, Isard, and
  Zisserman]{chum2007total}
Ondrej Chum, James Philbin, Josef Sivic, Michael Isard, and Andrew Zisserman.
\newblock Total recall: Automatic query expansion with a generative feature
  model for object retrieval.
\newblock In \emph{International Conference on Computer Vision}, 2007.

\bibitem[Contributors(2020)]{mmseg2020}
MMSegmentation Contributors.
\newblock {MMSegmentation}: Openmmlab semantic segmentation toolbox and
  benchmark.
\newblock \url{https://github.com/open-mmlab/mmsegmentation}, 2020.

\bibitem[Cubuk et~al.(2020)Cubuk, Zoph, Shlens, and Le]{cubuk2020randaugment}
Ekin~D Cubuk, Barret Zoph, Jonathon Shlens, and Quoc~V Le.
\newblock Randaugment: Practical automated data augmentation with a reduced
  search space.
\newblock In \emph{Proceedings of the IEEE/CVF Conference on Computer Vision
  and Pattern Recognition Workshops}, 2020.

\bibitem[d'Ascoli et~al.(2021)d'Ascoli, Touvron, Leavitt, Morcos, Biroli, and
  Sagun]{d2021convit}
St{\'e}phane d'Ascoli, Hugo Touvron, Matthew Leavitt, Ari Morcos, Giulio
  Biroli, and Levent Sagun.
\newblock Convit: Improving vision transformers with soft convolutional
  inductive biases.
\newblock \emph{arXiv preprint arXiv:2103.10697}, 2021.

\bibitem[Deng et~al.(2009)Deng, Dong, Socher, Li, Li, and
  Fei-Fei]{deng2009imagenet}
Jia Deng, Wei Dong, Richard Socher, Li-Jia Li, Kai Li, and Li~Fei-Fei.
\newblock Imagenet: A large-scale hierarchical image database.
\newblock In \emph{Computer Vision and Pattern Recognition}, 2009.

\bibitem[Ding et~al.(2021)Ding, Zhang, Han, and Ding]{ding2021repmlp}
Xiaohan Ding, Xiangyu Zhang, Jungong Han, and Guiguang Ding.
\newblock {RepMLP}: Re-parameterizing convolutions into fully-connected layers
  for image recognition.
\newblock \emph{arXiv preprint arXiv:2105.01883}, 2021.

\bibitem[Dosovitskiy et~al.(2021)Dosovitskiy, Beyer, Kolesnikov, Weissenborn,
  Zhai, Unterthiner, Dehghani, Minderer, Heigold, Gelly,
  et~al.]{dosovitskiy2020image}
Alexey Dosovitskiy, Lucas Beyer, Alexander Kolesnikov, Dirk Weissenborn,
  Xiaohua Zhai, Thomas Unterthiner, Mostafa Dehghani, Matthias Minderer, Georg
  Heigold, Sylvain Gelly, et~al.
\newblock An image is worth 16x16 words: Transformers for image recognition at
  scale.
\newblock In \emph{International Conference on Learning Representations}, 2021.

\bibitem[El-Nouby et~al.(2021)El-Nouby, Neverova, Laptev, and
  J{\'e}gou]{el2021training}
Alaaeldin El-Nouby, Natalia Neverova, Ivan Laptev, and Herv{\'e} J{\'e}gou.
\newblock Training vision transformers for image retrieval.
\newblock \emph{arXiv preprint arXiv:2102.05644}, 2021.

\bibitem[Fan et~al.(2021)Fan, Xiong, Mangalam, Li, Yan, Malik, and
  Feichtenhofer]{fan2021multiscale}
Haoqi Fan, Bo~Xiong, Karttikeya Mangalam, Yanghao Li, Zhicheng Yan, Jitendra
  Malik, and Christoph Feichtenhofer.
\newblock Multiscale vision transformers.
\newblock \emph{arXiv preprint arXiv:2104.11227}, 2021.

\bibitem[Gordo et~al.(2017)Gordo, Almaz{\'a}n, Revaud, and
  Larlus]{Gordo2017EndtoEndLO}
Albert Gordo, Jon Almaz{\'a}n, J{\'e}r{\^o}me Revaud, and Diane Larlus.
\newblock End-to-end learning of deep visual representations for image
  retrieval.
\newblock \emph{International journal of Computer Vision}, 124, 2017.

\bibitem[Graham et~al.(2021)Graham, El-Nouby, Touvron, Stock, Joulin,
  J{\'e}gou, and Douze]{graham2021levit}
Ben Graham, Alaaeldin El-Nouby, Hugo Touvron, Pierre Stock, Armand Joulin,
  Herv{\'e} J{\'e}gou, and Matthijs Douze.
\newblock Levit: a vision transformer in convnet's clothing for faster
  inference.
\newblock \emph{arXiv preprint arXiv:2104.01136}, 2021.

\bibitem[Han et~al.(2021)Han, Xiao, Wu, Guo, Xu, and Wang]{han2021transformer}
Kai Han, An~Xiao, Enhua Wu, Jianyuan Guo, Chunjing Xu, and Yunhe Wang.
\newblock Transformer in transformer.
\newblock \emph{arXiv preprint arXiv:2103.00112}, 2021.

\bibitem[He et~al.(2016)He, Zhang, Ren, and Sun]{he2016deep}
Kaiming He, Xiangyu Zhang, Shaoqing Ren, and Jian Sun.
\newblock Deep residual learning for image recognition.
\newblock In \emph{Computer Vision and Pattern Recognition}, 2016.

\bibitem[He et~al.(2017)He, Gkioxari, Doll{\'a}r, and Girshick]{he2017mask}
Kaiming He, Georgia Gkioxari, Piotr Doll{\'a}r, and Ross Girshick.
\newblock Mask r-cnn.
\newblock In \emph{International Conference on Computer Vision}, 2017.

\bibitem[Ho et~al.(2019)Ho, Kalchbrenner, Weissenborn, and
  Salimans]{ho2019axial}
Jonathan Ho, Nal Kalchbrenner, Dirk Weissenborn, and Tim Salimans.
\newblock Axial attention in multidimensional transformers.
\newblock \emph{arXiv preprint arXiv:1912.12180}, 2019.

\bibitem[Horn et~al.(2017)Horn, {Mac Aodha}, Song, Shepard, Adam, Perona, and
  Belongie]{Horn2019INaturalist}
Grant~Van Horn, Oisin {Mac Aodha}, Yang Song, Alexander Shepard, Hartwig Adam,
  Pietro Perona, and Serge~J. Belongie.
\newblock The {iNaturalist} species classification and detection dataset.
\newblock \emph{arXiv preprint arXiv:1707.06642}, 2017.

\bibitem[Hu et~al.(2018)Hu, Shen, and Sun]{hu2018squeeze}
Jie Hu, Li~Shen, and Gang Sun.
\newblock Squeeze-and-excitation networks.
\newblock In \emph{Computer Vision and Pattern Recognition}, 2018.

\bibitem[Huang et~al.(2016)Huang, Sun, Liu, Sedra, and
  Weinberger]{huang2016deep}
Gao Huang, Yu~Sun, Zhuang Liu, Daniel Sedra, and Kilian~Q Weinberger.
\newblock Deep networks with stochastic depth.
\newblock In \emph{European Conference on Computer Vision}, 2016.

\bibitem[Jaegle et~al.(2021)Jaegle, Gimeno, Brock, Zisserman, Vinyals, and
  Carreira]{jaegle2021perceiver}
Andrew Jaegle, Felix Gimeno, Andrew Brock, Andrew Zisserman, Oriol Vinyals, and
  Joao Carreira.
\newblock Perceiver: General perception with iterative attention.
\newblock \emph{arXiv preprint arXiv:2103.03206}, 2021.

\bibitem[J{\'e}gou et~al.(2008)J{\'e}gou, Douze, and
  Schmid]{Jegou2008HammingEA}
Herv{\'e} J{\'e}gou, Matthijs Douze, and Cordelia Schmid.
\newblock Hamming embedding and weak geometric consistency for large scale
  image search.
\newblock In \emph{European Conference on Computer Vision}, 2008.

\bibitem[J{\'e}gou et~al.(2012)J{\'e}gou, Perronnin, Douze, S{\'a}nchez, Perez,
  and Schmid]{jegou2012aggregating}
Herv{\'e} J{\'e}gou, Florent Perronnin, Matthijs Douze, Jorge S{\'a}nchez,
  Patrick Perez, and Cordelia Schmid.
\newblock Aggregating local image descriptors into compact codes.
\newblock \emph{{\sc IEEE} Transactions on Pattern Analysis and Machine
  Intelligence}, 34\penalty0 (9), 2012.

\bibitem[Katharopoulos et~al.(2020)Katharopoulos, Vyas, Pappas, and
  Fleuret]{katharopoulos2020transformers}
Angelos Katharopoulos, Apoorv Vyas, Nikolaos Pappas, and Fran{\c{c}}ois
  Fleuret.
\newblock Transformers are {RNNs}: Fast autoregressive transformers with linear
  attention.
\newblock In \emph{International Conference on Machine Learning}, 2020.

\bibitem[Kirillov et~al.(2019)Kirillov, Girshick, He, and
  Doll{\'a}r]{kirillov2019panoptic}
Alexander Kirillov, Ross Girshick, Kaiming He, and Piotr Doll{\'a}r.
\newblock Panoptic feature pyramid networks.
\newblock In \emph{Computer Vision and Pattern Recognition}, 2019.

\bibitem[Krause et~al.(2013)Krause, Stark, Deng, and Fei-Fei]{Cars2013}
Jonathan Krause, Michael Stark, Jia Deng, and Li~Fei-Fei.
\newblock 3d object representations for fine-grained categorization.
\newblock In \emph{4th International IEEE Workshop on 3D Representation and
  Recognition (3dRR-13)}, 2013.

\bibitem[Krizhevsky(2009)]{Krizhevsky2009LearningML}
Alex Krizhevsky.
\newblock Learning multiple layers of features from tiny images.
\newblock Technical report, CIFAR, 2009.

\bibitem[Lee-Thorp et~al.(2021)Lee-Thorp, Ainslie, Eckstein, and
  Ontanon]{lee2021fnet}
James Lee-Thorp, Joshua Ainslie, Ilya Eckstein, and Santiago Ontanon.
\newblock Fnet: Mixing tokens with fourier transforms.
\newblock \emph{arXiv preprint arXiv:2105.03824}, 2021.

\bibitem[Lin et~al.(2014)Lin, Maire, Belongie, Hays, Perona, Ramanan,
  Doll{\'a}r, and Zitnick]{lin2014microsoft}
Tsung-Yi Lin, Michael Maire, Serge Belongie, James Hays, Pietro Perona, Deva
  Ramanan, Piotr Doll{\'a}r, and C~Lawrence Zitnick.
\newblock Microsoft coco: Common objects in context.
\newblock In \emph{European Conference on Computer Vision}, 2014.

\bibitem[Lin et~al.(2017)Lin, Doll{\'a}r, Girshick, He, Hariharan, and
  Belongie]{lin2017feature}
Tsung-Yi Lin, Piotr Doll{\'a}r, Ross Girshick, Kaiming He, Bharath Hariharan,
  and Serge Belongie.
\newblock Feature pyramid networks for object detection.
\newblock In \emph{Computer Vision and Pattern Recognition}, 2017.

\bibitem[Liu et~al.(2021)Liu, Lin, Cao, Hu, Wei, Zhang, Lin, and
  Guo]{liu2021swin}
Ze~Liu, Yutong Lin, Yue Cao, Han Hu, Yixuan Wei, Zheng Zhang, Stephen Lin, and
  Baining Guo.
\newblock Swin transformer: Hierarchical vision transformer using shifted
  windows.
\newblock \emph{arXiv preprint arXiv:2103.14030}, 2021.

\bibitem[Loshchilov and Hutter(2017)]{loshchilov2017decoupled}
Ilya Loshchilov and Frank Hutter.
\newblock Decoupled weight decay regularization.
\newblock \emph{arXiv preprint arXiv:1711.05101}, 2017.

\bibitem[Melas-Kyriazi(2021)]{melaskyriazi2021doyoueven}
Luke Melas-Kyriazi.
\newblock Do you even need attention? a stack of feed-forward layers does
  surprisingly well on imagenet.
\newblock \emph{arXiv preprint arXiv:2105.02723}, 2021.

\bibitem[Nilsback and Zisserman(2008)]{Nilsback08}
M-E. Nilsback and A.~Zisserman.
\newblock Automated flower classification over a large number of classes.
\newblock In \emph{Proceedings of the Indian Conference on Computer Vision,
  Graphics and Image Processing}, 2008.

\bibitem[Parmar et~al.(2018)Parmar, Vaswani, Uszkoreit, Kaiser, Shazeer, Ku,
  and Tran]{parmar2018image}
Niki Parmar, Ashish Vaswani, Jakob Uszkoreit, Lukasz Kaiser, Noam Shazeer,
  Alexander Ku, and Dustin Tran.
\newblock Image transformer.
\newblock In \emph{International Conference on Machine Learning}, 2018.

\bibitem[Philbin et~al.(2007)Philbin, Chum, Isard, Sivic, and
  Zisserman]{Philbin07}
J.~Philbin, O.~Chum, M.~Isard, J.~Sivic, and A.~Zisserman.
\newblock Object retrieval with large vocabularies and fast spatial matching.
\newblock In \emph{Computer Vision and Pattern Recognition}, 2007.

\bibitem[Qiu et~al.(2019)Qiu, Ma, Levy, Yih, Wang, and Tang]{qiu2019blockwise}
Jiezhong Qiu, Hao Ma, Omer Levy, Scott Wen-tau Yih, Sinong Wang, and Jie Tang.
\newblock Blockwise self-attention for long document understanding.
\newblock \emph{arXiv preprint arXiv:1911.02972}, 2019.

\bibitem[Radenovi{\'c} et~al.(2018{\natexlab{a}})Radenovi{\'c}, Iscen, Tolias,
  Avrithis, and Chum]{radenovic2018revisiting}
Filip Radenovi{\'c}, Ahmet Iscen, Giorgos Tolias, Yannis Avrithis, and
  Ond{\v{r}}ej Chum.
\newblock Revisiting oxford and paris: Large-scale image retrieval
  benchmarking.
\newblock In \emph{Computer Vision and Pattern Recognition},
  2018{\natexlab{a}}.

\bibitem[Radenovi{\'c} et~al.(2018{\natexlab{b}})Radenovi{\'c}, Tolias, and
  Chum]{radenovic2018fine}
Filip Radenovi{\'c}, Giorgos Tolias, and Ondrej Chum.
\newblock Fine-tuning {CNN} image retrieval with no human annotation.
\newblock \emph{{\sc IEEE} Transactions on Pattern Analysis and Machine
  Intelligence}, 2018{\natexlab{b}}.

\bibitem[Radosavovic et~al.(2020)Radosavovic, Kosaraju, Girshick, He, and
  Doll{\'a}r]{radosavovic2020designing}
Ilija Radosavovic, Raj~Prateek Kosaraju, Ross Girshick, Kaiming He, and Piotr
  Doll{\'a}r.
\newblock Designing network design spaces.
\newblock In \emph{Computer Vision and Pattern Recognition}, 2020.

\bibitem[Ranftl et~al.(2021)Ranftl, Bochkovskiy, and Koltun]{ranftl2021vision}
Ren{\'e} Ranftl, Alexey Bochkovskiy, and Vladlen Koltun.
\newblock Vision transformers for dense prediction.
\newblock \emph{arXiv preprint arXiv:2103.13413}, 2021.

\bibitem[Recht et~al.(2019)Recht, Roelofs, Schmidt, and
  Shankar]{recht2019imagenet}
Benjamin Recht, Rebecca Roelofs, Ludwig Schmidt, and Vaishaal Shankar.
\newblock Do imagenet classifiers generalize to imagenet?
\newblock In \emph{International Conference on Machine Learning}, 2019.

\bibitem[Shen et~al.(2021)Shen, Zhang, Zhao, Yi, and Li]{shen2021efficient}
Zhuoran Shen, Mingyuan Zhang, Haiyu Zhao, Shuai Yi, and Hongsheng Li.
\newblock Efficient attention: Attention with linear complexities.
\newblock In \emph{Proceedings of the IEEE/CVF Winter Conference on
  Applications of Computer Vision}, 2021.

\bibitem[Sukhbaatar et~al.(2019)Sukhbaatar, Grave, Bojanowski, and
  Joulin]{sukhbaatar2019adaptive}
Sainbayar Sukhbaatar, Edouard Grave, Piotr Bojanowski, and Armand Joulin.
\newblock Adaptive attention span in transformers.
\newblock \emph{arXiv preprint arXiv:1905.07799}, 2019.

\bibitem[Tan and Le(2019)]{tan2019efficientnet}
Mingxing Tan and Quoc Le.
\newblock Efficientnet: Rethinking model scaling for convolutional neural
  networks.
\newblock In \emph{International Conference on Machine Learning}. PMLR, 2019.

\bibitem[Thomee et~al.(2016)Thomee, Shamma, Friedland, Elizalde, Ni, Poland,
  Borth, and Li]{thomee2016yfcc100m}
Bart Thomee, David~A Shamma, Gerald Friedland, Benjamin Elizalde, Karl Ni,
  Douglas Poland, Damian Borth, and Li-Jia Li.
\newblock Yfcc100m: The new data in multimedia research.
\newblock \emph{Communications of the ACM}, 59\penalty0 (2):\penalty0 64--73,
  2016.

\bibitem[Tolias et~al.(2016{\natexlab{a}})Tolias, Avrithis, and
  J{\'e}gou]{tolias2016image}
Giorgos Tolias, Yannis Avrithis, and Herv{\'e} J{\'e}gou.
\newblock Image search with selective match kernels: aggregation across single
  and multiple images.
\newblock \emph{International journal of Computer Vision}, 116\penalty0 (3),
  2016{\natexlab{a}}.

\bibitem[Tolias et~al.(2016{\natexlab{b}})Tolias, Sicre, and
  J{\'e}gou]{tolias2016particular}
Giorgos Tolias, Ronan Sicre, and Herv{\'e} J{\'e}gou.
\newblock Particular object retrieval with integral max-pooling of cnn
  activations.
\newblock In \emph{International Conference on Learning Representations},
  2016{\natexlab{b}}.

\bibitem[Tolias et~al.(2020)Tolias, Jenicek, and Chum]{tolias2020learning}
Giorgos Tolias, Tomas Jenicek, and Ond{\v{r}}ej Chum.
\newblock Learning and aggregating deep local descriptors for instance-level
  recognition.
\newblock In \emph{European Conference on Computer Vision}, 2020.

\bibitem[Tolstikhin et~al.(2021)Tolstikhin, Houlsby, Kolesnikov, Beyer, Zhai,
  Unterthiner, Yung, Steiner, Keysers, Uszkoreit, Lucic, and
  Dosovitskiy]{Tolstikhin21mixer}
Ilya Tolstikhin, Neil Houlsby, Alexander Kolesnikov, Lucas Beyer, Xiaohua Zhai,
  Thomas Unterthiner, Jessica Yung, Andreas Steiner, Daniel Keysers, Jakob
  Uszkoreit, Mario Lucic, and Alexey Dosovitskiy.
\newblock {MLP-Mixer}: An all-{MLP} architecture for vision.
\newblock \emph{arXiv preprint arXiv:2105.01601}, 2021.

\bibitem[Touvron et~al.(2019)Touvron, Vedaldi, Douze, and
  J{\'e}gou]{touvron2019fixing}
H~Touvron, A~Vedaldi, M~Douze, and H~J{\'e}gou.
\newblock Fixing the train-test resolution discrepancy.
\newblock \emph{Advances in Neural Information Processing Systems}, 2019.

\bibitem[Touvron et~al.(2020{\natexlab{a}})Touvron, Cord, Douze, Massa,
  Sablayrolles, and J\'egou]{touvron2020deit}
Hugo Touvron, Matthieu Cord, Matthijs Douze, Francisco Massa, Alexandre
  Sablayrolles, and Herv\'e J\'egou.
\newblock Training data-efficient image transformers and distillation through
  attention.
\newblock \emph{arXiv preprint arXiv:2012.12877}, 2020{\natexlab{a}}.

\bibitem[Touvron et~al.(2020{\natexlab{b}})Touvron, Vedaldi, Douze, and
  J{\'e}gou]{touvron2020fixing}
Hugo Touvron, Andrea Vedaldi, Matthijs Douze, and Herv{\'e} J{\'e}gou.
\newblock Fixing the train-test resolution discrepancy: Fixefficientnet.
\newblock \emph{arXiv preprint arXiv:2003.08237}, 2020{\natexlab{b}}.

\bibitem[Touvron et~al.(2021{\natexlab{a}})Touvron, Bojanowski, Caron, Cord,
  El-Nouby, Grave, Joulin, Synnaeve, Verbeek, and J\'egou]{Touvron21ResMLP}
Hugo Touvron, Piotr Bojanowski, Mathilde Caron, Matthieu Cord, Alaaeldin
  El-Nouby, Edouard Grave, Armand Joulin, Gabriel Synnaeve, Jakob Verbeek, and
  Herv\'e J\'egou.
\newblock {ResMLP}: Feedforward networks for image classification with
  data-efficient training.
\newblock \emph{arXiv preprint arXiv:2105.03404}, 2021{\natexlab{a}}.

\bibitem[Touvron et~al.(2021{\natexlab{b}})Touvron, Cord, Sablayrolles,
  Synnaeve, and J{\'e}gou]{touvron2021going}
Hugo Touvron, Matthieu Cord, Alexandre Sablayrolles, Gabriel Synnaeve, and
  Herv{\'e} J{\'e}gou.
\newblock Going deeper with image transformers.
\newblock \emph{arXiv preprint arXiv:2103.17239}, 2021{\natexlab{b}}.

\bibitem[Vaswani et~al.(2017)Vaswani, Shazeer, Parmar, Uszkoreit, Jones, Gomez,
  Kaiser, and Polosukhin]{vaswani2017attention}
Ashish Vaswani, Noam Shazeer, Niki Parmar, Jakob Uszkoreit, Llion Jones,
  Aidan~N Gomez, {\L}ukasz Kaiser, and Illia Polosukhin.
\newblock Attention is all you need.
\newblock In \emph{Advances in Neural Information Processing Systems}, 2017.

\bibitem[Wang et~al.(2020)Wang, Li, Khabsa, Fang, and Ma]{wang2020linformer}
Sinong Wang, Belinda Li, Madian Khabsa, Han Fang, and Hao Ma.
\newblock Linformer: Self-attention with linear complexity.
\newblock \emph{arXiv preprint arXiv:2006.04768}, 2020.

\bibitem[Wang et~al.(2021)Wang, Xie, Li, Fan, Song, Liang, Lu, Luo, and
  Shao]{wang2021pyramid}
Wenhai Wang, Enze Xie, Xiang Li, Deng-Ping Fan, Kaitao Song, Ding Liang, Tong
  Lu, Ping Luo, and Ling Shao.
\newblock Pyramid vision transformer: A versatile backbone for dense prediction
  without convolutions.
\newblock \emph{arXiv preprint arXiv:2102.12122}, 2021.

\bibitem[Wightman(2019)]{rw2019timm}
Ross Wightman.
\newblock Pytorch image models.
\newblock \url{https://github.com/rwightman/pytorch-image-models}, 2019.

\bibitem[Wu and He(2018)]{wu2018group}
Yuxin Wu and Kaiming He.
\newblock Group normalization.
\newblock In \emph{European Conference on Computer Vision}, 2018.

\bibitem[Xiao et~al.(2018)Xiao, Liu, Zhou, Jiang, and Sun]{xiao2018unified}
Tete Xiao, Yingcheng Liu, Bolei Zhou, Yuning Jiang, and Jian Sun.
\newblock Unified perceptual parsing for scene understanding.
\newblock In \emph{European Conference on Computer Vision}, 2018.

\bibitem[Xie et~al.(2017)Xie, Girshick, Doll{\'a}r, Tu, and
  He]{xie2017aggregated}
Saining Xie, Ross Girshick, Piotr Doll{\'a}r, Zhuowen Tu, and Kaiming He.
\newblock Aggregated residual transformations for deep neural networks.
\newblock In \emph{Computer Vision and Pattern Recognition}, 2017.

\bibitem[Xie et~al.(2021)Xie, Lin, Yao, Zhang, Dai, Cao, and Hu]{xie2021self}
Zhenda Xie, Yutong Lin, Zhuliang Yao, Zheng Zhang, Qi~Dai, Yue Cao, and Han Hu.
\newblock Self-supervised learning with swin transformers.
\newblock \emph{arXiv preprint arXiv:2105.04553}, 2021.

\bibitem[Xiong et~al.(2021)Xiong, Zeng, Chakraborty, Tan, Fung, Li, and
  Singh]{xiong2021nystr}
Yunyang Xiong, Zhanpeng Zeng, Rudrasis Chakraborty, Mingxing Tan, Glenn Fung,
  Yin Li, and Vikas Singh.
\newblock Nystr\" omformer: A nystr\" om-based algorithm for approximating
  self-attention.
\newblock \emph{arXiv preprint arXiv:2102.03902}, 2021.

\bibitem[Yuan et~al.(2021{\natexlab{a}})Yuan, Guo, Liu, Zhou, Yu, and
  Wu]{yuan2021incorporating}
Kun Yuan, Shaopeng Guo, Ziwei Liu, Aojun Zhou, Fengwei Yu, and Wei Wu.
\newblock Incorporating convolution designs into visual transformers.
\newblock \emph{arXiv preprint arXiv:2103.11816}, 2021{\natexlab{a}}.

\bibitem[Yuan et~al.(2021{\natexlab{b}})Yuan, Chen, Wang, Yu, Shi, Jiang, Tay,
  Feng, and Yan]{yuan2021tokens}
Li~Yuan, Yunpeng Chen, Tao Wang, Weihao Yu, Yujun Shi, Zihang Jiang, Francis~EH
  Tay, Jiashi Feng, and Shuicheng Yan.
\newblock Tokens-to-token {ViT}: Training vision transformers from scratch on
  {ImageNet}.
\newblock \emph{arXiv preprint arXiv:2101.11986}, 2021{\natexlab{b}}.

\bibitem[Zaheer et~al.(2020)Zaheer, Guruganesh, Dubey, Ainslie, Alberti,
  Ontanon, Pham, Ravula, Wang, Yang, et~al.]{zaheer2020big}
Manzil Zaheer, Guru Guruganesh, Avinava Dubey, Joshua Ainslie, Chris Alberti,
  Santiago Ontanon, Philip Pham, Anirudh Ravula, Qifan Wang, Li~Yang, et~al.
\newblock Big bird: Transformers for longer sequences.
\newblock \emph{arXiv preprint arXiv:2007.14062}, 2020.

\bibitem[Zhang et~al.(2021)Zhang, Dai, Yang, Xiao, Yuan, Zhang, and
  Gao]{zhang2021multi}
Pengchuan Zhang, Xiyang Dai, Jianwei Yang, Bin Xiao, Lu~Yuan, Lei Zhang, and
  Jianfeng Gao.
\newblock Multi-scale vision longformer: A new vision transformer for
  high-resolution image encoding.
\newblock \emph{arXiv preprint arXiv:2103.15358}, 2021.

\bibitem[Zhao et~al.(2020)Zhao, Jia, and Koltun]{zhao20cvpr}
Hengshuang Zhao, Jiaya Jia, and Vladlen Koltun.
\newblock Exploring self-attention for image recognition.
\newblock In \emph{Computer Vision and Pattern Recognition}, 2020.

\bibitem[Zheng et~al.(2020)Zheng, Lu, Zhao, Zhu, Luo, Wang, Fu, Feng, Xiang,
  Torr, et~al.]{zheng2020rethinking}
Sixiao Zheng, Jiachen Lu, Hengshuang Zhao, Xiatian Zhu, Zekun Luo, Yabiao Wang,
  Yanwei Fu, Jianfeng Feng, Tao Xiang, Philip~HS Torr, et~al.
\newblock Rethinking semantic segmentation from a sequence-to-sequence
  perspective with transformers.
\newblock \emph{arXiv preprint arXiv:2012.15840}, 2020.

\bibitem[Zhou et~al.(2017)Zhou, Zhao, Puig, Fidler, Barriuso, and
  Torralba]{zhou2017scene}
Bolei Zhou, Hang Zhao, Xavier Puig, Sanja Fidler, Adela Barriuso, and Antonio
  Torralba.
\newblock Scene parsing through ade20k dataset.
\newblock In \emph{Computer Vision and Pattern Recognition}, 2017.

\end{thebibliography}
}

\fi

\appendix
\clearpage
\counterwithin{figure}{section}
\counterwithin{table}{section}
\counterwithin{equation}{section}

\pagenumbering{Roman}  

\newpage
\vskip .375in
\begin{center}
{\Large \bf \inserttitle \\ \vspace{0.5cm} \large Appendix \par}
  \vspace*{24pt}
  {
  \par
  }
\end{center}

\section{Preliminary study on Vision Transformers (ViT)}
\label{sec:preliminary}

In this appendix we report the results associated with our preliminary study on high-resolution transformers. Most of the experiments were carried out on the ViT architecture~\cite{dosovitskiy2020image} with DeiT training~\cite{touvron2020deit}, and intended to analyze different aspects of transformers when considering images with varying resolution or high-resolution images specifically. 

\subsection{Impact of resolution versus patch size} 
\label{sec:res_patch}

\def \mysp {\hspace{12pt}}

\begin{figure*}[htb]
\begin{minipage}[c]{0.45\textwidth}
    \includegraphics[width=\linewidth]{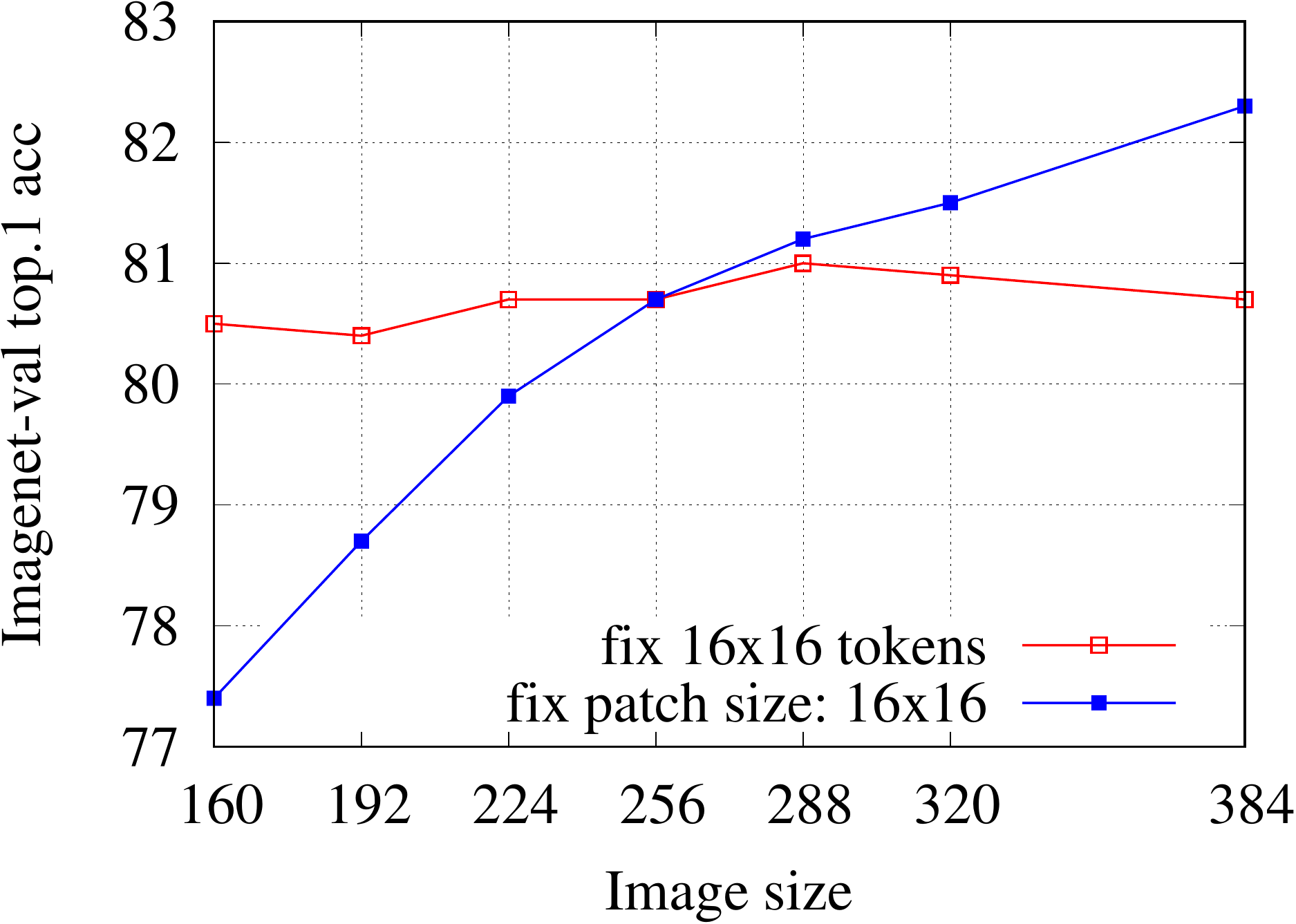}
\end{minipage}
    \hfill
\begin{minipage}[c]{0.49\textwidth}
    \scalebox{0.75}{
    \begin{tabular}{@{\mysp}l@{\mysp}|@{\mysp}c@{\mysp}c@{\mysp}c@{\mysp}c@{\mysp}c@{\mysp}c@{\mysp}c@{\mysp}}
    \multicolumn{7}{c}{Variable patch size\vphantom{\rule[-10pt]{1pt}{10pt}} } \\
    \toprule
        Image Size & 80 & 112 & 160  & 256 & 320 & 384\\
        \midrule
        Patch Size & 5 & 7 & 10 &  16 & 20 & 24 \\
        \midrule        
        Top-1 & 78.2 & 79.7 & 80.5 & 80.7 & 80.9 & 80.7 \\
         \bottomrule
    \end{tabular}}
    \vspace*{10pt} \\ 
    \scalebox{0.75}{
    \begin{tabular}{@{\mysp}c@{\mysp}|@{\mysp}c@{\mysp}c@{\mysp}c@{\mysp}c@{\mysp}c@{\mysp}c@{\mysp}c@{\mysp}}
    \multicolumn{7}{c}{Variable number of tokens size \vphantom{\rule[-10pt]{1pt}{10pt}}} \\
    \toprule
        Image Size &  160  & 224  & 256 & 288 & 320 & 384\\
        \midrule
        \# of tokens  & 100  & 196  & 256 & 324 & 400 & 576  \\
        \midrule        
        Top-1 & 77.4  & 79.9  & 80.7 & 81.2 & 81.5 & 82.3 \\
         \bottomrule
    \end{tabular}}
\end{minipage}    
    \caption{\textbf{Impact of input resolution on accuracy for DeiT-S}. We consider different image resolutions, and either (1) \textcolor{red!80}{increase the patch size while keeping the number of tokens fixed}; or (2) \textcolor{blue!80}{keep the patch size fixed and use more tokens}. Larger input images are beneficial if the number of tokens increases. The impact of a change of a resolution for a constant number of patches (of varying size) is almost neutral.  
As one can observe, the main driver of performance is the number of patches. The patch size has a limited impact on the accuracy, except when considering very small ones. We have observed and confirmed similar trends with \ours models.
    \label{fig:imsize_patch_tokens}}

\end{figure*}

\subsection{Approximate attention models in ViT with DeiT training}
\label{sec:approx_att}

In Table A.1, we report the results that we obtain by replacing the Multi-headed Self-attention operation with efficient variants \cite{ho2019axial, shen2021efficient, wang2020linformer, wang2021pyramid} in the DeiT-S backbone.  First, we can notice that for all efficient self-attention choices there is a clear drop in performance compared to the Deit-S baseline. The spatial reduction attention (SRA) proposed in PVT \cite{wang2021pyramid} has a significantly weaker performance compared to the full-attention with a quadratic complexity that is more efficient than full-attention by only a constant factor $R^2$. Linformer \cite{wang2020linformer} provides a better accuracy compared to SRA, however, it is also clearly weaker than full-attention. Moreover, Linformer does not have the flexibility of processing variable length sequences which limits its application in many computer vision tasks. Efficient attention \cite{shen2021efficient} provides a better trade-off than the aforementioned methods, with improved accuracy and linear complexity. However, it has a 3.6\% drop in performance compared to full-attention. Finally, axial attention \cite{ho2019axial} provides the strongest performance among the efficient attention variants we studied with a 1.5\% drop in accuracy compared to the baseline. We observe a saving in memory usage, but a drop in speed due to the separate row and column attention operations. Our observations are consistent with \cite{dosovitskiy2020image}.
\begin{table}[h!]
\small
\captionof{table}{\footnotesize \textbf{ImageNet Top-1 accuracy of efficient self-attention variants} (after 300 epochs of training).}\vspace*{3pt}
\centering 
\begin{tabular}{l|c|c}
     \toprule    
     Model & Complexity & Top-1  \\
     \midrule
     DeiT-S~\cite{touvron2020deit} & $\mathcal{O}(N^2)$ & 79.9  \\
     \midrule
     SRA (Average Pool)~\cite{wang2021pyramid} & $\mathcal{O}(N^2 / R^2)$  & 73.5 \\
     SRA (Convolutional)~\cite{wang2021pyramid} & $\mathcal{O}(N^2 / R^2)$ & 74.0 \\
     Linformer (k=$\sqrt n$)~\cite{wang2020linformer} & $\mathcal{O}(k N)$ & 75.7 \\
     Efficient Transformer~\cite{shen2021efficient} & $\mathcal{O}(N)$ & 76.3 \\
    Axial~\cite{ho2019axial} & $\mathcal{O}(N \sqrt{N})$ & 78.4 \\
     \bottomrule
\end{tabular}    
\end{table}

\subsection{Training and testing with varying resolution}

As discussed in the main manuscript, for several tasks it is important that the network is able to handle images of varying resolutions. This is the case, for instance, for image segmentation, image detection, or image retrieval where the object of interest may have very different sizes. We present an analysis of train/test resolution trade-off in Table A.2.

{\centering 
\begin{table}[h!]
    \captionof{table}{\footnotesize \textbf{Trade-off between train and test resolutions for DeiT.} MS refers to multi-scale training, where the models have seen images from different resolutions at training time.}\vspace*{3pt}
    \centering 
    \scalebox{0.85}{
    \begin{tabular}{c|ccccc|c}
        \toprule
         Test / Train & 160 & 224 & 256 & 288 & 320  & MS  \\
         \toprule
         160 & \textbf{77.2} & 75.9 & 73.3 & 68.2 & 59.6   &  76.3  \\% 76.1 & - \\ 
         224 & 78.0 & \textbf{79.9} & 79.9 & 79.0 & 77.9   & 79.6 \\ %
         256 & 77.3 & 80.4 & \textbf{80.7} & 80.2 & 79.9   & 80.6  \\ %
         288 & 76.3 & 80.4 & 81.0 & \textbf{81.2} & 80.8   & 81.0 \\ %
         320 & 75.0 & 80.1 & 80.9 & 81.3 & \textbf{81.5}   & 81.3 \\ %
         \bottomrule
    \end{tabular}}
    \label{tab:multi-scale}
    
\end{table}}

\section{Additional details of training and our architecture} 

\subsection{Sinusoidal Positional Encoding}
\label{sec:position_encoding}

We adopt a sinusoidal positional encoding as proposed by \citet{vaswani2017attention} and adapted to the 2D case by \citet{carion2020end}. However we depart from this method in that we first produce this encoding in an intermediate 64-d space before projecting it to the working space of the transformers. More precisely, in our implementation each of the $x$ and $y$ coordinates is encoded using 32 dimensions corresponding to cosine and sine functions with different frequencies (16 frequency for each function). The encoding of both coordinates are eventually concatenated to obtain a 64 dimension 2D positional encoding. Finally, the 64 dimension positional encoding is linearly projected to the working dimension of the model $d$.

\subsection{Obtaining Feature Pyramid for Dense Prediction}
\label{sec:fpn_api}

For state-of-the-art detection and segmentation models, FPN is an important component which provides features of multiple scales. We adapt \ours to be compatible with FPN detection and segmentation methods through a simple re-scaling of the features extracted from different layers. In particular, for models with 12 layers, we extract features from the 4$^{\text{th}}$, 6$^{\text{th}}$, 8$^{\text{th}}$ and 12$^{\text{th}}$ layers respectively. As for models with 24 layers, we extract features from 8$^{\text{th}}$, 12$^{\text{th}}$, 16$^{\text{th}}$ and 24$^{\text{th}}$ layers. Concerning the re-scaling of the features, the 4 feature levels are downsized by a ratio of 4, 8, 16 and 32 compared to the input image size. Feature downsizing is performed with max pooling and upsampling is achieved using a single layer of transposed convolutions with kernel size $k=2$ and stride $s=2$.  

\subsection{Hyper-parameters: LayerScale initialization and Stochastic Depth drop-rate} 
\label{sec:ls_sd_values}

We list the stochastic depth $d_r$ and LayerScale initialization $\epsilon$ hyperparameters used by each of our models in Table~\ref{tab:drop_layerscale}.

\begin{table}[htb]
    \centering
    \caption{\textbf{Hyperparameters used for training our models}, including the Stochastic depth drop rate $d_r$ and LayerScale initialization $\epsilon$.}
    \scalebox{0.9}{
    \begin{tabular}{c c |r r}
         \toprule
         Model & Patch size & $d_r$ &  $\epsilon$  \\
         \midrule
         \ours-N12 & 8 \& 16 & 0.0 & $1.0$ \\ 
         \ours-T12 & 8 \& 16 & 0.0 & $1.0$ \\ 
         \ours-T24 & 8 \& 16 & 0.05 & $10^{-5}$ \\ 
         \ours-S12 & 8 \& 16 & 0.05 & 1.0 \\
         \ours-S24 & 8 \& 16 & 0.1 &  $10^{-5}$ \\ 
         \ours-M24 & 8 \& 16 & 0.15 & $10^{-5}$ \\ 
         \ours-L24 & 16 & 0.25 & $10^{-5}$ \\ 
         \ours-L24 & 8  & 0.3 & $10^{-5}$ \\ 
         \bottomrule
    \end{tabular}}
    \label{tab:drop_layerscale}
\end{table}

\section{Pseudo-code}

We provide a PyTorch-style pseudo code of the Cross-covariance attention operation. The pseudo code resembles the Timm library \cite{rw2019timm} implementation of token self-attention. We show that XCA only requires few modifications, namely the $\ell_2$ normalization, setting the learnable temperature parameters and a transpose operation of the keys, queries and values. 

\begin{algorithm}[htb]
\caption{Pseudocode of XCA  in a PyTorch-like style.}
\label{alg:code}
\definecolor{codeblue}{rgb}{0.25,0.5,0.5}
\lstset{
  backgroundcolor=\color{white},
  basicstyle=\fontsize{7.2pt}{7.2pt}\ttfamily\selectfont,
  columns=fullflexible,
  breaklines=true,
  captionpos=b,
  commentstyle=\fontsize{7.2pt}{7.2pt}\color{codeblue},
  keywordstyle=\fontsize{7.2pt}{7.2pt},
}
\begin{lstlisting}[language=python]
# self.qkv: nn.Linear(dim, dim * 3, bias=qkv_bias)
# self.temp: nn.Parameter(torch.ones(num_headss, 1, 1))

def forward(self, x):
    B, N, C = x.shape
    qkv = self.qkv(x).reshape(B, N, 3, self.num_heads, C // self.num_heads)
    qkv = qkv.permute(2, 0, 3, 1, 4)
    q, k, v = qkv[0], qkv[1], qkv[2]   # split into query, key and value

    q = q.transpose(-2, -1) 
    k = k.transpose(-2, -1)         # Transpose to shape (B, h, C, N)
    v = v.transpose(-2, -1)

    q = F.normalize(q, dim=-1, p=2)    # L2 Normalization across the token dimension
    k = F.normalize(k, dim=-1, p=2)    

    attn = (k @ q.transpose(-2, -1))       # Computing the block diagonal cross-covariance matrix
    attn = attn * self.temp                # Adjusting the activations scale with temperature parameter
    attn = attn.softmax(dim=-1)            # d x d attention map

    x = attn @ v    # Apply attention to mix channels per token 
    x = x.permute(0, 3, 1, 2).reshape(B, N, C)  
    x = self.proj(x)
    return x
\end{lstlisting}
\end{algorithm}

\section{Additional results}

\subsection{More \ours models}

We present additional results for our \ours models in Table~\ref{tab:xct_models_details}. We include performance of 384$\times$384 images using a 16$\times$16 patch size as well as results for images with 224$\times$224 resolution using patch size of 8$\times$8.

\begin{table}[htb]
    \caption{\footnotesize \textbf{ImageNet-1k top-1 accuracy of \ours} for additional combinations of image and patch sizes.\vspace*{3pt}
    }
    \centering
    \scalebox{0.72}{
    \begin{tabular}{l|cccr|rccc|rccc}
    \toprule
         Models & Depth & $d$ & \#Blocks & \@params & \multicolumn{4}{c|}{$P=(16\times16)$} & \multicolumn{4}{c}{$P=(8\times8)$}   \\ 
         \cmidrule{6-9}
         \cmidrule{10-13}
         & & &   &  & GFLOPs & @224 & @224\alambic & @384$\uparrow$ 
         & GFLOPs & @224 & @224\alambic   & @384$\uparrow$   \\
        \midrule
        \OURS-N12    & 12 & \multirow{1}{*}{128} & \multirow{1}{*}{4}  & 3M & 0.5 & 69.9 & 72.2 & 75.4 &    2.1 & 73.8 & 76.3 & 77.8 \\
        \OURS-T12    & 12 & \multirow{1}{*}{192} & \multirow{1}{*}{4}  & 7M & 1.2 & 77.1 & 78.6 & 80.9    & 4.8 & 79.7 & 81.2 & 82.4   \\
        \OURS-T24    & 24 & 192 & 4 & 12M &  2.3 & 79.4 & 80.4  & 82.6   & 9.2  & 81.9 & 82.6 &  83.7 \\
        \OURS-S12      & 12 & \multirow{1}{*}{384} & \multirow{1}{*}{8} & 26M & 4.8 & 82.0 & 83.3 & 84.7    & 18.9 & 83.4 & 84.2 &  85.1  \\ 
        \OURS-S-24      & 24 & 384 & 8 & 48M & 9.1 & 82.6 & 83.9 & 85.1 &   36.0 & 83.9 &  84.9 & 85.6  \\ 
        \OURS-M24      & 24 & 512 & 8  & 84M & 16.2  & 82.7 & 84.3 & 85.4  & 63.9 & 83.7 & 85.1 & 85.8    \\
        \OURS-L24      & 24 &\multirow{1}{*}{768} & \multirow{1}{*}{16} &  189M & 36.1 & 82.9 & 84.9 & 85.8 & 142.2 & 84.4 & 85.4 & 86.0 \\
    \bottomrule     
 \end{tabular}
 } %
 \label{tab:xct_models_details}
\end{table}

\subsection{Transfer Learning}

 In order to further demonstrate the flexibility and generality of our models, we report transfer learning experiments in Table~\ref{tab:transfer} for models that have been pre-trained using ImageNet-1k and finetuned for other datasets including CIFAR-10, CIFAR-100~\cite{Krizhevsky2009LearningML}, Flowers-102~\cite{Nilsback08}, Stanford Cars~\cite{Cars2013} and iNaturalist~\cite{Horn2019INaturalist}.  We observe that the \ours models provide competitive performance when compared to strong baselines like ViT-B, ViT-L, DeiT-B and EfficientNet-B7.

\begin{table}
    \centering
    \caption{
    \textbf{Evaluation on transfer learning.}
    }\vspace*{3pt}
    \scalebox{0.9}{
    \begin{tabular}{@{\ }c@{\mysp}c@{\mysp}c@{\mysp}c@{\mysp}c@{\mysp}c@{\mysp}c@{\ }}
    \toprule
    Architecture              
        & {CIFAR$_{10}$}
        & {CIFAR$_{100}$}  
        & {Flowers102} 
        & {Cars} 
        & {iNat$_{18}$} 
        & {iNat$_{19}$} 
      \\
    \midrule
    EfficientNet-B7~\cite{tan2019efficientnet}  & \underline{98.9} & \textbf{91.7}  & \textbf{98.8} & \textbf{94.7} & \_ & \_  \\
    ViT-B/16~\cite{dosovitskiy2020image}   & 98.1 & 87.1 & 89.5 & \_   & \_   & \_   \\
    ViT-L/16~\cite{dosovitskiy2020image}  & 97.9 & 86.4 & 89.7 & \_   & \_   & \_   \\
    Deit-B/16~\cite{touvron2020deit} \alambic  & \textbf{99.1} & 91.3 & \textbf{98.8} & 92.9 & \underline{73.7} & \underline{78.4}  \\
    \midrule
    \ours-S24/16 \alambic  & \textbf{99.1} & 91.2 & 97.4 & 92.8 & 68.8 & 76.1 \\
    \ours-M24/16 \alambic & \textbf{99.1} & \underline{91.4} & 98.2 & 93.4 & 72.6 & 78.1   \\
    \ours-L24/16 \alambic & \textbf{99.1} & 91.3 & \underline{98.3} & \underline{93.7} & \textbf{75.6} & \textbf{79.3 }\\

    \bottomrule
    \end{tabular}}

    \label{tab:transfer}
\end{table}

\subsection{Image Retrieval}

\paragraph{Context of this study. } 
Vision-based retrieval tasks such as landmark or particular object retrieval have been dominated in the last years by  methods extracting features from high-resolution images. Traditionally, the image description was obtained as the aggregation of local descriptors, like in VLAD~\cite{jegou2012aggregating}. Most of the modern methods now rely on convolutional neural networks~\cite{berman2019multigrain,Gordo2017EndtoEndLO,tolias2016particular}. 
In a recent paper, El-Nouby \emph{et al.}~\cite{el2021training} show promising results with vision transformers, however they also underline the inherent scalability limitation associated with the fact that ViT models do not scale well with image resolution. Therefore, it  cannot compete with convolutional neural networks whose performance readily improve with higher resolution images. Our \OURS models do not suffer from this limitation: our models scale linearly with the number of pixels, like convnets, and therefore makes it possible to use off-the-shelf methods initially developed for retrieval with high-resolution images.

\subsubsection{Datasets and evaluation measure} 
In each benchmark, a set of query images is searched in a database of images and the performance is measured as the mean average precision. %

The Holidays~\cite{Jegou2008HammingEA} dataset contains images of 500 different objects or scenes. 
We use the version of the dataset where the orientation of images (portrait or landscape) has been corrected.
Oxford~\cite{Philbin07} is a dataset of building images, which corresponds to famous landmark in Oxford. A similar dataset has been produced for famous monuments in Paris and referred to as Paris6k~\cite{chum2007total}. 

\begin{table}[h!]
\begin{center}
\caption{\textbf{The basic statistics on the image retrieval datasets.}}\vspace*{3pt}
\scalebox{0.9}{%
    \begin{tabular}{l|rccc}
    \toprule
         & \multicolumn{2}{c}{number of images} & \multirow{2}{*}{nb of instances}  \\
                 \cmidrule(lr){2-3}
         Dataset & database & queries &  &   \\
    \midrule
    Holidays     & 1491 & 500 & 500 \\
    R-Oxford     & 4993 & 70 & 26  \\
    \bottomrule
    \end{tabular}
}
\end{center}
\end{table}

We use the revisited version of the Oxford benchmark~\cite{radenovic2018revisiting}, which breaks down the evaluation into easy, medium and hard categories. 
We report results on the "medium" and "hard" settings, as we observed that the ordering of techniques does not change under the easy measures.

\subsubsection{Image representation: global and local description with \OURS} 

We consider three existing methods to extract an image vector representations from the pre-trained \OURS  models. 
Note that to the best of our knowledge, for the first time we extract local features from the output layer of a transformer layer, and treat them as patches fed to traditional state-of-the-art methods based on matching local descriptors or CNN. 
\paragraph{CLS token.} Similar to \citet{el2021training} with ViT, we use the final vector as the image descriptor. In this context, the introduction of class-attention layers can be regarded as a way to learn the aggregation method. 

\paragraph{VLAD. } We treat the patches before the class-attention layers as individual local descriptors, and aggregate them into a higher-dimensional vector by employing the Vector of locally aggregated Descriptors~\cite{jegou2012aggregating}. 

\paragraph{AMSK. } We also apply the aggregated selective match kernel from Tolias \emph{et al.}~\cite{tolias2016image}. This method was originally introduced for local descriptors, but got adapted to convolutional networks. To the best of our knowledge this is the state of the art on several benchmarks~\cite{tolias2020learning}.  
\medskip

For all these methods, we use the models presented in our main paper, starting from the version fine-tuned at resolution 384$\times$384. By default  the resolution is 768. This is comparable to the choice adopted in the literature for ResNet (e.g., 800 in the work by Berman et al.~\cite{berman2019multigrain}).

\begin{table}[t]
\begin{center}
\caption{\label{tab:instance}
    \textbf{Instance retrieval experiments.} The default resolution is 768. The default class token size is 128 dimensions. The "local descriptor" representation extracted from the activations is in 128 dimensions. To our knowledge the state of the art with ResNet-50 on Holidays with Imagenet pre-training only is the Multigrain method~\cite{berman2019multigrain}, which achieves mAP=92.5\%. Here we compare against this method under the same training setting, i.e., off-the-shelf network pre-trained on Imagenet1k only and with the same training procedure and resolution. We refer the reader to Tolias \emph{et al.} \cite{tolias2020learning} for the state of the art on {$\mathcal {R}$}Oxford, which involves some training on the target domain with images depicted building and fine-tuning at the target resolution. 
    \label{tab:imageretrieval} 
}\vspace*{3pt}
\scalebox{0.9}{\begin{tabular}{l|c|ccc}
\toprule
Base model & parameters & \multicolumn{2}{c}{ROxford5k (mAP)} & Holidays (mAP) \\
\cmidrule{3-4}
 & &  Medium & Hard &   \\
\midrule
\multicolumn{5}{c}{\bf \ours -- class token} \\
\midrule
\ours-S12/16  &                & 30.1 &   8.7    &  86.0     \\
\ours-S12/8   &                & 33.2 & 12.1\pzo &  86.4    \\
\midrule  
\ours-S12/16  & resolution 224       & 12.7 &   2.4    &  71.5 \\
\ours-S12/16  & resolution 384       & 20.1 &   4.6    &  83.4 \\
\ours-S12/16  & resolution 512       & 26.6 &   5.8    &  84.6 \\
\ours-S12/16  & resolution 768       & 30.1 &   8.7    &  86.0 \\
\ours-S12/16  & resolution 1024      & 30.3 & 11.2\pzo &  86.3 \\
\midrule  
\ours-S12/16  & self-supervised DINO & 35.1 & 11.9\pzo &  87.3  \\
\ours-S12/8   & self-supervised DINO & 30.9 &   7.9    &  88.3   \\
\midrule   
\multicolumn{5}{c}{\bf \ours --  VLAD} \\  
\midrule  
\ours-S12/16  & k=256          & 36.6 & 11.6\pzo &  89.9        \\
\ours-S12/16  & k=1024         & 40.0 & 13.0\pzo &  90.7        \\
\midrule
\multicolumn{5}{c}{\bf \ours --  ASMK}\\
\midrule
\ours-S12/8  & k=1024          & 36.5 &   9.4    &  90.4        \\
\ours-S12/8  & k=65536         & 42.0 & 12.9\pzo &  92.3        \\
\ours-S12/16 & k=1024          & 35.2 & 11.5\pzo &  90.4        \\
\ours-S12/16 & k=65536         & 40.0 & 15.0\pzo &  92.0  \\
\midrule
\multicolumn{5}{c}{\bf ResNet-50 --  ASMK}\\
\midrule
Resnet50  & k=1024             & 41.6 &  14.6\pzo   &  86.0    \\
Resnet50  & k=65536            & 41.9 &  14.5\pzo   &  87.9 \\
Multigrain-resnet50 & k=1024   & 32.9 &  9.4        &  87.9 \\
\bottomrule
\end{tabular}}
\end{center}
\end{table}

\subsubsection{Experimental setting: Image retrieval with models pretrained on Imagenet1k only}

We only consider models pre-trained on Imagenet-1k. Note that the literature reports significant improvement when learning or fine-tuning networks~\cite{radenovic2018fine,tolias2020learning}  on specialized datasets (e.g., of buildings for Oxford5k and Paris6k). We consider only \ours-S12 models, since they have a  number of  parameters comparable to that of ResNet-50. 
We report the results in Table~\ref{tab:imageretrieval}. 

\paragraph{Scaling resolution.} As expected increasing the resolution with \ours improves the performance steadily up to resolution 768. This shows that our models are very tolerant to resolution changes considering that they have been fine-tuned at resolution 384. 
The performance starts to saturates at resolution 1024, which led us to keep 784 as the pivot resolution. 

\paragraph{Self-supervision.} The networks \ours pre-trained with self-supervision achieve a comparatively better performance than their supervised counterpart on Holidays, however, we have the opposite observation for {$\mathcal{R}$}Oxford. 

\paragraph{Impact of Image description. } We adopt the class-token as the descriptor, and in our experiments we verified that this aggregation method is better than average and GeM pooling \cite{Boureau2010ATA,radenovic2018fine}. In Table~\ref{tab:imageretrieval} one can see there is a large benefit in employing a patch based method along with our \ours transformers: \OURS-VLAD performs significantly better than the CLS token, likely thanks to the higher dimensionality. This is further magnified with AMSK, where we obtain results approaching the absolute state of the art on Holidays, despite a sub-optimal training setting for image retrieval. This is interesting since our method has not been fine-tuned for retrieval tasks and we have not been adapted in any significant way beyond applying off-the-shelf this aggregation technique. 
A direct comparison with ResNet-50 shows that our \ours method obtains competitive results in this comparable setting, slightly below the ResNet-50 on {$\mathcal{R}$}Oxford but significantly better on Holidays. 

\nocite{thomee2016yfcc100m} %

\subsection{Runtime and Memory Usage}

\begin{figure}[htb]
\centering
\includegraphics[trim=0 0 0 0, clip, scale=0.3]{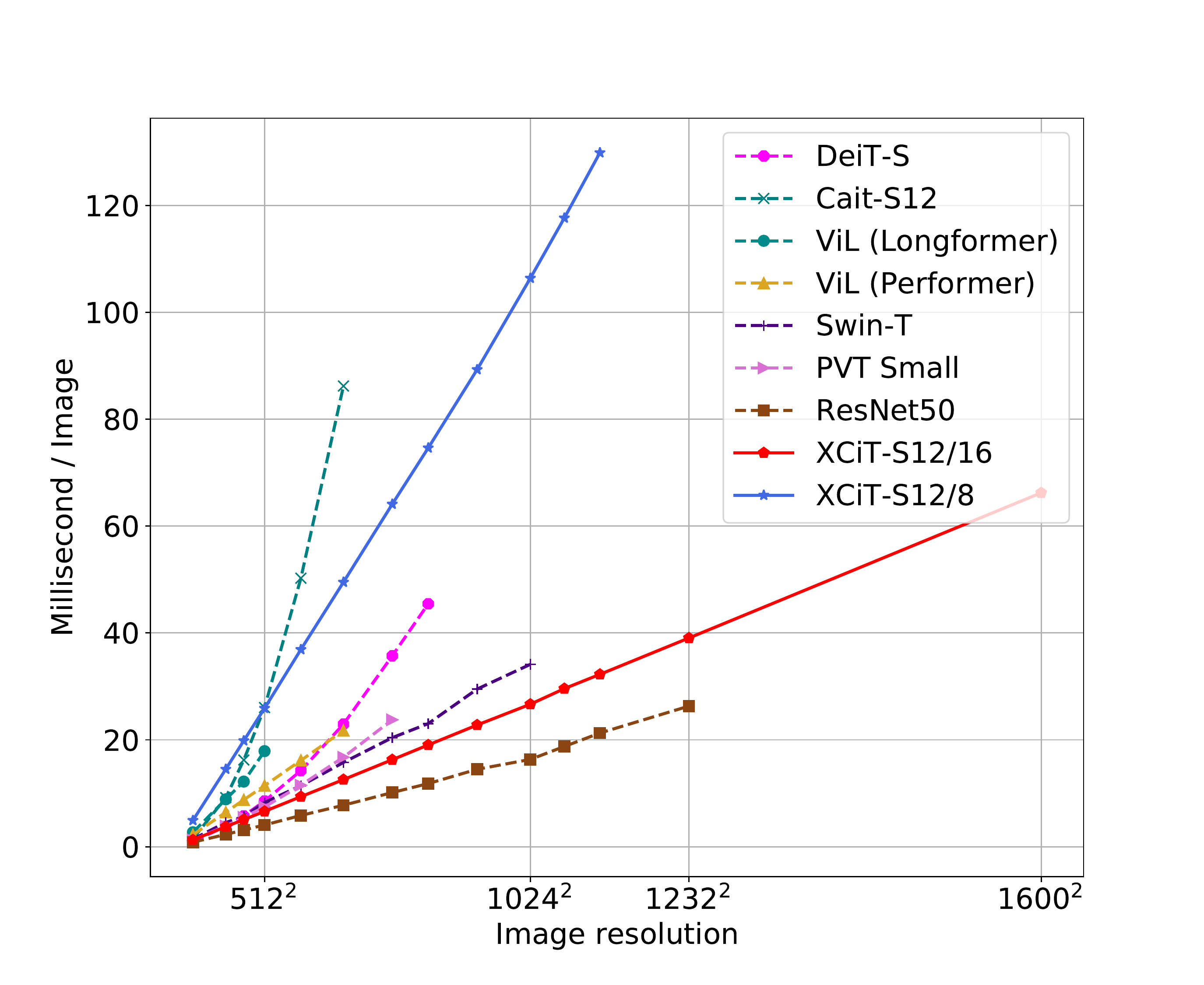}
\scriptsize
\caption{\textbf{We present the millisecond per image during inference of multiple models.} Our \ours-S12/16 model provides a speed up for images with higher resolution compared to existing vision transformers, especially the ones with quadratic complexity like DeiT and CaiT.
\label{fig:throughput}}
\end{figure}

We present the peak memory usage as well as the throughput of multiple models including full-attention and efficient vision transformers in Table~\ref{tab:peakmem}. Additionally, in Figure~\ref{fig:throughput} we plot the processing speed represented as millisecond per image as a function of image resolution for various models. We can observe that \ours provides a strong trade-off, possessing the best scalability in terms of peak memory, even when compared to ResNet-50. Additionally, the processing time scales linearly with respect to resolution, with only ResNet-50 providing a better trade-off on that front.

\begin{table}[ht]
    \centering
    \caption{\footnotesize \textbf{Inference throughput and peak GPU memory usage} for our \OURS small model compared to other models of comparable size that include token self-attention. All models tested using batch size of 64 on a V100 GPU with 32GB memory.  \label{tab:peakmem}}\vspace*{3pt}
    \scalebox{0.74}{
    \begin{tabular}{l | c | c |c | c| c | c| c | c | c | c}
         \toprule
         Model & \#params  & ImNet & \multicolumn{8}{c}{Image Resolution}  \\
         \cmidrule{4-11}
          & ($\times10^6$) & Top-1 & \multicolumn{2}{c|}{224$^2$} & \multicolumn{2}{c|}{384$^2$} & \multicolumn{2}{c|}{512$^2$} & \multicolumn{2}{c}{1024$^2$} \\
         \cmidrule{4-11}
         & & @224 & im/sec & mem (MB) & im/sec & mem (MB) & im/sec & mem (MB) & im/sec & mem (MB) \\
         \midrule
         ResNet-50        & 25 & 79.0 & 1171 & 772      & 434 & 2078    & 245    & 3618 & 61 & 14178 \\ 
         \midrule
         DeiT-S           & 22 & 79.9 & 974 & \pzo433   & 263 & 1580    & 116    & 4020 & N/A & OOM \\
         CaiT-S12         & 26 & 80.8 & 671 & \pzo577   & 108 & 2581    & \pzo38 & 7117 & N/A & OOM \\
         PVT-Small        & 25 & 79.8 & 777 & 1266      & 256 & 3142    & 134    & 5354 & N/A & OOM \\
         Swin-T           & 29 & 81.3 & 704 & 1386      & 220 & 3890    & 120    & 6873 & 29 & 26915 \\
         \rowcolor{Goldenrod} \OURS-S12/16\!\! & 26 & 82.0 & 781 & \pzo731 & 266 & 1372    & 151    & 2128 & 37 & \pzo7312  \\
         \bottomrule
    \end{tabular}}
   
\end{table}

\subsection{Queries and Keys magnitude visualizations}

\begin{figure}[htb]
    \centering
    \vspace{0pt}
    \setlength\tabcolsep{0pt} %
    \renewcommand{\arraystretch}{0.5}
    \scalebox{1.4}{\tiny 
\begin{tabular}{@{\hspace{-0pt}}l@{\hspace{2pt}}c@{\hspace{2pt}}c@{}}
            & $\|{\hat{Q}}\|$ & $\|\hat{K}\|$ \\[3pt]
          \includegraphics[width=0.18\linewidth]{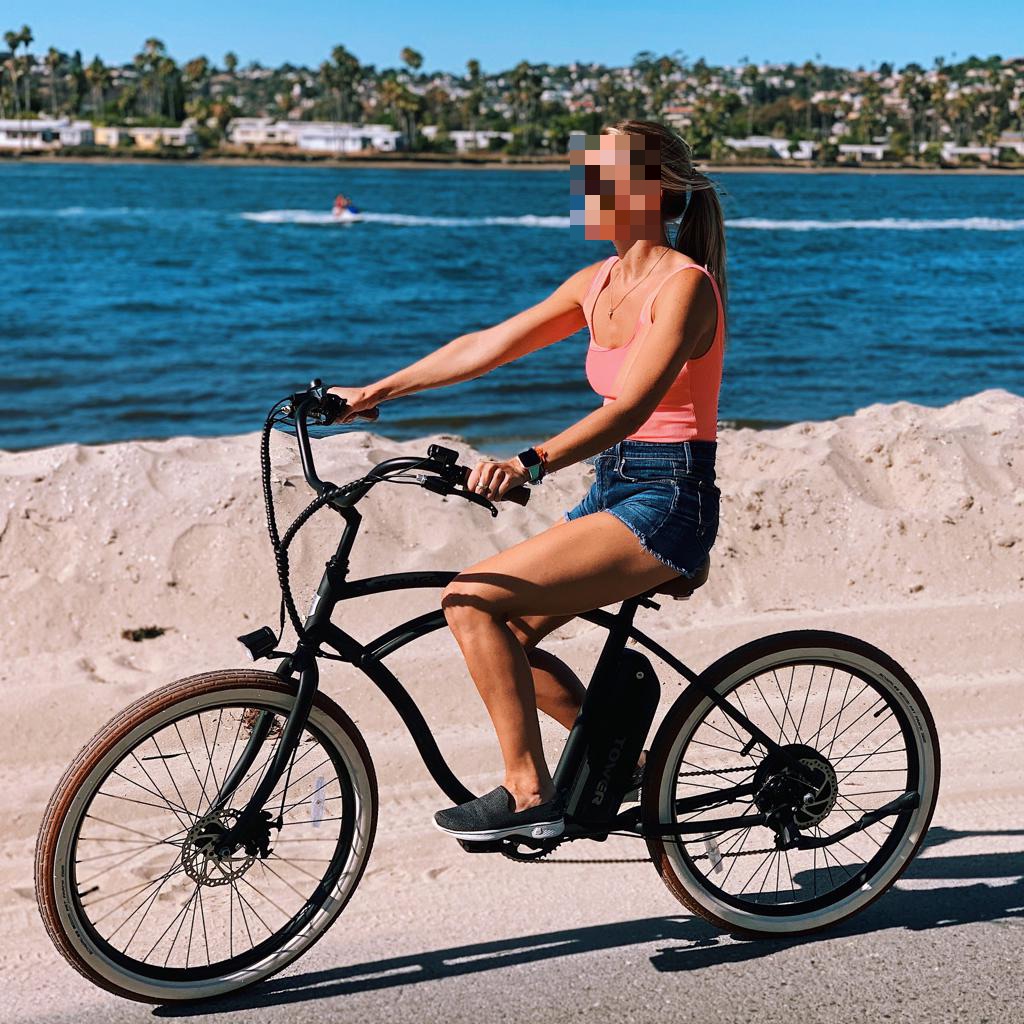} &
          \includegraphics[width=0.18\linewidth]{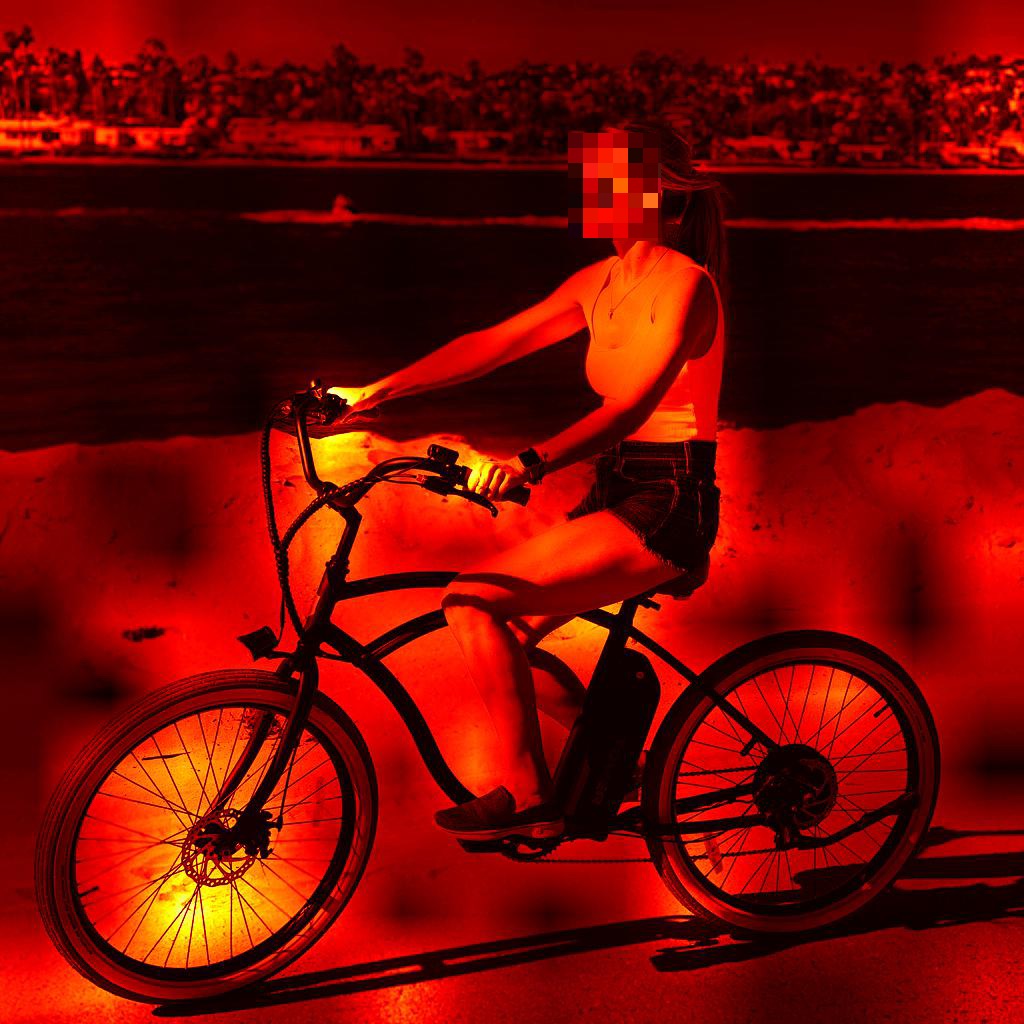} &
          \includegraphics[width=0.18\linewidth]{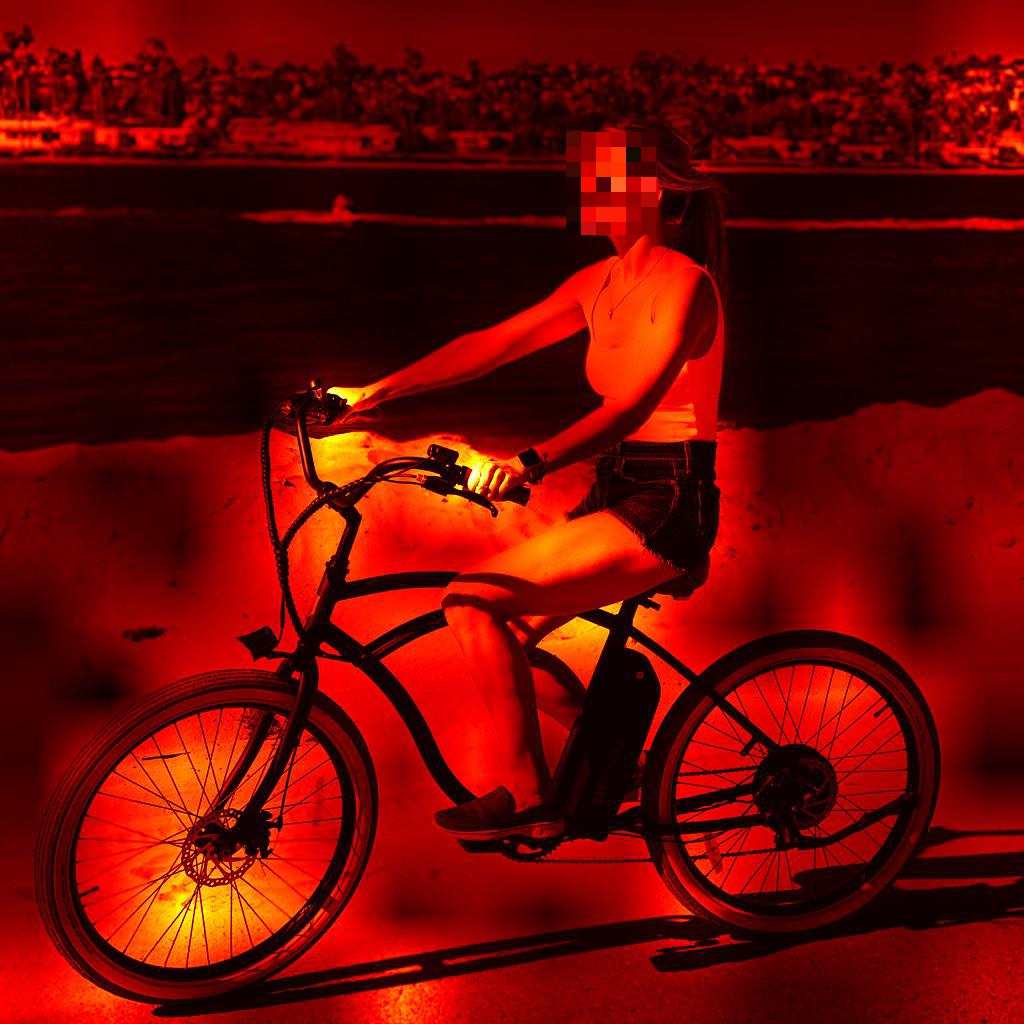} \\ 
          \includegraphics[width=0.18\linewidth]{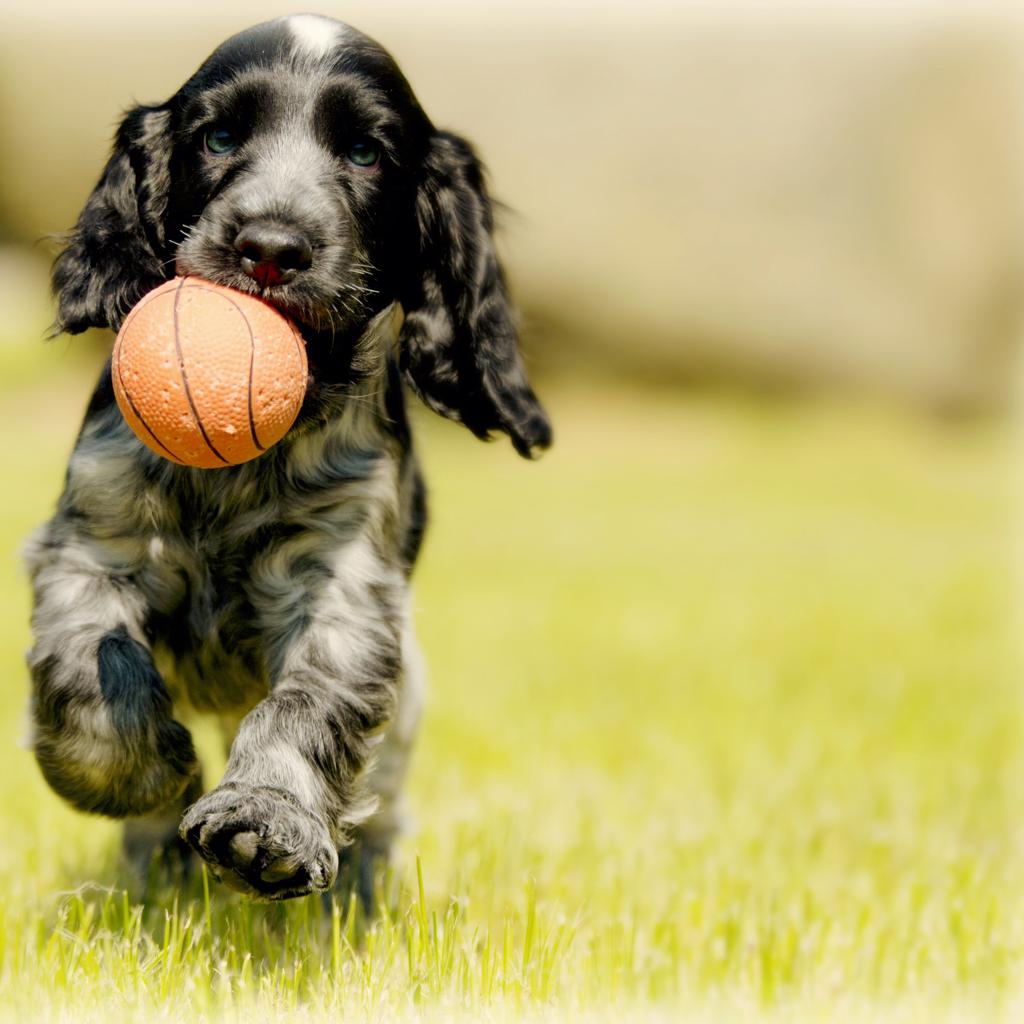} &
          \includegraphics[width=0.18\linewidth]{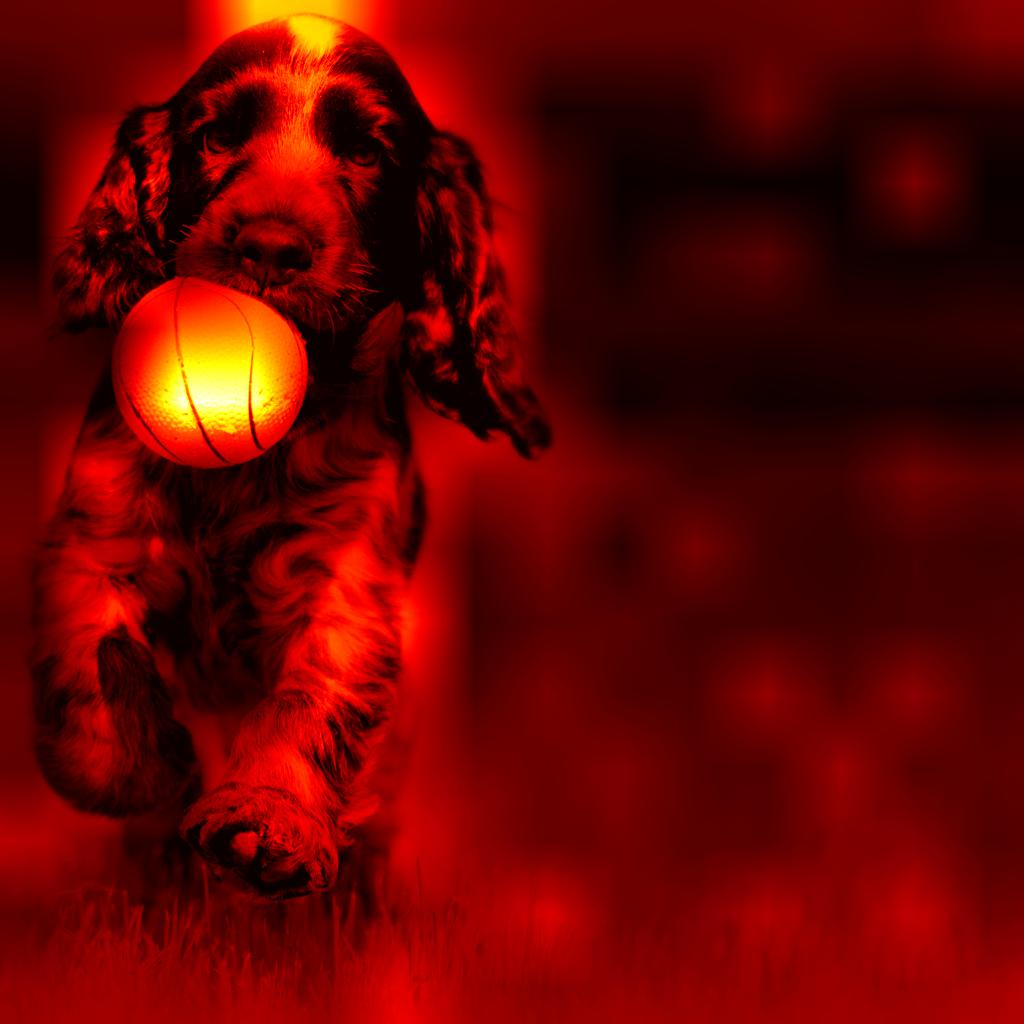} &
          \includegraphics[width=0.18\linewidth]{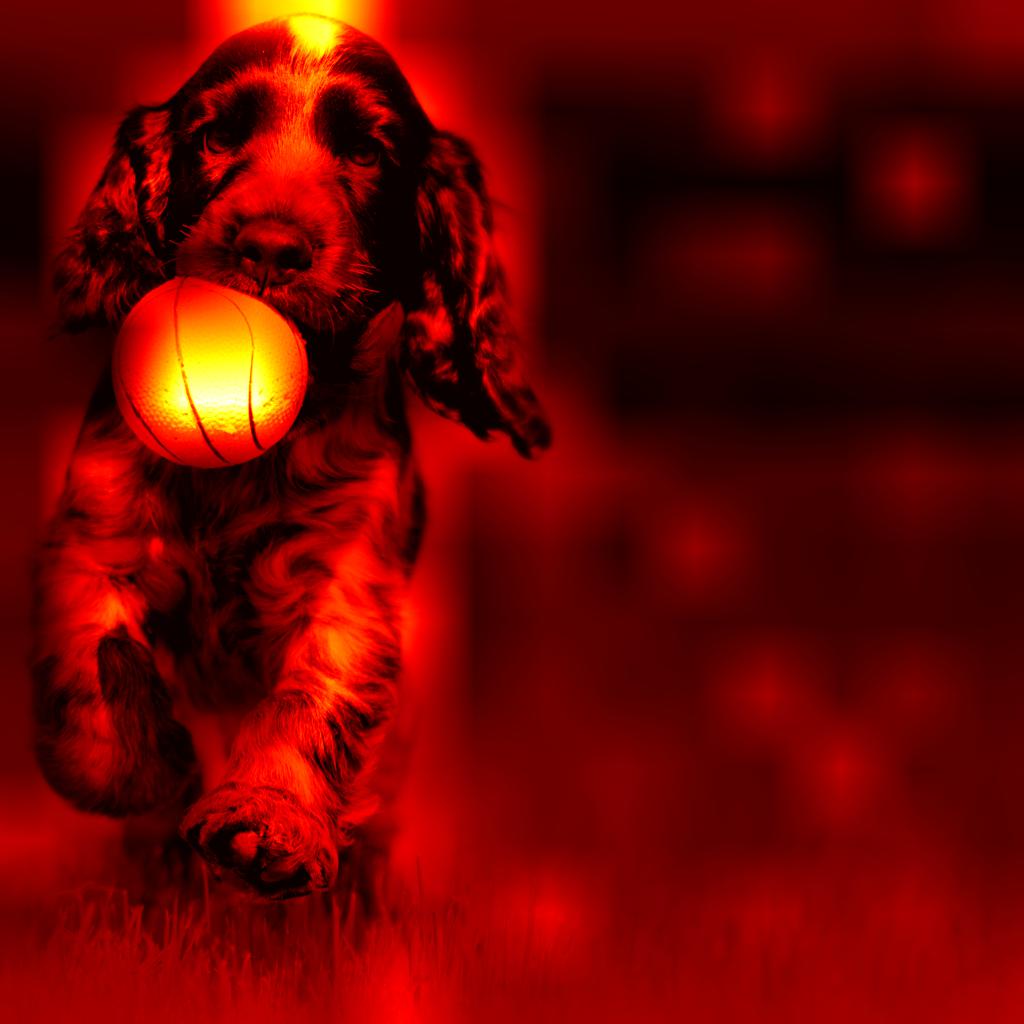} \\
          \includegraphics[width=0.18\linewidth]{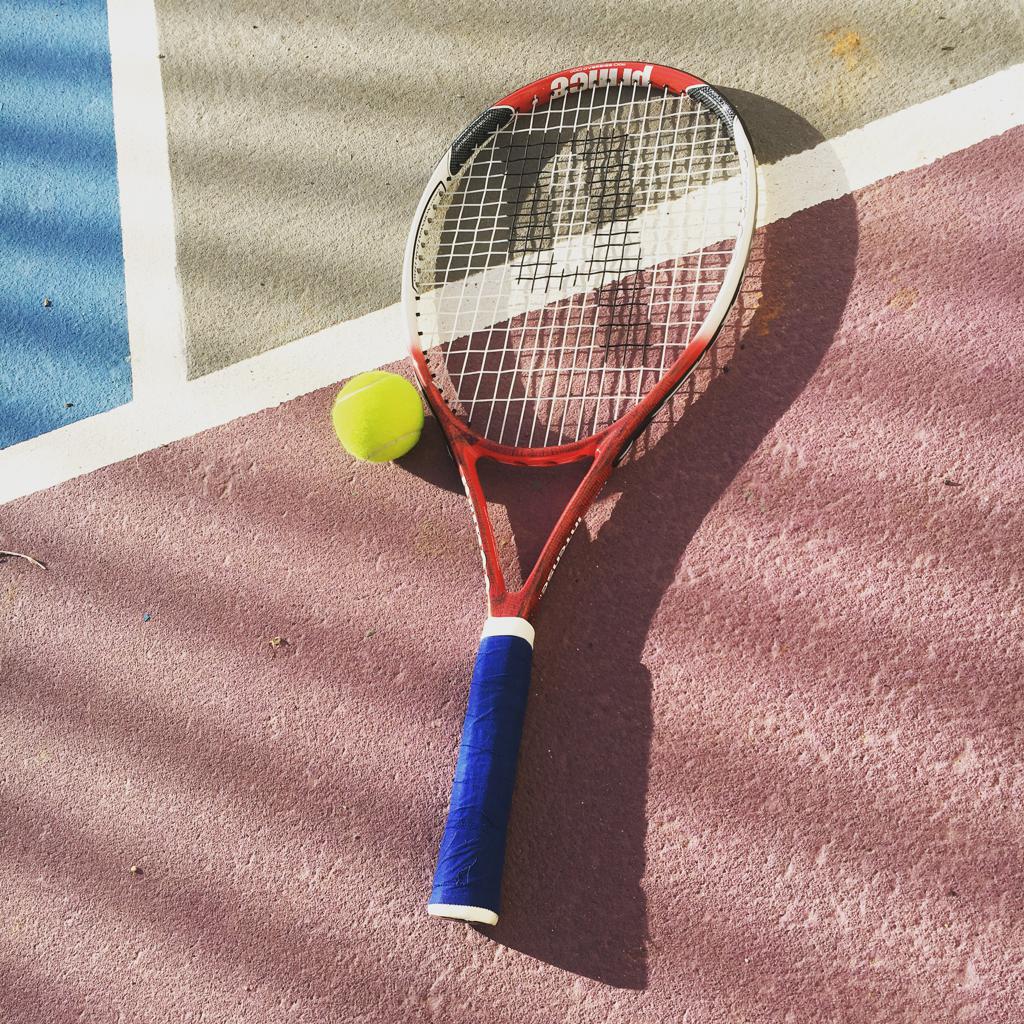} &
          \includegraphics[width=0.18\linewidth]{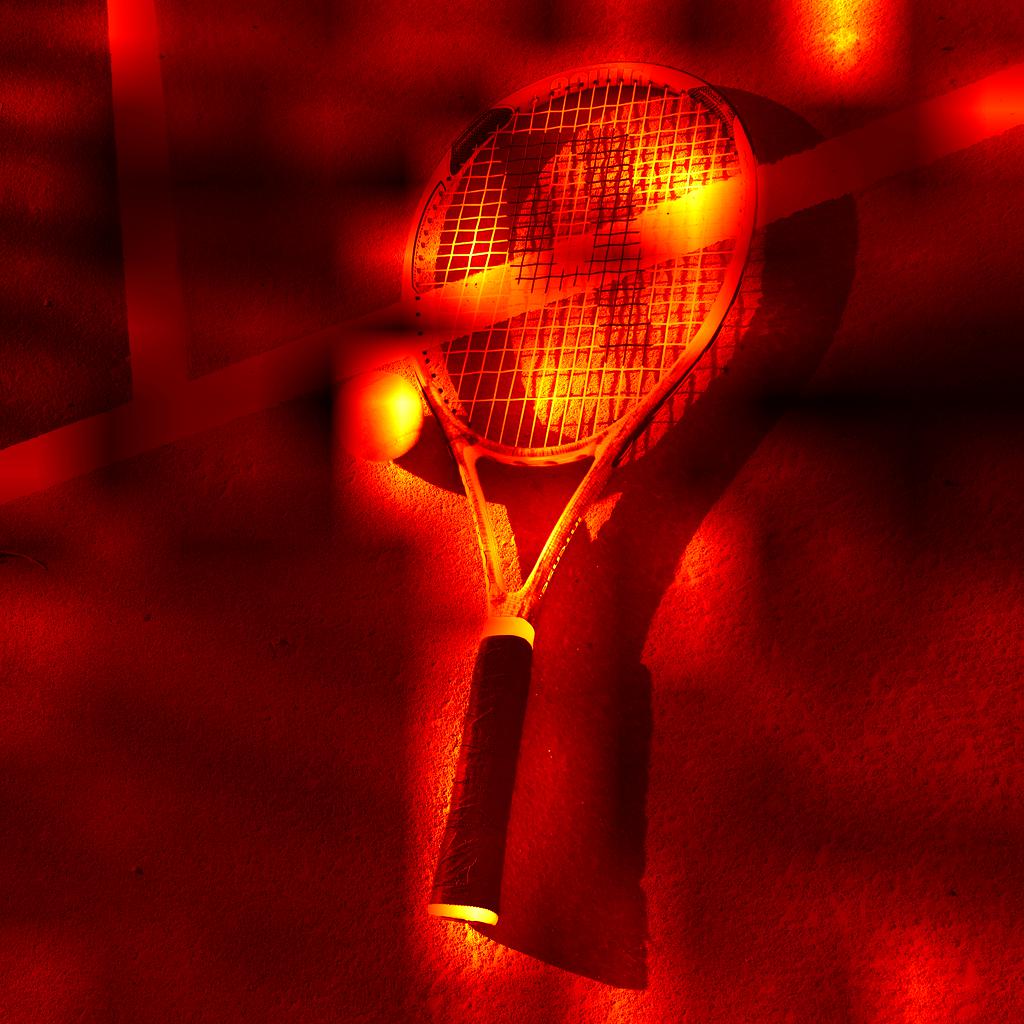} &
          \includegraphics[width=0.18\linewidth]{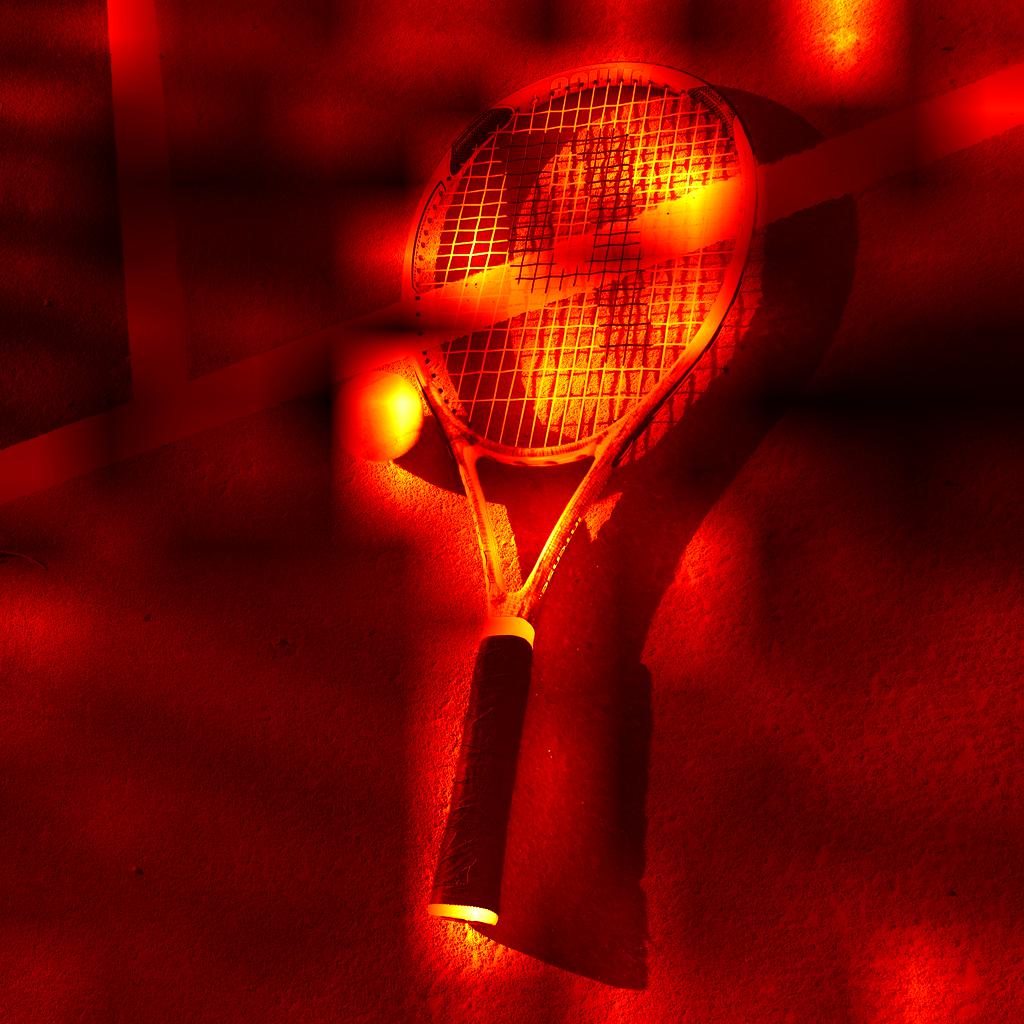} \\
        \end{tabular}
        }%
    \captionof{figure}{
    \textbf{Visualization of the queries $\hat{Q}$ and keys $\hat{K}$ norm across the feature dimension.} We empirically observe that magnitude of patch embeddings in the queries and keys correlates with the saliency of their corresponding region in the image. 
    }
    \label{fig:l2_vis}
\end{figure}

Our XCA operation relies on the cross-covariance matrix of the queries $\hat{Q}$ and keys $\hat{K}$ which are $\ell_2$ normalized across the patch dimension. Therefore, each element in the $d\times d$ matrix represents a cosine similarity whose value is strongly influenced by the magnitude of each patch. In Figure~\ref{fig:l2_vis} we visualize the magnitude of patch embeddings in the queries and keys matrices. We observe that patch embeddings with higher magnitude corresponds to more salient regions in the image, providing a very cheap visualization and interpretation of which regions in the image contribute more in the cross-covariance attention.

\end{document}